\PassOptionsToPackage{numbers, compress}{natbib}
\documentclass{article}

 \usepackage[eandd, final]{neurips_2026}

\usepackage[utf8]{inputenc} %
\usepackage[T1]{fontenc}    %
\usepackage{hyperref}       %
\usepackage{url}            %
\usepackage{booktabs}       %
\usepackage{amsfonts}       %
\usepackage{nicefrac}       %
\usepackage{microtype}      %
\usepackage{xcolor}         %
\usepackage{amsmath}
\usepackage{graphicx} 
\usepackage{multirow}
\usepackage{lipsum}
\usepackage[capitalize]{cleveref}
\usepackage{mathtools}
\usepackage{makecell}
\usepackage{arydshln}
\usepackage{comment}
\usepackage{subcaption}

\title{Do Image Editing Models Understand Lighting?}

\author{%
  \textbf{Tim Küchler}$^{\boldsymbol{\ast}1,3}$ \quad
  \textbf{Johann-Friedrich Feiden}$^{\boldsymbol{\ast}1}$ \quad
  \textbf{Matthias Nießner}$^2$ \quad
  \textbf{Carsten Rother}$^1$ \\[0.2cm]
  $^1$Heidelberg University \quad
  $^2$Technical University of Munich \quad
  $^3$Zuse School ELIZA
}

\begin{document}

\maketitle

\begingroup
\renewcommand{\thefootnote}{}
\footnotetext{$^{\boldsymbol{\ast}}$Equal contribution.}
\endgroup

\begin{abstract}
While recent advancements in generative image editing models have achieved stunning visual fidelity, it remains an open question whether these systems possess an intrinsic knowledge of real-world lighting. %
Existing benchmarks typically evaluate high-level plausibility of perceptual light transport on curated internet imagery, using VLMs or human judgement, or they rely on synthetically generated datasets. 
In this work, we introduce the 3D-anchored Light Probe (3DLP) benchmark, for which we have captured a new high-fidelity HDR dataset of real-world lighting changes.  
The dataset consists of 1K image pairs of diverse indoor scenery in which light probes are physically turned on and off. 
To allow for a granular performance analysis, we annotated specific image regions such as cast shadows or metallic surfaces. 
With this data, we evaluate a range of state-of-the-art image editing models by measuring how well their light probe edits align with reality. The evaluation uses two new scores to compensate for AI-generated photographic effects, such as adjusted white balance.     
Our results show that the overall performance of models differs considerably, with differences slightly less pronounced for specular highlights. The best image editing models are remarkably consistent with real-world physics, however, they still leave room for improvement. 
We observe that image regions that receive less light from the light probe are more prone to errors for all models.     
Furthermore, building on their success in evaluating macroscopic lighting plausibility, we test VLMs on our task but find that they are unsuitable for pixel-level light transport analysis.
We will make the benchmark, together with the real-world dataset, publicly available to encourage future research on this topic. 

\textbf{Project page:} \url{https://tim-kuechler.github.io/3DLP}
\end{abstract}

\section{Introduction}
\label{sec:intro}

\begin{figure}[t]
    \centering
    \includegraphics[width=\textwidth]{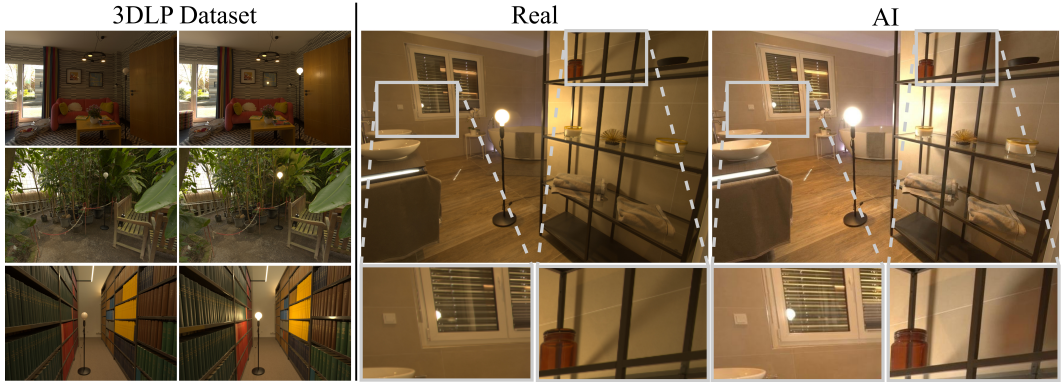}
    \caption{\textbf{Our 3DLP Task.} (Left) Three examples from our 3DLP dataset with the light probe turned off and on, respectively. The dataset includes a wide range of materials, geometries and ambient lighting effects. (Right) A result from a strong image editing model (Nano Banana 2) for the turn-on task, compared to the real image. The image editing model has turned on the light with a different brightness, and the photo has a different exposure. Our proposed metrics are invariant to such effects. Zooming in, we see that the model does remarkably well in casting new shadows, adding reflections, and leaving existing ambient lights unaltered. In our 3DLP benchmark we conduct a detailed analysis of individual effects.\vspace{-0.5cm}} 
    \label{fig:Teaser}
\end{figure}

Instruction-based image editing has advanced tremendously in recent years. While older models~\cite{brooks2023instructpix2pix, openai2025gptimage1, labs2025flux1kontext} often struggled with scene consistency, unintentionally altering unprompted regions or producing noticeable misalignments~\cite{openai2025gptimage1}, the latest generation of editing models~\cite{wu2025qwenimagetechnicalreport, flux2_max_2025, blackforestlabs2025flux2dev, google2026nanobanana2, google2025nanobananapro, openai2025gptimage15} achieves near pixel-perfect scene preservation, while performing only the targeted edits. This critical advancement in isolating edits allows us to perform physically based benchmarking at a pixel-level rather than relying on global plausibility. As a result, these systems are rapidly becoming practical tools in professional visual media, enabling complex tasks such as seamlessly inserting virtual products into commercial photography~\cite{chittersu2025insert, adobe2026firefly} or relighting background plates for visual effects~\cite{beeble2025switchlight3}. Furthermore, open-source variants~\cite{wu2025qwenimagetechnicalreport, blackforestlabs2025flux2dev, podell2023sdxl, rombach2022high} frequently serve as foundational backbones for downstream vision pipelines, including intrinsic image decomposition~\cite{zeng2024rgb, ke2025marigold}, depth prediction~\cite{ke2023marigoldoriginal, zavadski2024primedepth}, and light estimation~\cite{phongthawee2024diffusionlight, tong2025spatiotemporally, Chinchuthakun2025DiffusionLightTurbo}. A key question thus arises: do these models genuinely understand lighting - specifically, can they consistently synthesise relighting effects, accurately insert virtual objects, and model illumination across varying materials - or are they merely reproducing appearance priors that look perceptually plausible to humans? This distinction is vital, particularly for downstream tasks that implicitly rely on their base models to possess a robust understanding of illumination. Currently, it remains unclear whether performance bottlenecks in these applications stem from task-specific methodological limitations, or from base models that lack physical grounding. This critical ambiguity motivates our work to rigorously benchmark the lighting capabilities of foundational image editing models.

Despite this critical need, the physically grounded evaluation of image editing models remains underdeveloped. Existing benchmarks provide essential frameworks for measuring semantic faithfulness~\cite{pu2025picabench, sun2025t2i, meng2024phybench} and instruction adherence~\cite{li2024genai, zhu2025gobench} on regional~\cite{sarkar2024shadows, pu2025picabench} or global~\cite{sun2025t2i, li2024genai, zhu2025gobench, meng2024phybench} scales. Frequently, Vision-Language Models (VLMs)~\cite{pu2025picabench, meng2024phybench, sun2025t2i} or human judgement~\cite{zhu2025gobench, li2024genai} are employed to assess physical visual plausibility. While these evaluation schemes offer a valuable signal, they are inherently optimised for perceptual alignment rather than radiometric accuracy. Consequently, they struggle to quantify the pixel-level physical correctness of intricate light transport effects. A common strategy to mitigate this limitation is the use of synthetic datasets, which have been successfully leveraged across a wide variety of tasks, including intrinsic image decomposition~\cite{kocsis2025iif, zeng2024rgb, Zhu_2022_CVPR, liang2026pi, he2025unirelight} and relighting~\cite{liang2026pi, he2025unirelight, serrano2026synclight, magar2025lightlab, liang2026luxremix}. Prominent examples rely on renderings from open-source environments like Hypersim~\cite{roberts2021hypersim}, Infinigen~\cite{infinigen2024indoors}, OpenRooms~\cite{li2021openrooms} or InteriorVerse \cite{zhu2022learning}. Another line of work captures high-fidelity data under controlled lighting, such as light stages. Although these setups yield highly accurate measurements, One-Light-at-a-Time (OLAT)~\cite{zhou2025olatverse} environments are typically restricted to isolated objects rather than full, complex scenes. While approaches leveraging VLMs, synthetic datasets, and OLAT captures have driven tremendous progress, evaluating generated lighting effects within diverse, real-world environments remains a formidable challenge. Addressing this gap is particularly vital for the evaluation of modern generative image editing methods, which serve as the backbone for numerous downstream applications.

In this work, we address this gap by introducing the 3D-anchored Light Probe (3DLP) benchmark, designed for evaluating light transport consistency using real-world data (see \cref{fig:Teaser}). 
To this end, we captured 1K high-fidelity HDR image pairs across a wide array of diverse real-world indoor environments. An image pair captures the identical scene with a light probe turned on and off, where both the bulb and the base of the light probe are visible in the image.
To enable granular analysis, we further provide fine-grained annotations for critical optical phenomena, including cast shadows, specular highlights, and challenging interactions with metallic, mirrored, and transparent surfaces. We also introduce two new physically motivated metrics. These scores allow the editing model to freely choose the temperature and brightness of the turned-on light and to also adjust the white balance and exposure of the edited photograph. Our analysis across six state-of-the-art image editing models shows that the model performance differs considerably, slightly less pronounced for specular highlights.
The leading systems have achieved a remarkable overlap with real-world physics, but still leave room for improvement. In particular, image regions that receive less light from the light probe perform poorly.      

To summarise, our contributions are as follows
\begin{itemize}
\item A new dataset of 1K high-fidelity HDR image pairs that capture diverse real-world indoor environments. For granular performance analysis, we manually labelled challenging image regions such as shadows, highlights, and metallic surfaces.
\item The 3D-anchored Light Probe (3DLP) benchmark with two new evaluation metrics, allowing image editing models to freely choose the temperature and brightness of the turned-on light and adjust the exposure and white balance of the full image. 
\item We systematically evaluate six cutting-edge image editing models, and observe that leading models have a remarkably good understanding of light transport. The top-ranked model is the commercial model Nano Banana Pro and the third-ranked is the open-source model Qwen-Image-Edit.

\end{itemize}

\section{Related work}
\label{sec:relatedwork}

\paragraph{Instruction-based image editing models} Instruction-based image editing has evolved from simple object replacements~\cite{hertz2022prompt, avrahami2023blended} and limited editing capabilities~\cite{brooks2023instructpix2pix, bar2022text2live} to complex, global edits like lighting adjustments~\cite{google2025nanobananapro, blackforestlabs2025flux2dev, wu2025qwenimagetechnicalreport}. Historically, even advanced models suffered from unintended stylistic drift and background degradation~\cite{openai2025gptimage1, labs2025flux1kontext, hidreami1technicalreport}, rendering them inapplicable for pixel-aligned physics benchmarking where non-targeted edits falsely penalise physical accuracy. Consequently, our benchmark exclusively evaluates state-of-the-art commercial~\cite{openai2025gptimage15, google2026nanobanana2, google2025nanobananapro, flux2_max_2025} and open-source~\cite{blackforestlabs2025flux2dev, wu2025qwenimagetechnicalreport, wu2025omnigen2, deng2025emergingbagel} models capable of processing universal prompts while preserving untargeted regions. While some are specialised editing models~\cite{wu2025qwenimagetechnicalreport}, many are foundational multimodal models, also supporting text-to-image generation~\cite{flux2_max_2025, blackforestlabs2025flux2dev, google2026nanobanana2, google2025nanobananapro, openai2025gptimage15, wu2025omnigen2, deng2025emergingbagel}. It is worth noting that there are recent task-specific relighting approaches~\cite{serrano2026synclight, magar2025lightlab, liang2026luxremix} that are also capable of tasks such as turning on a light probe that is visible in an image. Due to the lack of publicly available source code and model weights, we were unable to include these methods in our evaluation. Other relighting methods~\cite{zhang2025iclight, jin2024neural, zeng2024dilightnet, bharadwaj2025genlit} are open source, but are not applicable to our task because they do not allow the direct control over individual lamps or do not apply relighting on the entire scene.

\paragraph{Image generation and editing benchmarks} Image generation and editing models are typically benchmarked across diverse domains, evaluating prompt adherence~\cite{wang2026everything, wang2025lmm4lmm, li2024genai, huang2024paralleledits, zhang2023magicbrush, wang2023imagen, sun2025t2i}, spatial positioning and transformations~\cite{wang2026everything, wang2025genspace, zhao2025envisioning, krojer2024learning, mahajan2025undo, wang2023imagen}, logic, causal effects and common sense~\cite{wang2026everything, meng2024phybench, fu2024commonsense, zhao2025envisioning, mahajan2025undo, wu2025kris, niu2025wise, sun2025t2i}, and broad physical phenomena~\cite{gupta2026physicsbasedbenchmarkingmetricsmultimodal, pu2025picabench, meng2024phybench, zhu2025gobench, sarkar2024shadows}. These benchmarks generally rely on one of three evaluation schemes: (1) semantic or pixel-level metrics (e.g., CLIP, DINO, $L_1$, classifier models)~\cite{sarkar2024shadows, krojer2024learning, mahajan2025undo, ma2024i2ebench}, which primarily capture surface-level similarity; (2) Vision-Language Models (VLMs)~\cite{gupta2026physicsbasedbenchmarkingmetricsmultimodal, pu2025picabench, meng2024phybench, wang2025genspace, wang2023imagen, wang2026everything, fu2024commonsense, zhao2025envisioning, wang2025lmm4lmm, wu2025kris, niu2025wise, sun2025t2i}, which assess global plausibility but often hallucinate or miss local physical inconsistencies~\cite{pu2025picabench}; or (3) human evaluation~\cite{zhu2025gobench, fu2024commonsense, wang2025lmm4lmm, li2024genai, wang2023imagen}, which is costly, unscalable, and prone to bias. While these benchmarks demonstrate strong alignment with human judgement~\cite{pu2025picabench}, they assess spatial and physical edits via high-level semantic plausibility. By relying on categorical grading schemes or binary condition checks, they establish a vital baseline for perceptual realism but leave a crucial gap for precise, quantitative, pixel-level evaluation against objective physical metrics. To address this, our benchmark strictly compares edits on the pixel-level against captured real-world ground truths.

\paragraph{Relighting evaluation datasets} 
One approach to capture high-fidelity data is to use controlled lighting, such as a light stage as in a One-Light-at-a-Time (OLAT) system, e.g.~\cite{zhou2025olatverse, toschi2023relight, yang2026ictpolarreal}. While light stages provide highly accurate measurements, it is typically restricted to capture isolated objects rather than full, complex scenes~\cite{zhou2025olatverse}. 
To capture full scenes with varying illumination, most works~\cite{he2025unirelight, zeng2024rgb, kocsis2025iif, magar2025lightlab, serrano2026synclight, liang2026pi, liang2026luxremix} resort to synthetic datasets. Examples include renderings from open-source environments like Hypersim~\cite{roberts2021hypersim}, Infinigen~\cite{infinigen2024indoors} or OpenRooms \cite{li2021openrooms}, as well as proprietary internal 3D assets~\cite{he2025unirelight}. 
With respect to real data, little is available for the particular relighting task we are interested in, i.e. scenes where a toggled light probe is visible.  
In~\cite{magar2025lightlab} the authors used an internal dataset of 600 image pairs with lights turned on and off. However, this unpublished dataset is of considerably lower quality than ours, since it was captured with a mobile device not focusing on high precision radiance. Furthermore, their data deliberately includes diverse and complex lamp geometries and we show experimentally that the quality of light transport can be measured less well for complex lamps. 
In concurrent work,~\cite{serrano2026synclight} proposed a small real-world dataset of about 40 such image pairs. In contrast to these works, we capture a high-fidelity, HDR dataset of 1K image pairs.

\section{Method}

\subsection{3DLP Benchmark}
\label{subsec:3DLP_benchmark_method}

\begin{figure}[t]
    \centering
    \includegraphics[width=\textwidth]{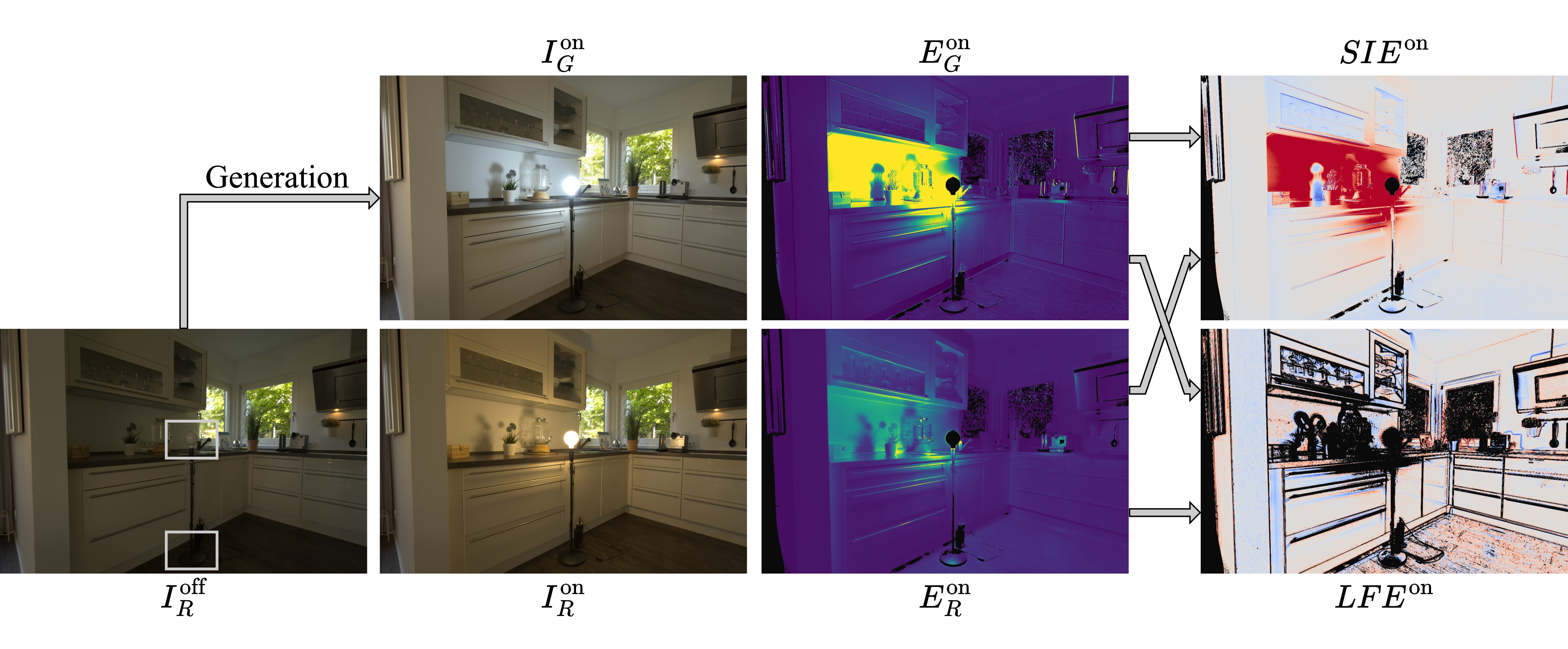}
    \caption{\textbf{3DLP Pipeline} illustrating the turn-on task. (Left) Image with a light bulb turned off, $I^{\textrm{off}}_{R}$, where the stand as well as the bulb are visible (white boxes). The model is tasked to turn on the bulb with a bright, white light, producing $I^{\textrm{on}}_{G}$. To isolate the light transport contribution of the single light bulb, we compute the ratio image of the generative model, $E^{\textrm{on}}_{G} = I^{\textrm{on}}_{G}/I^\textrm{off}_R$, and the real ratio image, $E^{\textrm{on}}_{R} = I^{\textrm{on}}_{R}/I^\textrm{off}_R$.
    Black areas indicate regions masked out due to clipping, low signal, or window labels.
    The two measures (right) compare the ratio images. The Standardised Intensity Error
    (SIE), visualised as map $\mathcal{S}(E^{\textrm{on}}_{R}) - \mathcal{S}(E^{\textrm{on}}_{G})$, shows that the pixels left to the light bulb have a high intensity error (red). In contrast, the Low-Frequency Error (LFE), visualised as a map $\mathcal{S}(M^{\textrm{on}}_R) - \mathcal{S}(M^{\textrm{on}}_{G})$, penalises errors in gradients. Here, pixels with high intensity error (SIE) can have a low LFE error. Note that LFE is additionally masked by the 80th percentile highest gradients in $E^{\textrm{on}}_{R}$ or $E^{\textrm{on}}_{G}$. Also note that both errors are invariant to photographic effects of the model, such as different exposure, as well as the colour and intensity of the light bulb. 
    \vspace{-0.3cm}
    }
    \label{fig:teaser_pipeline}
\end{figure}

We propose the 3D-anchored Light Probe (3DLP) benchmark, which is based on real-world captured image pairs. 
We chose a binary task: Turning on and off a single spherical light bulb (``light probe'') mounted on a floor lamp. By ensuring that the base of the lamp is clearly visible, we ``anchor'' the light probe within the scene, allowing models to infer its precise 3D position, which is a prerequisite for accurate light transport. This setup, combined with the radial emission pattern of the bulb, ensures that the benchmark evaluates a model's physical light transport capabilities rather than its ability to interpret vague prompts or ambiguous lamp geometries. While the light source is simple, the task remains non-trivial as the model has to decouple the probe's contribution from the existing complex ambient light and accurately simulate its physical interaction with the scene's diverse materials and geometry.
Formally, we utilise real-world ground-truth radiance measurements $I^\text{on}_R$ and $I^\text{off}_R$ for the on and off state, respectively. The model is tasked with performing the opposite of the input state: Given $I^\text{off}_R$ to produce an ``on'' state, $I^\text{on}_{G}$, and vice versa. 
The input image to the model is passed as an 8-bit, tonemapped sRGB image. The model's output is subsequently linearised for evaluation. As shown in the supplement (\cref{supp:quantisation_uncertainty}), 8-bit linearisation has a negligible effect on metric accuracy.
Rather than relying on subjective human judgement or VLMs, we evaluate $I^\text{on/off}_\text{G}$ against the ground-truth measurements that capture all the intricate effects of real-world light transport, using the metrics defined in \cref{sec:metrics}. An overview of this pipeline is given in \cref{fig:teaser_pipeline}. We will make the 3DLP benchmark, alongside the real-world dataset, publicly available.

\subsection{Dataset capturing and annotation}
\label{subsec:Dataset}

For public release, we have compiled a dataset of 1K image pairs, captured in diverse indoor scenes, each featuring a fully visible, 3D-anchored 
light probe within the image. Each pair consists of a light-on and a light-off high dynamic range (HDR) image. For every individual HDR image, we captured two 14-bit, 9-exposure brackets with an exposure value (EV) stop of 1.0 per image. These two brackets are subsequently averaged (stacking) to reduce noise and merged into a single HDR representation. The resulting image is then corrected for lens distortion and vignetting effects utilising the EXIF data and saved in a linear format. Across views, the focus distance is fixed at around 1.5\,m with a constant aperture of f/8.0.
To ensure that the dataset remains applicable to potential future applications, we recorded exactly five image pairs per camera viewpoint. All pairs from a single viewpoint share the same base exposure time, set according to the light-on illumination, so that all images have the same ambient light noise level. Since the model needs a well-lit input image and the off-image $I^\text{off}_R$ is oftentimes underexposed, we re-adjust the image post-capture to a baseline exposure (0 EV shift) using the EXIF data. For consistency, the generated on-image $I^\text{on}_{G}$ is converted back to the same base exposure time of the real on-image $I^\text{on}_R$. 
While this rigorous procedure requires an acquisition time of approximately 5 minutes per image pair, it yields exceptionally high-quality, low-noise captures. To maximise utility, the final linearised HDR images will be published alongside their corresponding EXIF metadata.

We manually annotated three types of regions in the images: i) challenging materials (metallic, mirror or transparent), ii) regions that have clear shadows or highlights cast by the light probe, iii) regions that have ambient highlights or shadows visible in the off-image. 
Examples of individual cases are illustrated in \cref{fig:Teaser,fig:ambientfailure,fig:Transparent}, and more are included in supplemental \cref{supp:dataset}.
The annotation was performed with a polygon tool. Our aim was not to reach completeness but to have correct and consistent annotations. Overall, 10.92\% of pixels were assigned a label.

\subsection{Evaluation metrics}
\label{sec:metrics}

While prior works~\cite{magar2025lightlab, serrano2026synclight, liang2026luxremix, kocsis2025iif, zeng2024rgb} use PSNR, SSIM~\cite{wang2004imagessim}, and LPIPS~\cite{zhang2018unreasonablelpips} to assess relighting accuracy, these metrics suffer from several weaknesses for our purpose. In particular, they are not invariant to global adjustments of the gen. model, e.g. white balance, and a model which does nothing sometimes even scores best (\cref{supp:comparison_SIE_LFE_vs_PSNR_SSIM_LPIPS}). Consequently, we introduce two new metrics that overcome these limitations. %

Our benchmark evaluates how accurately generated light transport matches the real ground-truth 
light transport for two tasks, turning \textit{on} and \textit{off} a light bulb. We observe that some image editing models~\cite{flux2_max_2025, blackforestlabs2025flux2dev} mimic human photographers by adjusting global exposure and white balance ($K$) when a new light source is introduced or an existing one is removed. Additionally, we lack control over the exact colour and intensity of the generated light source ($C$)\footnote{If not done by default by the generative model, the prompt includes ``bright, white light'' to encourage large changes in intensity.}. We therefore formulate our testing hypothesis for the \textit{turn-on} task as
\begin{equation}
I^{\textrm{on}}_{G} = C \odot L_R + K \odot I^{\textrm{off}}_R,
\label{eq:AIassumption_on}
\end{equation}
where $\odot$ denotes the Hadamard product and $C>0$. The radiance of the light bulb is defined by $L_R$, and $I^{\textrm{on}}_R = I^{\textrm{off}}_R + L_R$.
While $I$ and $L$ are spatially varying, $C$ and $K$ are image-level constant 3D vectors. Dividing by the background image $I^\textrm{off}_R$ gives the ratio image $E^{\textrm{on}}_{G}\coloneq I^{\textrm{on}}_{G}/I^\textrm{off}_R$, which is related to the ground-truth ratio $E^{\textrm{on}}_R\coloneq I^{\textrm{on}}_R/I^\textrm{off}_R$ by an unknown per-channel affine transformation
\begin{equation}
E^{\textrm{on}}_{G}
=
C \odot E^{\textrm{on}}_R + K - C.
\label{eq:EnergyRatios_on}
\end{equation}
The two $E^{\textrm{on}}_{R/G}$ images are visualised in \cref{fig:teaser_pipeline}. We see that they capture only the intensity of the turned-on light bulb. We masked out clipped and low-signal regions, as well as regions being labelled as an uncovered window (see supplement \cref{supp:Proofs_and_Scores} for details).

For the \textit{turn-off} task, we use the analogous hypothesis: The model may apply a global exposure or white-balance change to the on-image and should subtract the real light transport field,
\begin{equation}
I^{\textrm{off}}_{G}
=
K \odot I^{\textrm{on}}_R - C \odot L_R.
\label{eq:AIassumption_off}
\end{equation}
We define $E^{\textrm{off}}_{G}\coloneq I^{\textrm{off}}_{G}/I^{\textrm{on}}_R$ and $E^{\textrm{off}}_R\coloneq I^{\textrm{off}}_R/I^{\textrm{on}}_R$. Since $I^{\textrm{off}}_R=I^{\textrm{on}}_R-L_R$, we arrive at a symmetric equation as for the turn-on case
\begin{equation}
E^{\textrm{off}}_{G}
=
C \odot E^{\textrm{off}}_R + K - C.
\label{eq:EnergyRatios_off}
\end{equation}
As above, $E^{\textrm{off}}_{G}$ and $E^{\textrm{off}}_R$ are related up to a global per-channel affine transformation. To compare the ratio images objectively, our metrics must therefore be invariant to the unknown global scale ($C$) and shift ($K$).
Note that the invariance with respect to $C$ is not strictly required for the turn-off task, as the on-image already defines the lamp's intensity and colour. However, maintaining $C$-invariance ensures our metrics remain symmetric and comparable across both tasks. Although this invariance fails to penalise a model that only dims the lamp instead of turning it fully off, we do not observe this phenomenon among the tested models.

\paragraph{Standardised Intensity Error (SIE)} This metric measures how accurately the generative model predicts the changes in light intensity across the scene. %
We map the ratio images $E_R^\textrm{on/off}$ and $E^\textrm{on/off}_{G}$ to a standardised space to achieve invariance to $C$ and $K$. Because generated images can exhibit artefacts, e.g. high-frequency noise, and the assumed Gaussian noise in sRGB images becomes long-tailed during linearisation, we utilise a robust standardisation operator $\mathcal{S}(\cdot)$ based on the median and the Median Absolute Deviation (MAD):
\begin{equation}
\mathcal{S}(X) = \frac{X - \text{median}(X)}{\text{MAD}(X)},
\label{eq:Standardisation}
\end{equation}
with elementwise division. The Standardised Intensity Error is then computed as the Mean Absolute Error (MAE) over the image and colour channels between the two standardised ratio images:
\begin{equation}
\textrm{SIE}^t = \text{MAE}(\mathcal{S}(E_{R}^t), \mathcal{S}(E^t_{G})),\quad t\in\{\textrm{on},\textrm{off}\}.
\label{eq:SIE_on}
\end{equation}

\paragraph{Low-Frequency Error (LFE)}

This metric assesses the physical realism of the decay of light as well as surface interactions. By focusing on the first-order derivatives of the ratios, we evaluate whether the model captures the smooth spatial transitions, which are characteristic of the inverse-square fall-off and Lambertian shading. This is in contrast to the intensity-based SIE metric. 
Raw spatial gradients are dominated by high-frequency edges, which generative models often fail to align perfectly and which are typically caused by texture or geometry rather than the light fall-off itself. We therefore remove strong edges before comparing gradients. For each colour channel, we compute the Sobel gradient magnitudes of both the generated ratio image and true ratio image,
\begin{equation}
M_E=|\nabla E| .
\label{eq:Gradient}
\end{equation}
We form an evaluation mask of those pixels for which the generated and real gradient magnitudes both lie below the 80th percentile.
Let $\Omega_{80}$ denote this edge-filtered evaluation mask. We robustly standardise the retained gradient magnitudes and compute
\begin{equation}
\textrm{LFE}^t = \text{MAE}_{\Omega_{80}}(\mathcal{S}(M_R^t),\mathcal{S}(M_{G}^t)),\quad t\in\{\textrm{on},\textrm{off}\}.
\label{eq:LFE_on}
\end{equation}
It is important to note that both scores, SIE and LFE, apply to arbitrary materials and textures, since each colour channel is compared separately. Proofs for invariance of SIE and LFE with respect to $C$ and $K$ are provided in the supplement \cref{sec:Proof_of_metric_invariances}.

\section{Experiments}
\label{sec:Experiments}
Following our evaluation protocol (\cref{sec:experimental_setup}), we discuss benchmark results (\cref{subsec:Results}), further ablations (\cref{sec:LightSourceTypes}), and VLM-based evaluation comparisons (\cref{sec:comparison_to_vlm}). Additional details and experimental details are provided in the supplement (\cref{supp:additional_experiments,supp:details_to_experiments}).

\subsection{Experimental setup}
\label{sec:experimental_setup}

We identified six foundational models that understand our task well: Nano Banana Pro~\cite{google2025nanobananapro}, Nano Banana 2~\cite{google2026nanobanana2}, Qwen-Image-Edit (2511)~\cite{wu2025qwenimagetechnicalreport}, Flux 2 Dev~\cite{blackforestlabs2025flux2dev}, Flux 2 Max~\cite{flux2_max_2025}, and GPT Image 1.5~\cite{openai2025gptimage15}. We have exhaustively tested two other models, Bagel 7B~\cite{deng2025emergingbagel} and OmniGen2~\cite{wu2025omnigen2}, but they did not consistently understand the task, and hence they were only included in the supplement (\cref{supp:All_Models_All_Percentiles}). As mentioned above, none of the task-specific models~\cite{serrano2026synclight, magar2025lightlab, liang2026luxremix} provided any code or model weights.
For each model, we searched for the optimal prompt, which was then used consistently for every image. For example, for Nano Banana Pro the prompt is ``Identify the light bulb on the black pole. Turn it on with a bright white light.''  
To exclude cases where the model misinterpreted the task, we compute each metric using only the $80\%$ of images with the lowest error. Notably, visual inspection confirmed that all six models featured in the main benchmark successfully perform the requested tasks in over $80\%$ of the images. %
All models receive the input image at a resolution of 1024$\times$1536. Since all models produce different output resolutions, the outputs are rescaled to the smallest resolution among the tested models (832$\times$1248). 
This ensures that the scaling operations do not negatively impact our metrics.

\subsection{Benchmark and evaluation}
\label{subsec:Results}

\begin{figure}[t]
    \centering
    \includegraphics[width=\textwidth]{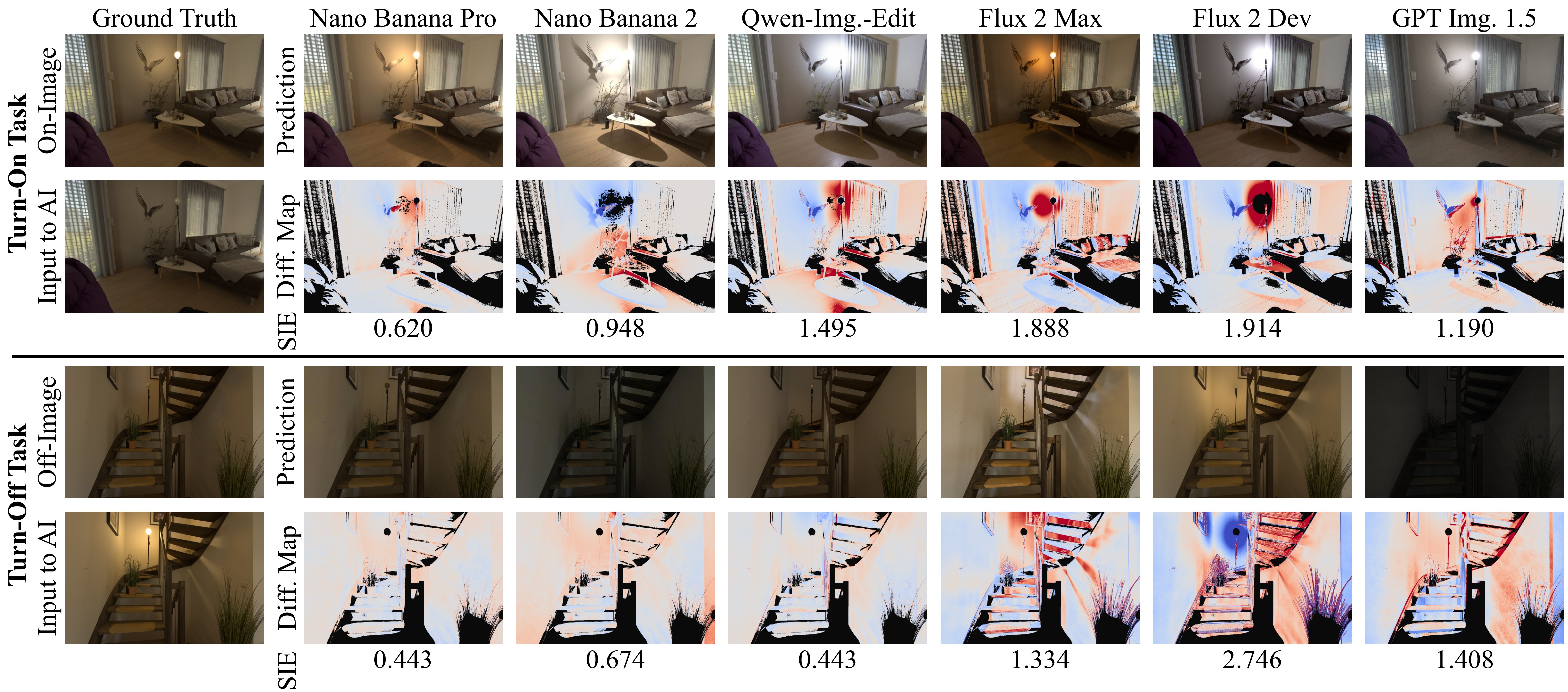}
    \caption{\textbf{Qualitative results (best viewed zoomed-in)} for the turn-on (top) and turn-off (bottom) task. The SIE error map is visualised as $\mathcal{S}(E^t_{R}) - \mathcal{S}(E^t_{G})$, where red indicates an excess of light relative to the rest of the image, blue indicates a deficit, and black pixels denote regions excluded due to clipping, low signal, or window labels. Note that the model is free to adapt the global exposure and white balance, as well as the colour and intensity of the added light bulb.
    Overall, the results differ considerably, while Nano Banana Pro is best in both tasks. The results show many interesting effects. For example, in the turn-on case, Nano Banana 2 identifies a real flying dove and casts an additional shadow of it on the wall. Flux 2 Max forgets to remove the shadows cast by the staircase in the turn-off case.
    \vspace{-0.4cm}
    }
    \label{fig:Qualitative}
\end{figure}

Quantitative results are presented in \cref{tab:main_benchmark} and qualitative results in \cref{fig:Qualitative}. The two commercial models, Nano Banana Pro and Nano Banana 2, clearly outperform the others. The best open-source model, Qwen-Image-Edit, is also clearly superior to the other open-source model, Flux 2 Dev.

\begin{table}[t]
\centering
\caption{\textbf{3DLP Benchmark results.} Quantitative evaluation of light transport for the turn-on and turn-off task. We report SIE (Standardised Intensity Error) and LFE (Low-Frequency Error) over the best 80\% of images. Best performance is marked in \textbf{bold}, and runner-up is \underline{underlined}.}
\label{tab:main_benchmark}
\resizebox{0.71\linewidth}{!}{
\begin{tabular}{lccccc}
\toprule
& \multicolumn{2}{c}{\textbf{Turning Light On}} & \multicolumn{2}{c}{\textbf{Turning Light Off}} & \\
\cmidrule(lr){2-3} \cmidrule(lr){4-5}
\textbf{Model}
& \textbf{SIE} $\downarrow$
& \textbf{LFE} $\downarrow$
& \textbf{SIE} $\downarrow$
& \textbf{LFE} $\downarrow$
& \textbf{Avg. Rank} $\downarrow$ \\
\midrule
Nano Banana Pro$^\ast$ & \textbf{0.813} & \textbf{1.641} & \textbf{0.769} & \textbf{1.651} & \textbf{1.0} \\
Nano Banana 2$^\ast$ & \underline{0.882} & \textbf{1.641} & \underline{0.871} & \underline{1.652} & \underline{1.8} \\
Qwen-Img.-Edit$^\dagger$ & 1.239 & 1.772 & 0.982 & 1.656 & 3.3 \\
Flux 2 Max$^\ast$ & 1.571 & 1.710 & 1.303 & 1.691 & 3.8 \\
Flux 2 Dev$^\dagger$ & 1.747 & 1.775 & 2.136 & 1.797 & 5.5 \\
GPT Img. 1.5$^\ast$ & 2.336 & 1.933 & 1.887 & 1.734 & 5.5 \\
\bottomrule
\end{tabular}
}
\vspace{0.05cm}

{\footnotesize $^\ast$Commercial model. $^\dagger$Open-source model.}
\end{table}

\begin{figure}[t]
    \centering
    \includegraphics[width=1\textwidth]{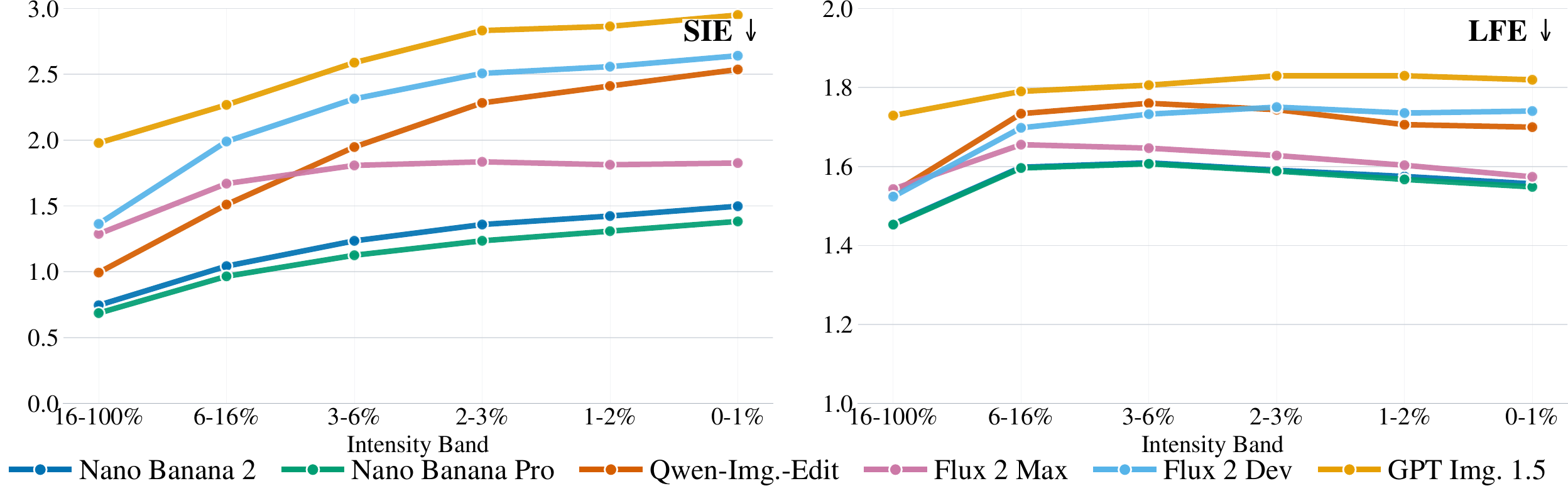}
    \caption{\textbf{Light intensity band analysis} for the turn-on task.
    Performance of various generative models evaluated across light intensity bands cast by the light probe. These bands serve as a proxy for the 3D distance to the light source, starting with the highest intensity band (closest to the source) on the left. Both metrics are calculated for the turn-on task. For both metrics, the lowest error is always within the highest intensity band, i.e. likely close to the light probe.
    \vspace{-0.4cm}
    }
    \label{fig:ablation_light_bands}
\end{figure}

\paragraph{Light intensity band analysis}
\label{subsec:IntensityBands}
For this analysis, we partition the difference image $I^{\textrm{on}}_{R}-I^{\textrm{off}}_{R}$ into six intensity bands. The difference image captures the isolated intensity of only the light probe. The partitioning is according to the inverse-square law of light attenuation. This means, for instance, that the first band has $16\%$ to $100\%$ of the maximum light intensity and the second band $6\%$ to $16\%$. If the scene was to have Lambertian surfaces only, then this would roughly capture the 3D distance of a pixel to the light probe. The results for the turn-on task are illustrated in \cref{fig:ablation_light_bands}. All scores have been calculated per band individually.
The SIE scores of all methods consistently increase with decreased intensity. This trend may be expected since, for instance, errors do accumulate for complex, long-distance light transport.    
For LFE the lowest value is also observed in the brightest intensity band. For lower intensity bands the error stays relatively constant. This indicates that the majority of models may possess a robust understanding of light attenuation and the Lambertian shading law.

\begin{figure}[t]
    \centering
    \includegraphics[width=1\textwidth]{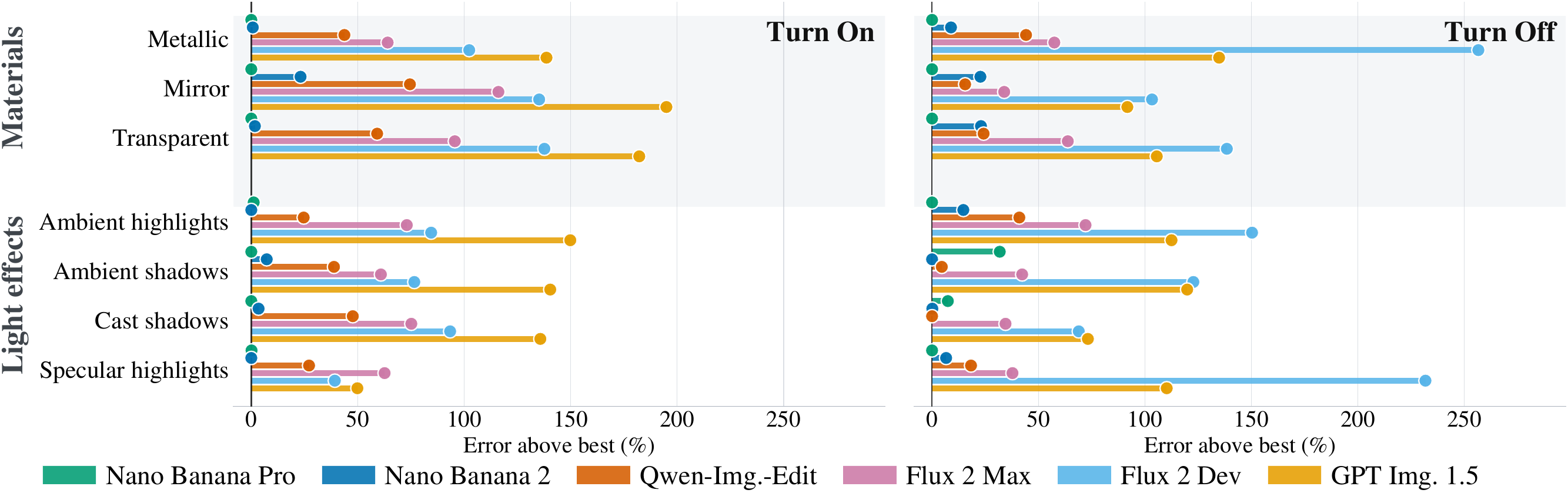}
    \caption{  \textbf{Material and light effect analysis.} For each annotated class the relative SIE scores are shown. This means that the best model always has a value of $0$, in most cases Nano Banana Pro, and we measure the relative increase (in percentage) of error for the remaining models. The spread of errors for some classes is higher than for other classes. 
    \vspace{-0.5cm}
    }
    \label{fig:ablation_annotations}
\end{figure}

\begin{figure}[t]
    \centering
    \includegraphics[width=\textwidth]{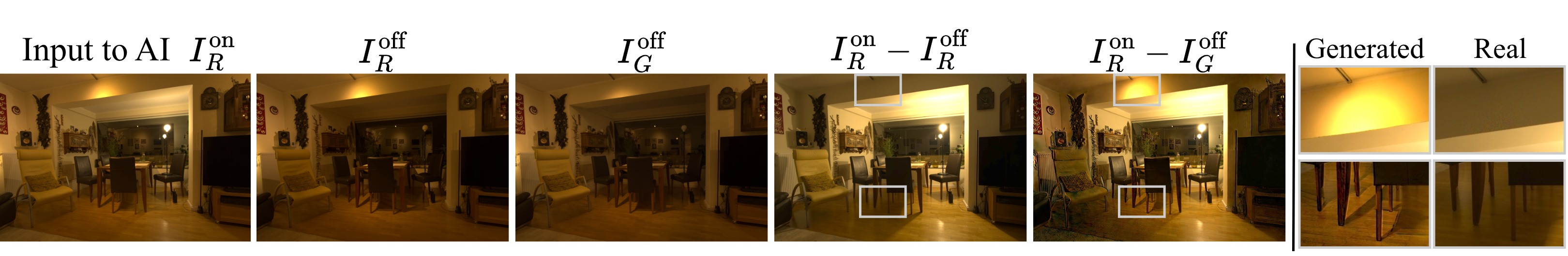}
    \caption{\textbf{Failure case for ambient highlights and shadows} for the turn-off task with the Qwen-Image-Edit model. Left are the real turned-on and turned-off images, as well as the respective difference images. The real difference image ($I^{\textrm{on}}_{R}-I^{\textrm{off}}_{R}$ ) only shows lighting effects caused by the light probe. In contrast, the difference image of the generative model ($I^{\textrm{on}}_{R}-I^{\textrm{off}}_{G}$) has additional incorrect effects.
    We see in the zoom-in that the model does not reason about the ambient highlight at the wall and the shadows cast by the chairs correctly, as they are still visible in the difference image.
    \vspace{-0.4cm}
    }
    \label{fig:ambientfailure}
\end{figure}

\begin{figure}[t]
    \centering
    \includegraphics[width=1\textwidth]{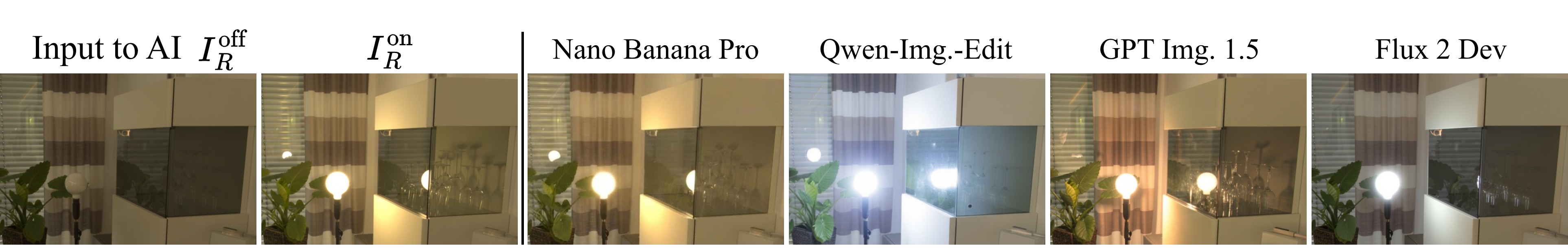}
    \caption{\textbf{Challenging transparent object.} Real images (left) alongside the results of four different models. None of the models is able to perform the task correctly for this highly challenging scene. Nonetheless the result of Nano Banana Pro is visually quite convincing. %
    \vspace{-0.5cm}
    }
    \label{fig:Transparent}
\end{figure}

\paragraph{Materials and light effects analysis}
\label{subsec:PerformanceOnLabels}

Given the annotations (see \cref{subsec:Dataset}), we evaluate individual models with respect to the different classes. \cref{fig:ablation_annotations} illustrates the respective SIE scores for the turn-on and turn-off task. Here we normalised the results so that the best model has always a value of $0$, in most cases Nano Banana Pro, and we measure the relative increase of error for the remaining models. For the turn-on task we observe that the spread in errors is largest for mirrors and smallest for specular highlights cast by the light probe. For the turn-off task, performance varies more among models than for the turn-on task. In particular, the Flux 2 Dev model underperforms for metallic surfaces and specular highlights.
\cref{fig:Teaser} shows an example where the model (Nano Banana 2) predicts a shadow and a highlight both cast by the light probe, although being visually pleasing they are not physically correct. 
The case of an ambient highlight and an ambient shadow which are both present in the off-image but not correctly handled by the model is illustrated in \cref{fig:ambientfailure}. Finally, \cref{fig:Transparent} shows a very challenging scene with a transparent vitrine that holds glasses. The real image contains higher-order bounces which, as expected, none of the models can handle perfectly. %

\subsection{Comparison with VLM scoring}
\label{sec:comparison_to_vlm}

Benchmarks like PICABench~\cite{pu2025picabench} use VLMs to evaluate lighting plausibility. However, while~\cite{pu2025picabench} ranks GPT Image 1.5 first in most categories, it ranks last in our benchmark (\cref{tab:main_benchmark}). To investigate VLMs for our benchmark, we used Gemini 3.1 Pro and GPT-5.5 in a binary A/B test to identify the more realistic lighting between real and generated on-images.
As shown in \cref{tab:vlm-percent-correct-all5}, the VLMs can disagree with each other and do not reflect our ranking. Scores below $50\%$ (e.g., Flux 2 Max) are especially notable, as they indicate that the VLM prefers generated artefacts over a real image. 
We conclude that while VLMs capture macroscopic plausibility and human-like perception well, they are not yet capable of precise, pixel-level physical evaluations of light transport.

\begin{table}[h]
\centering
\caption{\textbf{VLM A/B test.} Percentage of trials in which the VLM correctly preferred the real image $I_R^\textrm{on}$ over the generated $I_{G}^\textrm{on}$. Scores $< 50\%$ indicate a preference for generated images over physical ground truth, and $50\%$ represents chance. Columns are ordered by 3DLP benchmark ranks (left is best).}
\label{tab:vlm-percent-correct-all5}
\resizebox{\linewidth}{!}{
\begin{tabular}{lcccccc}
\toprule
\textbf{Evaluator} & \textbf{Nano Banana Pro} & \textbf{Nano Banana 2} & \textbf{Qwen-Img.-Edit} & \textbf{Flux 2 Max} & \textbf{Flux 2 Dev} & \textbf{GPT Img. 1.5} \\
\midrule
Gemini 3.1 Pro & 65.7\% & 76.0\% & 65.4\% & 24.6\% & 60.3\% & 65.9\% \\
GPT-5.5        & 48.6\% & 66.0\% & 73.3\% & 28.0\% & 66.6\% & 67.6\% \\
\bottomrule
\end{tabular}
}
\end{table}

\subsection{Light probe ablation}
\label{sec:LightSourceTypes}

To motivate our choice of a spherical lamp, we capture an additional 150 images for different lamp geometries (see details in \cref{supp:lamp_images}).  \cref{tab:ablation_light_type} presents the scores of the best-performing models across different light probes with varying emission patterns. As expected, for the turn-on task the SIE increases with complex geometries, since the model also has to predict the more complex emission patterns. As we want to measure light transport capabilities of different models, we chose the light probe with simplest geometry. 
Note, the turn-off task mitigates the emission pattern complexity since it is already visible in the input image, making the turn-off task extensible to other lamp geometries.

\begin{table*}[htbp]
    \centering
    \caption{\textbf{Light probe evaluation.} SIE and LFE scores for different light probe geometries. For the \textit{turn-on} task the SIE degrades with more complex lamp geometries due to emission pattern ambiguity. In contrast, for the \textit{turn-off} task the emission patterns are visible in the input image. Here, lower SIE values, compared to ours, can be explained by a smaller illuminated area that needs adjustment.} %
    \label{tab:ablation_light_type}
    \resizebox{\textwidth}{!}{
    \begin{tabular}{ll cccccccccccc}
        \toprule
        & & \multicolumn{2}{c}{\textbf{Spherical (Ours)}}
        & \multicolumn{2}{c}{\textbf{Shaded}}
        & \multicolumn{2}{c}{\textbf{Light Bar}}
        & \multicolumn{2}{c}{\textbf{Directional}}
        & \multicolumn{2}{c}{\textbf{Daylight}}
        & \multicolumn{2}{c}{\textbf{Basket}} \\
        \cmidrule(lr){3-4} \cmidrule(lr){5-6} \cmidrule(lr){7-8}
        \cmidrule(lr){9-10} \cmidrule(lr){11-12} \cmidrule(lr){13-14}
        \textbf{Task} & \textbf{Method}
        & \textbf{SIE} $\downarrow$ & \textbf{LFE} $\downarrow$
        & \textbf{SIE} $\downarrow$ & \textbf{LFE} $\downarrow$
        & \textbf{SIE} $\downarrow$ & \textbf{LFE} $\downarrow$
        & \textbf{SIE} $\downarrow$ & \textbf{LFE} $\downarrow$
        & \textbf{SIE} $\downarrow$ & \textbf{LFE} $\downarrow$
        & \textbf{SIE} $\downarrow$ & \textbf{LFE} $\downarrow$ \\
        \midrule
        \multirow{3}{*}{\rotatebox[origin=c]{90}{Turn-On}}
        & Nano Banana Pro
        & 1.210 & 2.202
        & 2.160 & 2.067
        & 3.607 & 2.652
        & 2.566 & 2.276
        & 3.481 & 2.636
        & 3.076 & 2.239 \\
        & Nano Banana 2
        & 1.547 & 2.163
        & 2.474 & 2.115
        & 3.401 & 2.680
        & 2.803 & 2.371
        & 3.747 & 2.669
        & 3.571 & 2.272 \\
        & Qwen-Img.-Edit
        & 1.767 & 2.662
        & 2.345 & 2.528
        & 4.387 & 3.198
        & 3.103 & 2.831
        & 5.190 & 3.295
        & 4.035 & 3.102 \\
        \midrule
        \multirow{3}{*}{\rotatebox[origin=c]{90}{Turn-Off}}
        & Nano Banana Pro
        & 1.058 & 1.873
        & 0.578 & 1.923
        & 0.761 & 2.008
        & 0.677 & 1.983
        & 0.956 & 1.907
        & 0.862 & 1.895 \\
        & Nano Banana 2
        & 1.173 & 1.871
        & 0.584 & 1.916
        & 0.681 & 1.992
        & 0.706 & 1.950
        & 1.070 & 1.949
        & 0.859 & 1.885 \\
        & Qwen-Img.-Edit
        & 1.314 & 1.955
        & 0.708 & 1.837
        & 1.044 & 1.992
        & 1.555 & 1.974
        & 1.313 & 1.939
        & 1.314 & 1.902 \\
        \bottomrule
    \end{tabular}
    }
\end{table*}

\section{Limitations and conclusions}
\label{sec:conclusions}
We introduce the 3DLP Benchmark and an accompanying high-fidelity dataset focused on indoor scenes. The limitation to indoor stems from our light probe's $\approx$2500-lumen output, which is easily overpowered by bright outdoor daylight and currently makes illumination differences unmeasurable. In response, we capture 100 outdoor HDR pairs at night (see \cref{supp:night_images}), but strong lens flare complicates the evaluation of light transport in these images. Additionally, the dataset predominantly contains rigid indoor surfaces and provides limited coverage of vegetation and materials with substantial subsurface scattering, such as human skin.

We benchmarked the light transport capabilities of six state-of-the-art image editing models and identified Nano Banana Pro as the overall top performer. 
Among the open-source models, we observe that Qwen-Image-Edit significantly outperforms Flux 2 Dev. These insights provide valuable guidance for researchers developing downstream applications that require a foundational backbone with a robust understanding of light transport. 
Given the scarcity of real-world relighting data, we envision that our dataset can help to improve and evaluate task-specific relighting models. Furthermore, our detailed annotations can help to foster research in specialised areas, such as the challenging problem of ambient shadow detection in indoor scenes.

\begin{ack}
We are grateful to Taha Erkoc and Jonathan Achim Hering for their generous time and essential support during dataset capture. We also thank the \href{https://www.botgart.cos.uni-heidelberg.de/en}{Botanical Garden of Heidelberg University}, \href{https://www.kampa.de}{KAMPA}, \href{https://www.okal.de}{OKAL}, \href{https://www.wolf-haus.de}{WOLF-HAUS}, \href{https://www.fingerhuthaus.de}{FINGERHUT}, \href{https://www.elkhaus.de}{ELK}, \href{https://www.rensch-haus.com}{RENSCH HAUS}, Eva Tilscher, Simon Pijahn, Juliane Lenz, Isabel Parnet, Nele Schikora, and Jonathan Achim Hering for providing access to locations used for data collection. Finally, we want to thank Angela Dai for the video voice-over. Tim Küchler is supported by the Konrad Zuse School of Excellence in Learning and Intelligent Systems (ELIZA) through the DAAD programme Konrad Zuse Schools of Excellence in Artificial Intelligence, sponsored by the Federal Ministry of Education and Research. This work was also supported by the ERC Consolidator Grant Gen3D (101171131).
\end{ack}

\clearpage
{
\small

\bibliographystyle{plainnat}
\bibliography{references}

}

\clearpage
\appendix
\section{Proofs and scores}
\label{supp:Proofs_and_Scores}

\subsection{Proof of metric invariance}
\label{sec:Proof_of_metric_invariances}
We show that our evaluation metrics, Standardised Intensity Error (SIE) and Low-Frequency Error (LFE), are invariant to the unknown global colour scale ($C > 0$) and exposure shift ($K$) introduced by the generative model. For readability, we consider each colour channel independently, denoting the scalar scale and shift for a given channel as $c, k \in \mathbb{R}$ with $c > 0$.

\textbf{Standardisation Invariance Lemma:} First, we prove that our robust standardisation operator $\mathcal{S}(x)$ is invariant to any affine transformation $x' = c x + k$ where $c > 0$.
Given the definition $\mathcal{S}(x) = \frac{x - \text{median}(x)}{\text{MAD}(x)}$, we determine the median and Median Absolute Deviation (MAD) of the transformed variable:
\begin{align}
    \text{median}(c x + k) &= c \cdot \text{median}(x) + k \\
    \text{MAD}(c x + k) &= \text{median}(|(c x + k) - (c \cdot \text{median}(x) + k)|) \nonumber \\
    &= \text{median}(c|x - \text{median}(x)|) \nonumber \\
    &= c \cdot \text{MAD}(x).
\end{align}
Substituting these into the standardisation operator gives:
\begin{equation}
    \mathcal{S}(c x + k) = \frac{c(x - \text{median}(x))}{c \cdot \text{MAD}(x)} = \mathcal{S}(x).
    \label{eq:standardisation_invariance}
\end{equation}
This basic property guarantees that any affine transformation applied to the pixel intensities is perfectly factored out.

\textbf{Standardised Intensity Error (SIE):} From our testing hypotheses, the ratio images for both the \textit{turn-on} and \textit{turn-off} tasks follow the exact same affine relationship per colour channel:
\begin{equation}
    E^t_{G} = c (E^t_R - 1) + k = c E^t_R + (k - c), \quad t \in \{\textrm{on}, \textrm{off}\}.
\end{equation}
This equation describes an affine transformation of the ground-truth ratio $E^t_R$, with an effective shift of $k \overset{\text{redefine}}{=} k - c$. Applying our standardisation lemma (Eq.~\ref{eq:standardisation_invariance}) directly to the generated ratio yields
\begin{equation}
    \mathcal{S}(E^t_{G}) = \mathcal{S}(c E^t_R + k) = \mathcal{S}(E^t_R).
\end{equation}
Therefore, the Standardised Intensity Error, $\textrm{SIE}_t = \text{MAE}(\mathcal{S}(E^t_R), \mathcal{S}(E^t_{G}))$, evaluates the structural distribution of light completely independently of the global scale $C$ and shift $K$.

\textbf{Low-Frequency Error (LFE):} This metric relies on the spatial gradients of the ratio images. As established, the generated ratio image is given by $E^t_{G} = c E^t_R + k$, where $k = k - c$. Applying the spatial gradient operator $\nabla$, the derivative of the constant $k$ vanishes:
\begin{equation}
    \nabla E^t_{G} = \nabla (c E^t_R + k) = c \nabla E^t_R + 0.
\end{equation}
Taking the gradient magnitude (knowing $c > 0$):
\begin{equation}
    M_{G}^t = |\nabla E^t_{G}| = |c \nabla E^t_R| = c |\nabla E^t_R| = c M_R^t.
\end{equation}
The shift $K$ has been completely eliminated, leaving a scaled gradient $c M_R^t$.

Before applying the standardisation, $\textrm{LFE}_t$ filters out high-frequency edges by only keeping pixels below the 80th percentile of gradient magnitudes ($\Omega_{80}$). Because multiplying by a strictly positive constant $c$ preserves the monotonic ordering of the gradient magnitudes, the 80th percentile threshold correctly scales by $c$ as well. Consequently, the identical set of pixels is selected for the mask $\Omega_{80}$ regardless of the value of $c$.

Finally, applying the standardisation lemma to the masked gradients gives
\begin{equation}
    \mathcal{S}(M_{G}^t) = \mathcal{S}(c M_R^t) = \mathcal{S}(M_R^t).
\end{equation}
This demonstrates that $\textrm{LFE}_t$ isolates and evaluates the physical realism of the light's decay completely invariant to both the exposure shift $K$ and the colour scale $C$.

\subsection{Effect of 8-bit quantisation uncertainty on metrics}
\label{supp:quantisation_uncertainty}
Generated images are available only as 8-bit sRGB files.  Hence, for an
observed channel value \(k \in \{0,\ldots,255\}\), the underlying normalised
sRGB intensity is only known up to the quantisation bin
\[
    q \sim \mathcal{U}\!\left(\frac{k-0.5}{255}, \frac{k+0.5}{255}\right),
\]
clipped to \([0,1]\).  To estimate how this uncertainty affects our reported
metrics, we use Monte Carlo propagation through the full metric pipeline.  For
each generated image, we repeatedly sample perturbed sRGB images from these
per-pixel quantisation bins, apply the same resizing and inverse sRGB
linearisation used in the benchmark, and recompute SIE and LFE with the
original evaluation mask held fixed. The per-sample quantisation uncertainty
is the empirical standard deviation over these Monte Carlo metric values.

For a model-level dataset mean, we propagate the independent per-sample
uncertainties \(\sigma_i\) as
\[
    \sigma_{\bar{m}}
    =
    \frac{1}{N}\sqrt{\sum_{i=1}^{N}\sigma_i^2},
\]
where \(N\) is the number of valid evaluated samples (Monte Carlo repetitions). We use this procedure with $N=64$ for both SIE and LFE and present the main benchmark table from \cref{tab:main_benchmark} with additional quantisation uncertainties in \cref{tab:quantization_uncertainty}. The uncertainty is negligible relative to the reported metric values.
\begin{table}[htbp]
\centering
\caption{\textbf{8-bit sRGB quantisation uncertainty.} Effect on the reported SIE and LFE metrics. We show the main benchmark's (\cref{tab:main_benchmark}) 80th-percentile SIE and LFE values with Monte Carlo propagated quantisation uncertainty. The uncertainties are negligible relative to the reported metric values.}
\label{tab:quantization_uncertainty}
\resizebox{\linewidth}{!}{
\begin{tabular}{lcccc}
\toprule
& \multicolumn{2}{c}{\textbf{Turning Light On}} & \multicolumn{2}{c}{\textbf{Turning Light Off}} \\
\cmidrule(lr){2-3} \cmidrule(lr){4-5}
\textbf{Model} & \textbf{SIE} $\downarrow$ & \textbf{LFE} $\downarrow$ & \textbf{SIE} $\downarrow$ & \textbf{LFE} $\downarrow$ \\
\midrule
Nano Banana Pro$^\ast$ & \textbf{0.813} $\pm 4.5{\times}10^{-6}$ & \textbf{1.641} $\pm 2.1{\times}10^{-5}$ & \textbf{0.769} $\pm 6.8{\times}10^{-6}$ & \textbf{1.651} $\pm 2.7{\times}10^{-5}$ \\
Nano Banana 2$^\ast$ & \underline{0.882} $\pm 4.1{\times}10^{-6}$ & \textbf{1.641} $\pm 2.1{\times}10^{-5}$ & \underline{0.871} $\pm 7.7{\times}10^{-6}$ & \underline{1.652} $\pm 2.7{\times}10^{-5}$ \\
Qwen-Img.-Edit$^\dagger$ & 1.239 $\pm 4.9{\times}10^{-6}$ & 1.772 $\pm 3.0{\times}10^{-5}$ & 0.982 $\pm 6.9{\times}10^{-6}$ & 1.656 $\pm 3.1{\times}10^{-5}$ \\
Flux 2 Max$^\ast$ & 1.571 $\pm 6.5{\times}10^{-6}$ & 1.710 $\pm 1.8{\times}10^{-5}$ & 1.303 $\pm 5.1{\times}10^{-6}$ & 1.691 $\pm 1.9{\times}10^{-5}$ \\
Flux 2 Dev$^\dagger$ & 1.747 $\pm 1.1{\times}10^{-5}$ & 1.775 $\pm 2.9{\times}10^{-5}$ & 2.136 $\pm 2.4{\times}10^{-5}$ & 1.797 $\pm 3.6{\times}10^{-5}$ \\
GPT Img. 1.5$^\ast$ & 2.336 $\pm 6.7{\times}10^{-6}$ & 1.933 $\pm 2.5{\times}10^{-5}$ & 1.887 $\pm 7.2{\times}10^{-6}$ & 1.734 $\pm 2.0{\times}10^{-5}$ \\
\bottomrule
\end{tabular}
}
\vspace{0.05cm}

{\footnotesize $^\ast$Commercial model. $^\dagger$Open-source model.}
\end{table}

\subsection{Details on evaluation masking}
\label{supp:evaluation-masks}

All metrics are evaluated on a validity mask $\Omega$ that keeps only pixels where the ground-truth light transport is observable. For the LFE metric, we define an additional validation mask, $\Omega_{80}$, in \cref{sec:metrics} to exclude regions with large gradient magnitudes. While the mask $\Omega$ described in this section applies to both SIE and LFE, the LFE evaluation specifically utilises the intersection of $\Omega$ and $\Omega_{80}$.

Let
\[
L_R=\max(I^{\textrm{on}}_R-I^{\textrm{off}}_R,0)
\]
be the ground-truth light contribution in linear RGB.  We first exclude saturated pixels, since clipping removes the radiometric information needed to compare light ratios.  Although SIE and LFE are computed in linear space, clipping is detected after mapping the real ground-truth image to its sRGB representation. Pixels are removed if any colour channel is clipped in either the ground-truth image or the generated edit image.

Secondly, we also exclude very low-signal pixels, where 8-bit quantisation and camera registration uncertainty dominate the light change.  The threshold is applied to a smoothed light map (Gaussian kernel with $\sigma=0.02\cdot\max(\textrm{height, \textrm{width}})$), so nearby illumination keeps high-frequency cast shadows valid while far-away dark regions are removed. We keep pixels whose smoothed mean light contribution is at least $1\%$ of its 99th percentile.

Finally, we remove pixels labelled as windows from all reported metrics.  Windows can show outdoor content that moves between captures, making the ground truth invalid for indoor relighting.  They are also highly sensitive to external illumination changes, such as passing clouds, which would negatively impact the measured light transport error.

\subsection{Extension to non-invariant global exposure and white balance}
\label{supp:non_invariant_exposure_white_balance}

In applications such as shadow detection, one may want to penalise global exposure and white-balance changes rather than be invariant to them. Our framework supports this by setting $K=\mathbf{1}$ in \cref{eq:AIassumption_on}, yielding $I^{\textrm{on}}_G=C\odot L_R+I^{\textrm{off}}_R$, and replacing the standardisation operator with $\mathcal{S}_1(X)=(X-1)/\text{MAD}(X)$. This makes SIE invariant to lamp intensity and colour while penalising global exposure and white-balance changes. The same adaptation is not possible for LFE, since its spatial derivative removes the spatially constant global term. \cref{tab:non_invariant_sie} reports SIE recomputed with this formulation. Nano Banana Pro remains best on both tasks, while the gap between the three leading models and the Flux/GPT models widens. This is consistent with our observation that the latter frequently change global exposure and/or white balance.

\begin{table}[htbp]
\centering
\caption{\textbf{SIE penalising global exposure and white-balance changes.} Results using $\mathcal{S}_1$ for the \textit{turn-on} and \textit{turn-off} tasks. Best performance is marked in \textbf{bold}, and runner-up is \underline{underlined}.}
\label{tab:non_invariant_sie}
\small
\begin{tabular}{lcc}
\toprule
\textbf{Model} & \textbf{Turn-On SIE} $\downarrow$ & \textbf{Turn-Off SIE} $\downarrow$ \\
\midrule
Nano Banana Pro & \textbf{1.601} & \textbf{4.190} \\
Nano Banana 2 & \underline{1.751} & 5.813 \\
Qwen-Img.-Edit & 2.179 & \underline{4.399} \\
Flux 2 Max & 3.697 & 7.814 \\
Flux 2 Dev & 3.923 & 7.457 \\
GPT Img. 1.5 & 4.557 & 16.496 \\
\bottomrule
\end{tabular}
\end{table}

\section{Additional experiments}
\label{supp:additional_experiments}

\subsection{Comparison of SIE and LFE to PSNR, SSIM and LPIPS}
\label{supp:comparison_SIE_LFE_vs_PSNR_SSIM_LPIPS}

\begin{figure}[htbp]
    \centering
    \includegraphics[width=\textwidth]{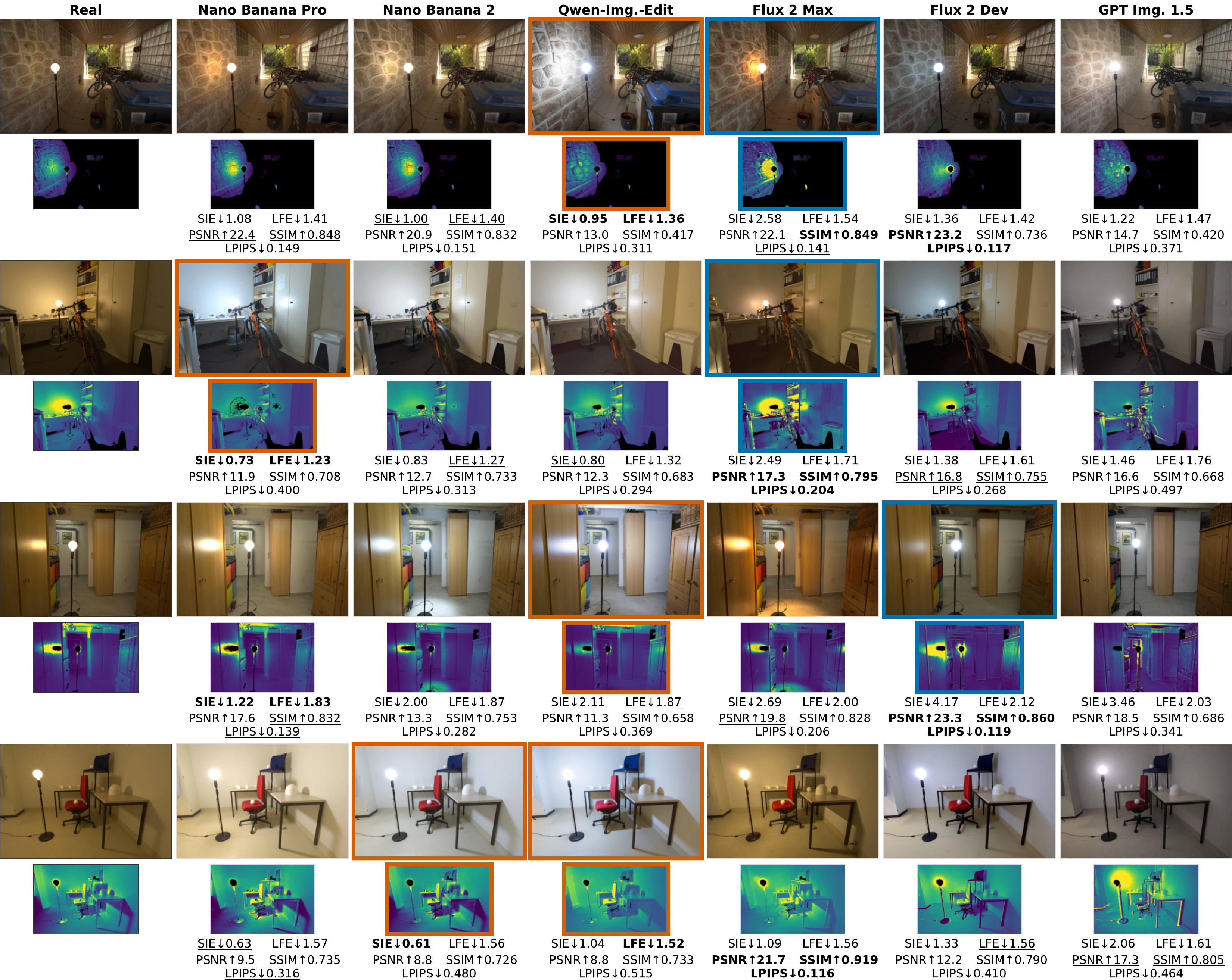}
    \caption{\textbf{Qualitative comparison of SIE and LFE versus PSNR, SSIM and LPIPS.} Major rows present the real on-image $I_R^\textrm{on}$ and the generated on-images $I_{G}^\textrm{on}$. The rows below show the intensity ratios $E_R^\textrm{on}$ and $E_{G}^\textrm{on}$. For each row the best value per metric is given \textbf{bold} and the second best \underline{underlined}. \textcolor{orange}{Orange} and \textcolor{blue}{blue} boxes indicate samples in which SIE/LFE and PSNR/SSIM/LPIPS are in exceptionally high disagreement, where orange indicates good SIE/LFE and bad PSNR/SSIM/LPIPS while blue indicates the opposite. Inspection of the ratio images shows that the different metrics pick up on very different aspects of light. \textbf{Best viewed zoomed in.}}
    \label{fig:sie_lfe_ssim_psnr_lpips_comparison}
\end{figure}

Prior works in intrinsic image decomposition and relighting~\cite{zeng2024rgb, kocsis2025iif} as well as light editing~\cite{magar2025lightlab, serrano2026synclight, liang2026luxremix} primarily rely on PSNR, SSIM~\cite{wang2004imagessim}, and LPIPS~\cite{zhang2018unreasonablelpips} for evaluation. \cref{tab:relighting_metrics_white_balance_full_reference} highlights two major advantages of our proposed metrics (SIE and LFE). First, standard metrics are highly sensitive to white balance shifts, which can arbitrarily alter model rankings. In contrast, our metrics are physically inspired and are invariant to changes in global exposure, white balance, and the intensity or colour of the activated light probe.
Second, traditional metrics fail to penalise complete task failures, such as failing to turn a light on or off (``Do Nothing'' model), frequently ranking these failed models ahead of successful ones. Conversely, our metrics strictly penalise such failures. Notably, PSNR, SSIM, and LPIPS show no correlation with SIE and LFE on the turn-on task, indicating they capture fundamentally different properties. Qualitative comparisons are provided in \cref{fig:sie_lfe_ssim_psnr_lpips_comparison}.

\begin{table*}[htbp]
\centering
\caption{\textbf{Comparison of SIE and LFE to PSNR, SSIM and LPIPS.} We report 80th-percentile SIE and LFE, together with PSNR, SSIM, and LPIPS, calculated on the default real HDR pairs. The \textcolor{red}{red} and \textcolor{blue}{blue} deltas below the scores show the changes in metrics when the real HDR pairs are shifted towards a \textcolor{red}{warmer} or \textcolor{blue}{colder} white balance. It is noticeable that our metrics (SIE/LFE) are invariant to such transformations and thus do not change. In contrast, PSNR, SSIM, and LPIPS do change noticeably. For the ``Do Nothing'' model we simulate a model failing to turn on/off the lamp, by calculating the metrics against $I_R^\textrm{off}$ for the \textit{turn-on} task and against $I_R^\textrm{on}$ for the \textit{turn-off} task. It can be seen that such a model is frequently ranked among the best for PSNR, SSIM and LPIPS, while our metrics (SIE/LFE) heavily punish such behaviour. Best reference performance is marked in \textbf{bold}, while the second-best is \underline{underlined}.}
\label{tab:relighting_metrics_white_balance_full_reference}
\resizebox{\linewidth}{!}{
\begin{tabular}{lcccccccccc}
\toprule
& \multicolumn{5}{c}{\textbf{Turning Light On}} & \multicolumn{5}{c}{\textbf{Turning Light Off}} \\
\cmidrule(lr){2-6} \cmidrule(lr){7-11}
\textbf{Model} & \textbf{SIE} $\downarrow$ & \textbf{LFE} $\downarrow$ & \textbf{PSNR} $\uparrow$ & \textbf{SSIM} $\uparrow$ & \textbf{LPIPS} $\downarrow$ & \textbf{SIE} $\downarrow$ & \textbf{LFE} $\downarrow$ & \textbf{PSNR} $\uparrow$ & \textbf{SSIM} $\uparrow$ & \textbf{LPIPS} $\downarrow$ \\
\midrule
Nano Banana Pro$^\ast$ & \makecell{\textbf{0.813}\\[-0.2ex]{\scriptsize\textcolor{red}{-0.000}/\textcolor{blue}{+0.000}}} & \makecell{\textbf{1.641}\\[-0.2ex]{\scriptsize\textcolor{red}{-0.000}/\textcolor{blue}{+0.000}}} & \makecell{20.14\\[-0.2ex]{\scriptsize\textcolor{red}{-1.66}/\textcolor{blue}{-0.68}}} & \makecell{0.877\\[-0.2ex]{\scriptsize\textcolor{red}{-0.067}/\textcolor{blue}{+0.006}}} & \makecell{0.126\\[-0.2ex]{\scriptsize\textcolor{red}{+0.079}/\textcolor{blue}{+0.027}}} & \makecell{\textbf{0.769}\\[-0.2ex]{\scriptsize\textcolor{red}{-0.000}/\textcolor{blue}{+0.000}}} & \makecell{\textbf{1.651}\\[-0.2ex]{\scriptsize\textcolor{red}{-0.000}/\textcolor{blue}{-0.000}}} & \makecell{\textbf{30.12}\\[-0.2ex]{\scriptsize\textcolor{red}{-6.20}/\textcolor{blue}{-3.42}}} & \makecell{\textbf{0.937}\\[-0.2ex]{\scriptsize\textcolor{red}{-0.063}/\textcolor{blue}{-0.007}}} & \makecell{0.086\\[-0.2ex]{\scriptsize\textcolor{red}{+0.082}/\textcolor{blue}{+0.045}}} \\
Nano Banana 2$^\ast$ & \makecell{\underline{0.882}\\[-0.2ex]{\scriptsize\textcolor{red}{-0.000}/\textcolor{blue}{+0.000}}} & \makecell{\underline{1.641}\\[-0.2ex]{\scriptsize\textcolor{red}{-0.000}/\textcolor{blue}{+0.000}}} & \makecell{17.58\\[-0.2ex]{\scriptsize\textcolor{red}{-1.63}/\textcolor{blue}{+0.11}}} & \makecell{0.840\\[-0.2ex]{\scriptsize\textcolor{red}{-0.070}/\textcolor{blue}{+0.014}}} & \makecell{0.174\\[-0.2ex]{\scriptsize\textcolor{red}{+0.114}/\textcolor{blue}{-0.012}}} & \makecell{\underline{0.871}\\[-0.2ex]{\scriptsize\textcolor{red}{-0.000}/\textcolor{blue}{+0.000}}} & \makecell{\underline{1.652}\\[-0.2ex]{\scriptsize\textcolor{red}{-0.000}/\textcolor{blue}{+0.000}}} & \makecell{27.27\\[-0.2ex]{\scriptsize\textcolor{red}{-5.01}/\textcolor{blue}{-1.55}}} & \makecell{0.919\\[-0.2ex]{\scriptsize\textcolor{red}{-0.070}/\textcolor{blue}{+0.000}}} & \makecell{0.108\\[-0.2ex]{\scriptsize\textcolor{red}{+0.093}/\textcolor{blue}{+0.026}}} \\
Qwen-Img.-Edit$^\dagger$ & \makecell{1.239\\[-0.2ex]{\scriptsize\textcolor{red}{-0.001}/\textcolor{blue}{+0.000}}} & \makecell{1.772\\[-0.2ex]{\scriptsize\textcolor{red}{-0.000}/\textcolor{blue}{+0.000}}} & \makecell{14.03\\[-0.2ex]{\scriptsize\textcolor{red}{-1.12}/\textcolor{blue}{+0.35}}} & \makecell{0.732\\[-0.2ex]{\scriptsize\textcolor{red}{-0.061}/\textcolor{blue}{+0.016}}} & \makecell{0.280\\[-0.2ex]{\scriptsize\textcolor{red}{+0.114}/\textcolor{blue}{-0.034}}} & \makecell{0.982\\[-0.2ex]{\scriptsize\textcolor{red}{-0.000}/\textcolor{blue}{+0.000}}} & \makecell{1.656\\[-0.2ex]{\scriptsize\textcolor{red}{-0.000}/\textcolor{blue}{-0.000}}} & \makecell{\underline{27.42}\\[-0.2ex]{\scriptsize\textcolor{red}{-4.33}/\textcolor{blue}{-2.04}}} & \makecell{0.907\\[-0.2ex]{\scriptsize\textcolor{red}{-0.056}/\textcolor{blue}{-0.009}}} & \makecell{\textbf{0.075}\\[-0.2ex]{\scriptsize\textcolor{red}{+0.081}/\textcolor{blue}{+0.043}}} \\
Flux 2 Max$^\ast$ & \makecell{1.571\\[-0.2ex]{\scriptsize\textcolor{red}{-0.000}/\textcolor{blue}{+0.000}}} & \makecell{1.710\\[-0.2ex]{\scriptsize\textcolor{red}{-0.000}/\textcolor{blue}{+0.000}}} & \makecell{\underline{21.96}\\[-0.2ex]{\scriptsize\textcolor{red}{-2.31}/\textcolor{blue}{-0.92}}} & \makecell{0.857\\[-0.2ex]{\scriptsize\textcolor{red}{-0.043}/\textcolor{blue}{-0.007}}} & \makecell{0.159\\[-0.2ex]{\scriptsize\textcolor{red}{+0.066}/\textcolor{blue}{+0.027}}} & \makecell{1.303\\[-0.2ex]{\scriptsize\textcolor{red}{-0.000}/\textcolor{blue}{+0.000}}} & \makecell{1.691\\[-0.2ex]{\scriptsize\textcolor{red}{-0.000}/\textcolor{blue}{-0.000}}} & \makecell{19.80\\[-0.2ex]{\scriptsize\textcolor{red}{-1.10}/\textcolor{blue}{-0.58}}} & \makecell{0.818\\[-0.2ex]{\scriptsize\textcolor{red}{-0.060}/\textcolor{blue}{+0.004}}} & \makecell{0.161\\[-0.2ex]{\scriptsize\textcolor{red}{+0.082}/\textcolor{blue}{+0.033}}} \\
Flux 2 Dev$^\dagger$ & \makecell{1.747\\[-0.2ex]{\scriptsize\textcolor{red}{-0.001}/\textcolor{blue}{+0.000}}} & \makecell{1.775\\[-0.2ex]{\scriptsize\textcolor{red}{-0.000}/\textcolor{blue}{+0.000}}} & \makecell{20.26\\[-0.2ex]{\scriptsize\textcolor{red}{-3.02}/\textcolor{blue}{+0.57}}} & \makecell{0.821\\[-0.2ex]{\scriptsize\textcolor{red}{-0.063}/\textcolor{blue}{+0.006}}} & \makecell{0.154\\[-0.2ex]{\scriptsize\textcolor{red}{+0.118}/\textcolor{blue}{-0.015}}} & \makecell{2.136\\[-0.2ex]{\scriptsize\textcolor{red}{+0.000}/\textcolor{blue}{-0.000}}} & \makecell{1.797\\[-0.2ex]{\scriptsize\textcolor{red}{-0.000}/\textcolor{blue}{+0.000}}} & \makecell{21.99\\[-0.2ex]{\scriptsize\textcolor{red}{-1.40}/\textcolor{blue}{-1.10}}} & \makecell{0.843\\[-0.2ex]{\scriptsize\textcolor{red}{-0.060}/\textcolor{blue}{-0.003}}} & \makecell{0.088\\[-0.2ex]{\scriptsize\textcolor{red}{+0.065}/\textcolor{blue}{+0.055}}} \\
GPT Img. 1.5$^\ast$ & \makecell{2.336\\[-0.2ex]{\scriptsize\textcolor{red}{-0.001}/\textcolor{blue}{+0.000}}} & \makecell{1.933\\[-0.2ex]{\scriptsize\textcolor{red}{-0.000}/\textcolor{blue}{+0.000}}} & \makecell{17.95\\[-0.2ex]{\scriptsize\textcolor{red}{-2.31}/\textcolor{blue}{+0.61}}} & \makecell{0.697\\[-0.2ex]{\scriptsize\textcolor{red}{-0.059}/\textcolor{blue}{+0.011}}} & \makecell{0.352\\[-0.2ex]{\scriptsize\textcolor{red}{+0.123}/\textcolor{blue}{-0.039}}} & \makecell{1.887\\[-0.2ex]{\scriptsize\textcolor{red}{-0.000}/\textcolor{blue}{+0.000}}} & \makecell{1.734\\[-0.2ex]{\scriptsize\textcolor{red}{-0.000}/\textcolor{blue}{+0.000}}} & \makecell{18.88\\[-0.2ex]{\scriptsize\textcolor{red}{-1.61}/\textcolor{blue}{+0.24}}} & \makecell{0.680\\[-0.2ex]{\scriptsize\textcolor{red}{-0.036}/\textcolor{blue}{+0.001}}} & \makecell{0.377\\[-0.2ex]{\scriptsize\textcolor{red}{+0.100}/\textcolor{blue}{-0.023}}} \\
Bagel 7B MoT$^\dagger$ & \makecell{2.655\\[-0.2ex]{\scriptsize\textcolor{red}{-0.000}/\textcolor{blue}{+0.000}}} & \makecell{1.851\\[-0.2ex]{\scriptsize\textcolor{red}{-0.000}/\textcolor{blue}{+0.000}}} & \makecell{\textbf{23.80}\\[-0.2ex]{\scriptsize\textcolor{red}{-4.39}/\textcolor{blue}{-0.45}}} & \makecell{\textbf{0.923}\\[-0.2ex]{\scriptsize\textcolor{red}{-0.066}/\textcolor{blue}{+0.000}}} & \makecell{\underline{0.087}\\[-0.2ex]{\scriptsize\textcolor{red}{+0.105}/\textcolor{blue}{+0.012}}} & \makecell{2.322\\[-0.2ex]{\scriptsize\textcolor{red}{+0.000}/\textcolor{blue}{+0.000}}} & \makecell{1.706\\[-0.2ex]{\scriptsize\textcolor{red}{-0.000}/\textcolor{blue}{+0.000}}} & \makecell{21.68\\[-0.2ex]{\scriptsize\textcolor{red}{-1.32}/\textcolor{blue}{-1.24}}} & \makecell{0.896\\[-0.2ex]{\scriptsize\textcolor{red}{-0.063}/\textcolor{blue}{-0.004}}} & \makecell{0.098\\[-0.2ex]{\scriptsize\textcolor{red}{+0.058}/\textcolor{blue}{+0.056}}} \\
OmniGen2$^\dagger$ & \makecell{3.159\\[-0.2ex]{\scriptsize\textcolor{red}{-0.001}/\textcolor{blue}{+0.000}}} & \makecell{1.959\\[-0.2ex]{\scriptsize\textcolor{red}{-0.000}/\textcolor{blue}{+0.000}}} & \makecell{20.71\\[-0.2ex]{\scriptsize\textcolor{red}{-1.86}/\textcolor{blue}{-0.74}}} & \makecell{0.823\\[-0.2ex]{\scriptsize\textcolor{red}{-0.022}/\textcolor{blue}{-0.011}}} & \makecell{0.143\\[-0.2ex]{\scriptsize\textcolor{red}{+0.067}/\textcolor{blue}{+0.027}}} & \makecell{2.700\\[-0.2ex]{\scriptsize\textcolor{red}{-0.000}/\textcolor{blue}{+0.000}}} & \makecell{1.771\\[-0.2ex]{\scriptsize\textcolor{red}{-0.000}/\textcolor{blue}{-0.000}}} & \makecell{20.48\\[-0.2ex]{\scriptsize\textcolor{red}{-0.75}/\textcolor{blue}{-1.13}}} & \makecell{0.854\\[-0.2ex]{\scriptsize\textcolor{red}{-0.035}/\textcolor{blue}{-0.014}}} & \makecell{0.133\\[-0.2ex]{\scriptsize\textcolor{red}{+0.044}/\textcolor{blue}{+0.057}}} \\
\midrule
Do Nothing & \makecell{39.901} & \makecell{5.513} & \makecell{21.19} & \makecell{\underline{0.922}} & \makecell{\textbf{0.077}} & \makecell{59.733} & \makecell{5.939} & \makecell{21.19} & \makecell{\underline{0.922}} & \makecell{\underline{0.077}} \\

\bottomrule
\end{tabular}
}
\vspace{0.05cm}

{\footnotesize $^\ast$Commercial model. $^\dagger$Open-source model}
\end{table*}

\begin{figure*}[ht!]
    \centering
        \begin{subfigure}{.14\linewidth}
        \centering
        \includegraphics[width=\linewidth, angle=-90]{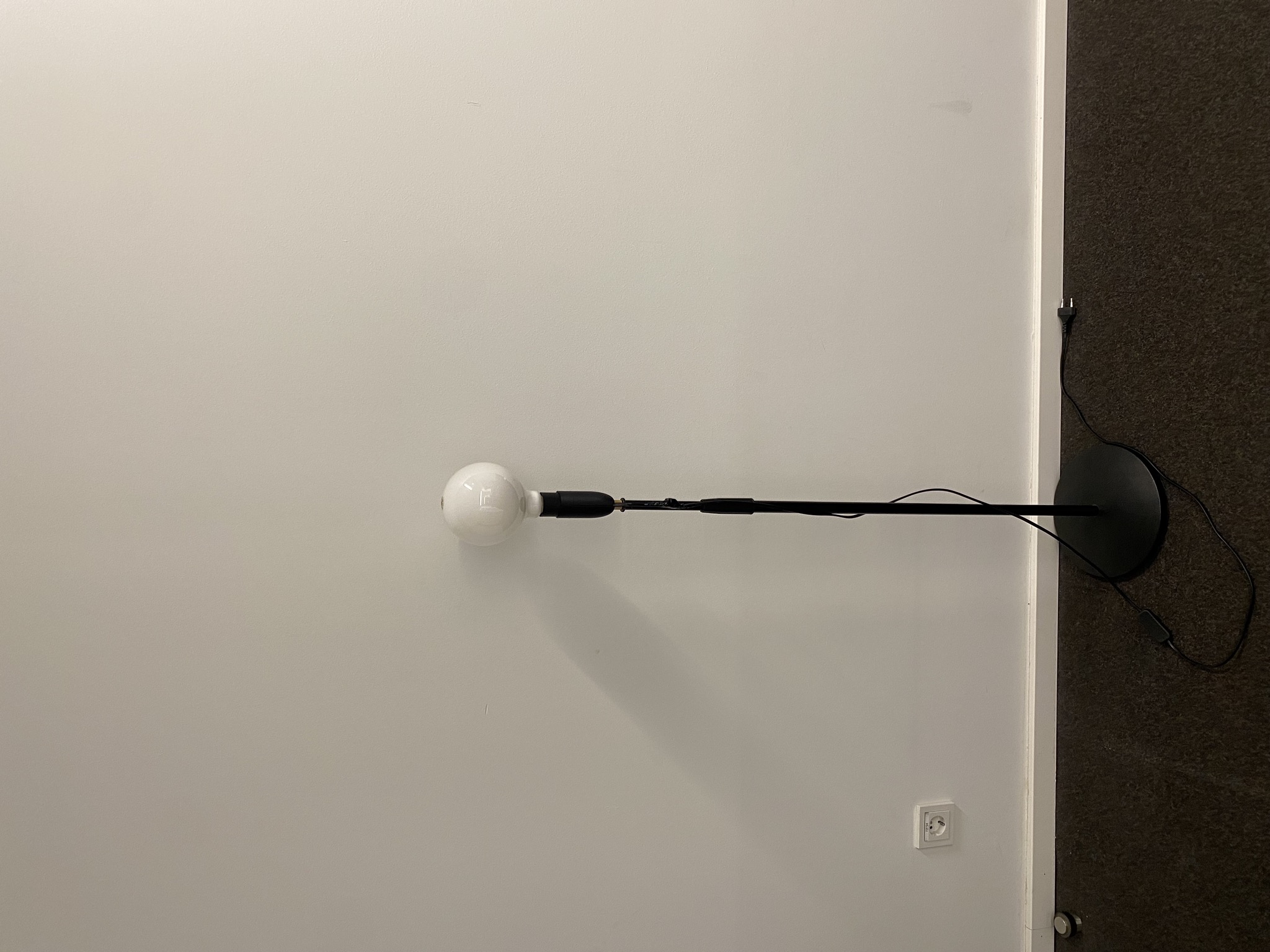}
        \caption*{Spherical}
    \end{subfigure}\hfill
    \begin{subfigure}{.14\linewidth}
        \centering
        \includegraphics[width=\linewidth, angle=-90]{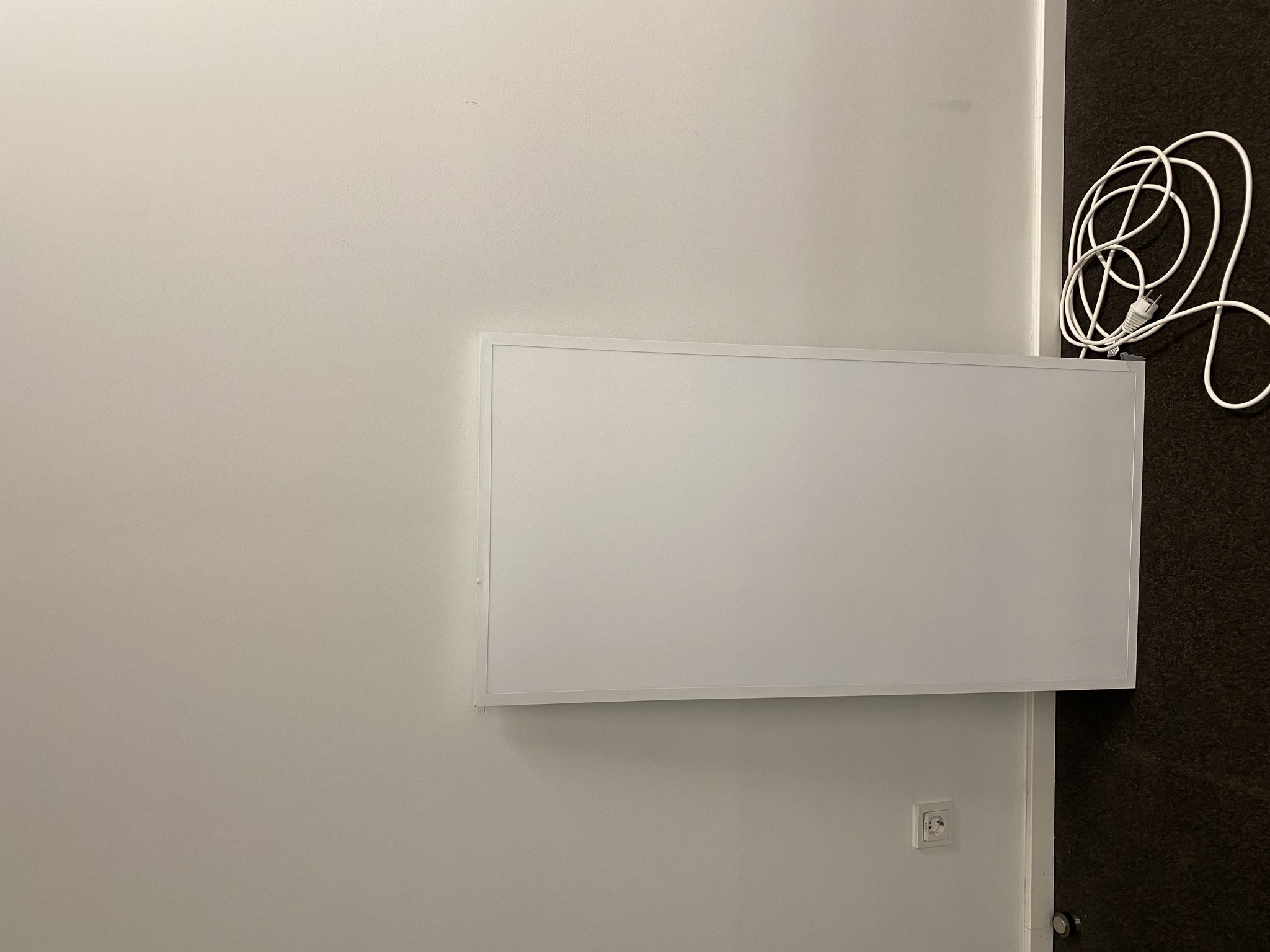}
        \caption*{Area}
    \end{subfigure}\hfill
    \begin{subfigure}{.14\linewidth}
        \centering
        \includegraphics[width=\linewidth, angle=-90]{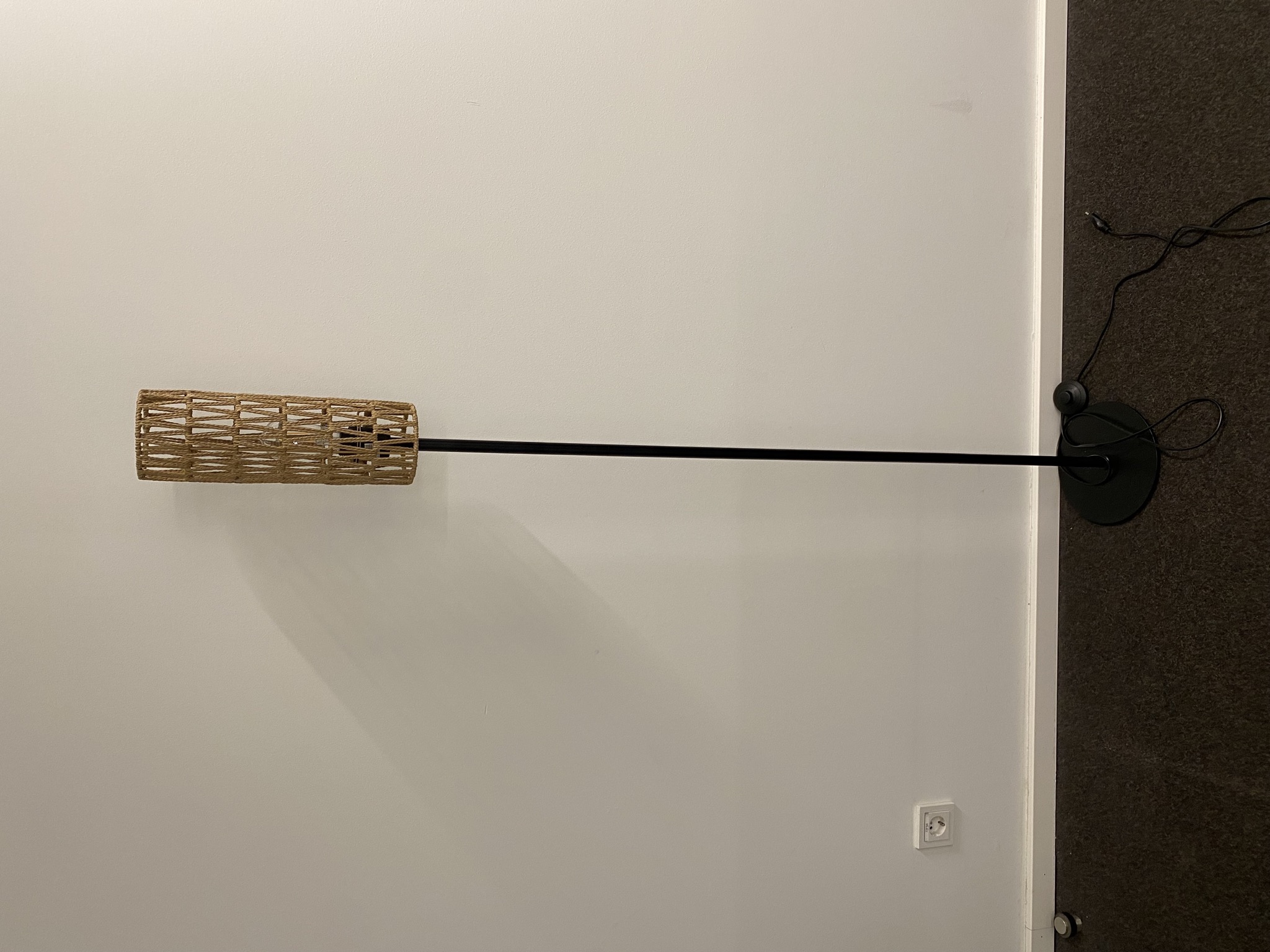}
        \caption*{Basket}
    \end{subfigure}\hfill
    \begin{subfigure}{.14\linewidth}
        \centering
        \includegraphics[width=\linewidth, angle=-90]{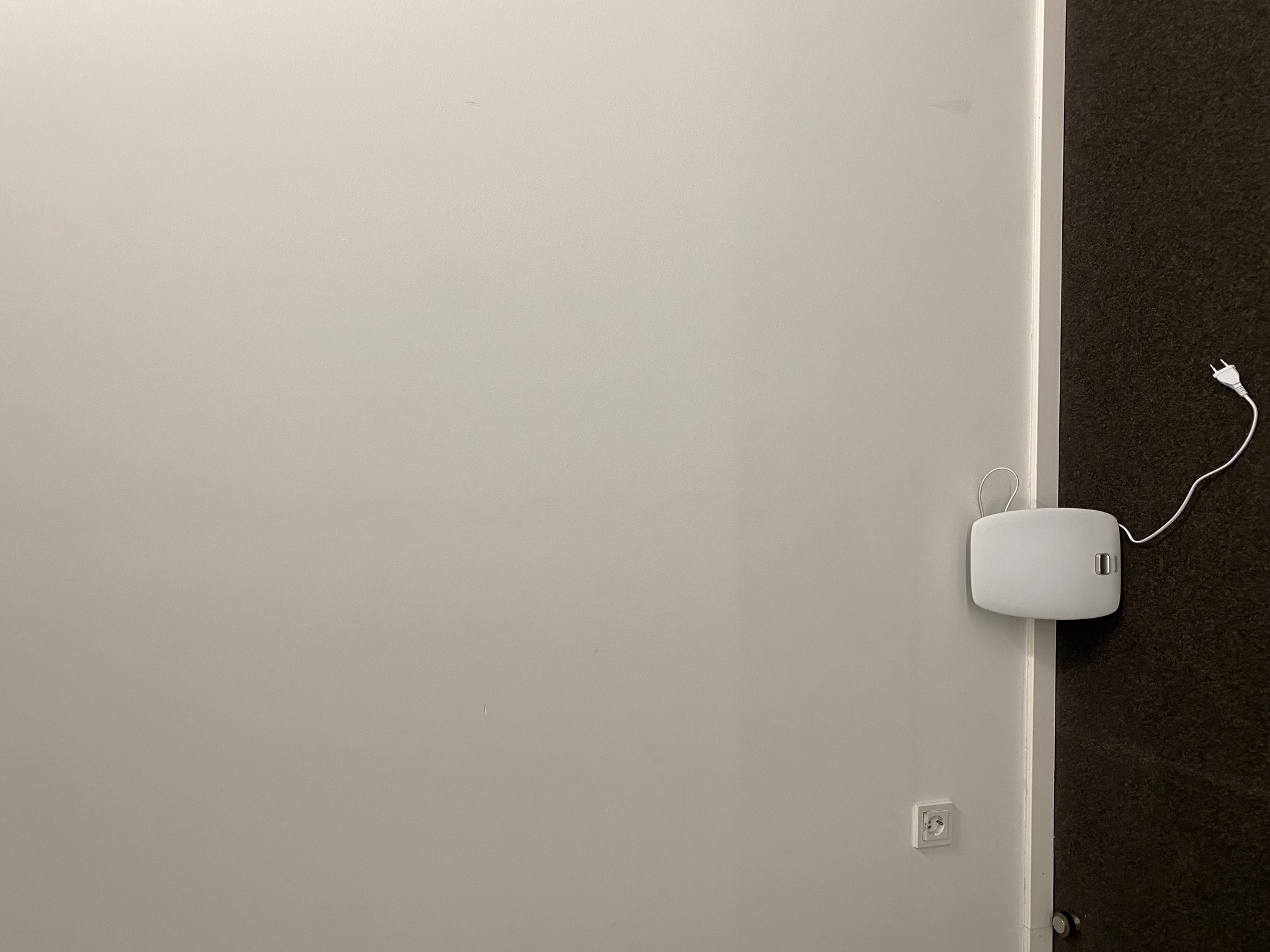}
        \caption*{Daylight}
    \end{subfigure}\hfill
    \begin{subfigure}{.14\linewidth}
        \centering
        \includegraphics[width=\linewidth, angle=-90]{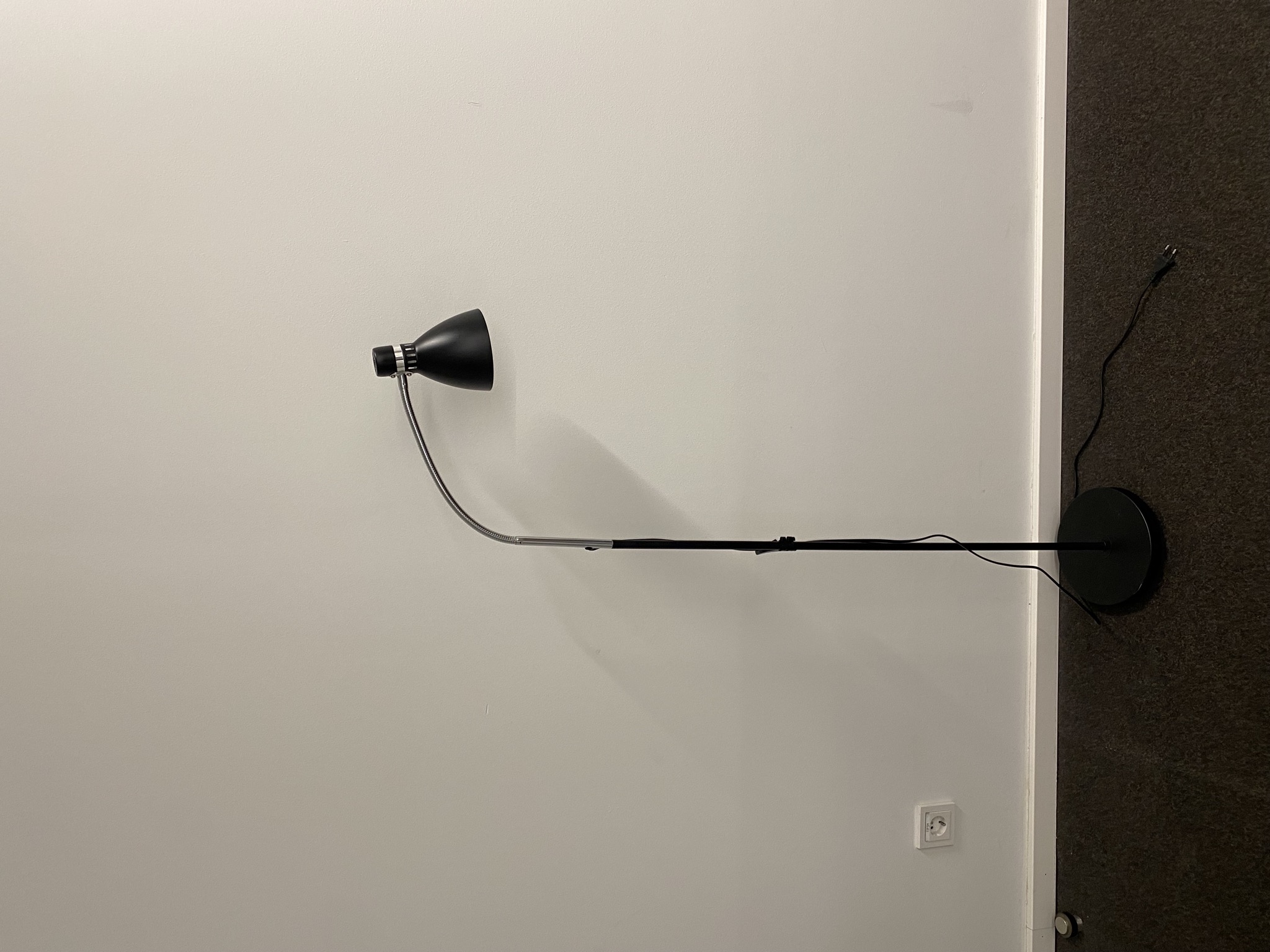}
        \caption*{Directional}
    \end{subfigure}\hfill
    \begin{subfigure}{.14\linewidth}
        \centering
        \includegraphics[width=\linewidth, angle=-90]{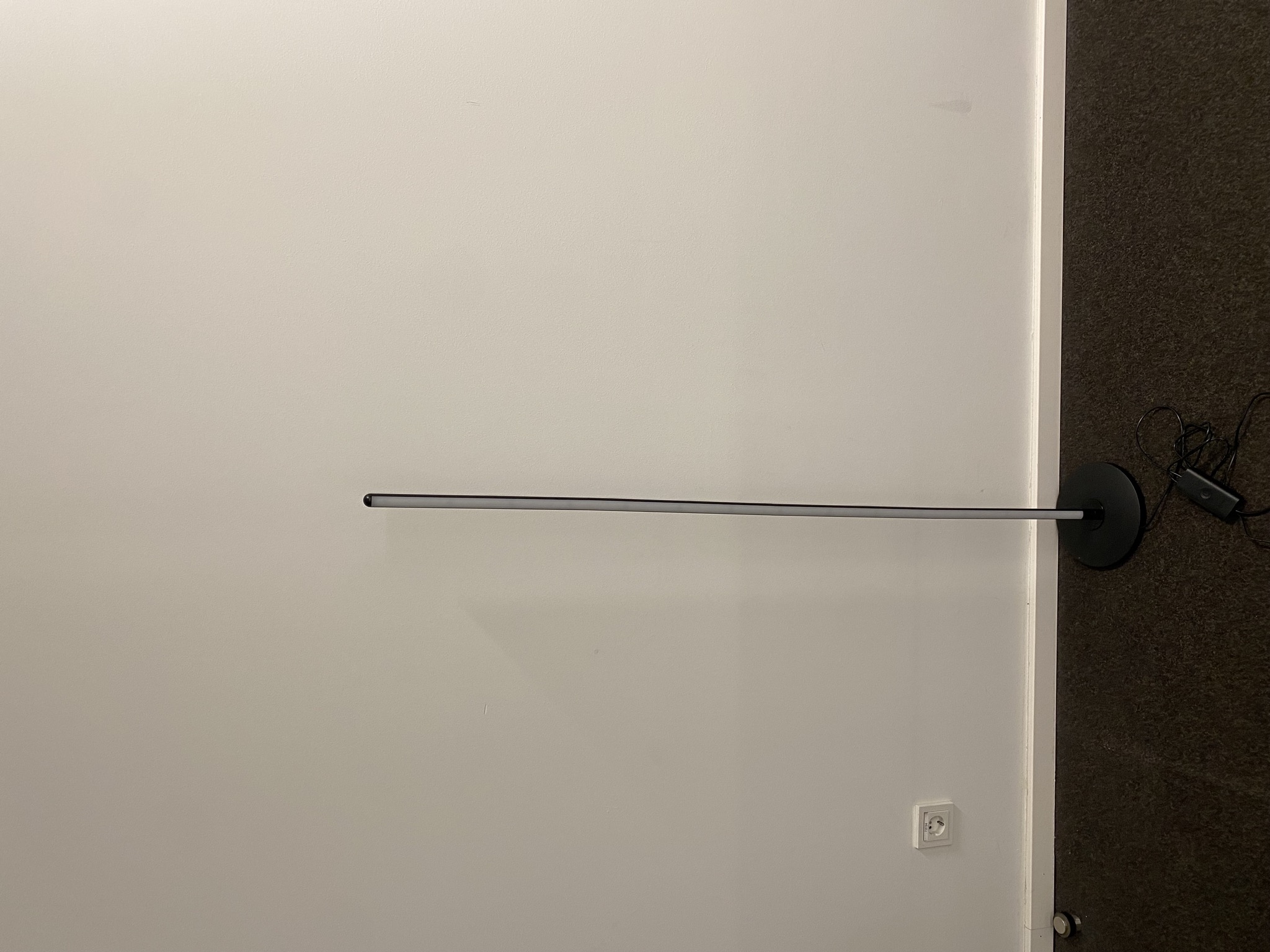}
        \caption*{Light Bar}
    \end{subfigure}\hfill
    \begin{subfigure}{.14\linewidth}
        \centering
        \includegraphics[width=\linewidth, angle=-90]{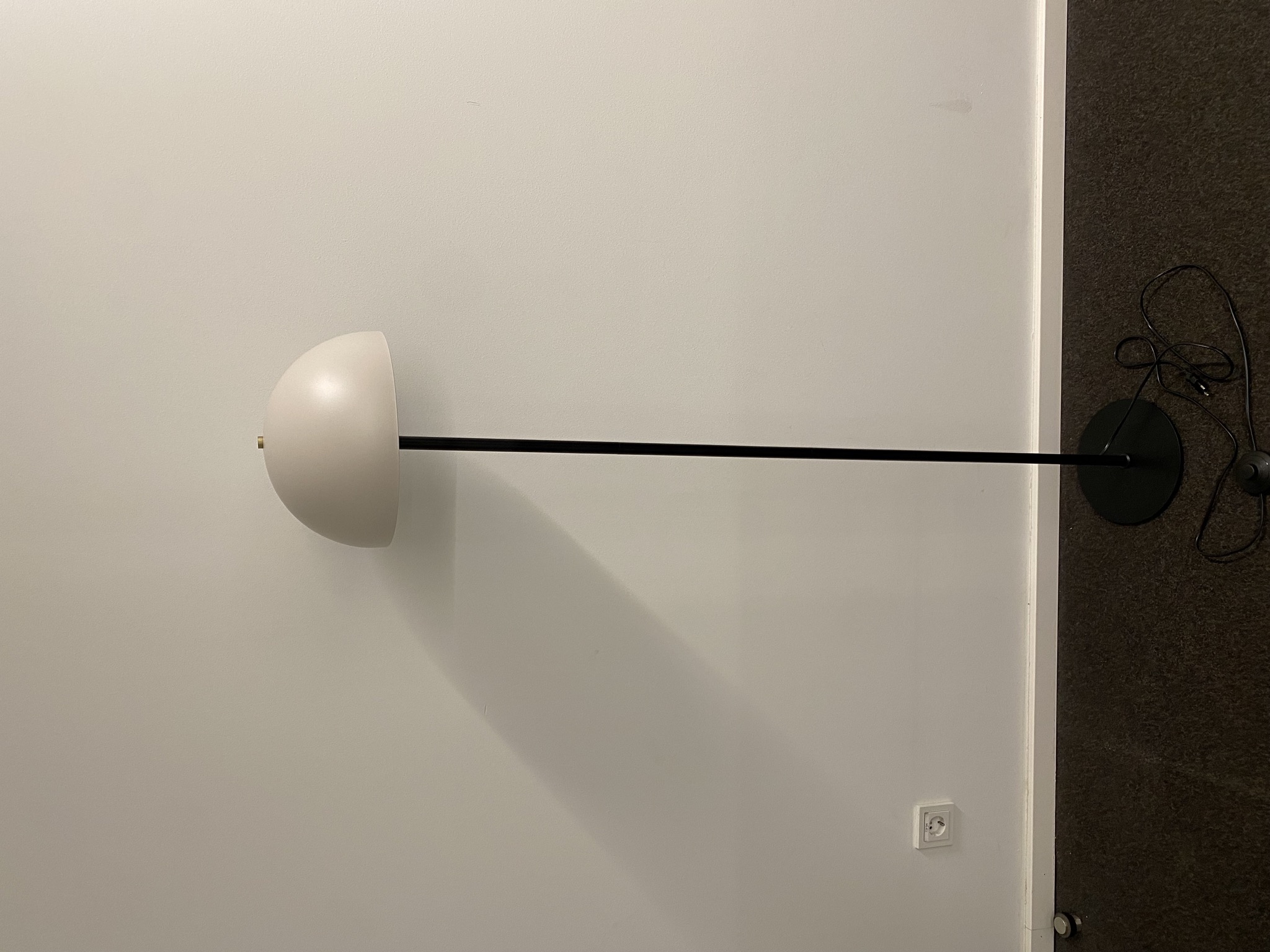}
        \caption*{Shaded}
    \end{subfigure}
    \caption{\textbf{Images of tested lamps.} Visual reference for the seven different lamp types used in \cref{sec:LightSourceTypes} and \cref{supp:ambient_light_complexity_evaluation}: Spherical, Area, Basket, Daylight, Directional, Light Bar, and Shaded. For the main 3DLP benchmark (\cref{tab:main_benchmark}) we used the spherical lamp.}
    \label{fig:light_types}
\end{figure*}

\subsection{Ambient light complexity evaluation}
\label{supp:ambient_light_complexity_evaluation}

Generating physically consistent illumination requires models to disentangle ambient light from underlying material properties, such as albedo. Consequently, we conjecture that image editing models perform implicitly intrinsic image decomposition, a task whose difficulty likely scales with the complexity and type of ambient lighting. To analyse this, we did two additional experiments. 

To assess model performance across different illumination regimes, we categorised our dataset into scenarios lit exclusively by distant sources (e.g., sunlight), near-field sources (e.g., indoor lamps), or a mixture of both. Our 3DLP dataset comprises 425 naturally lit, 190 artificially lit, and 385 mixed-lighting scenes. As shown in \cref{tab:ambient_lighting_conditions}, SIE and LFE metrics may suggest a marginal trend: models tend to perform better on naturally lit scenes and struggle more with mixed lighting conditions.

To further isolate the effects of light source geometry and spectral characteristics, we captured an additional targeted subset of 300 HDR pairs with systematically increasing ambient complexity, ranging from simple point light sources to complex multi-light configurations. This subset includes 10 new scenes with 5 views each, evaluated across 6 ambient lamp types (spherical bulb, 60\,cm$\times$120\,cm area light, single and dual directional lamps, and single and dual complex basket lamps), see \cref{{fig:light_types}}. These lamps were placed behind the camera to eliminate ambiguity regarding which lamp the model has to turn on/off (the target remains the spherical bulb from the main 3DLP benchmark, \cref{tab:main_benchmark}).

Results for these lamp types are presented in \cref{tab:ambient_light_ablation_on_off_80}. Based on this data, we cannot confidently conclude that ambient light type significantly impacts model performance. We hypothesise that the interaction of ambient light with scene geometry, specifically the casting of complex shadows and highlights, poses a more substantial challenge to generative models than the complexity of the ambient light sources themselves.

\begin{table}[htbp]
    \centering
    \caption{\textbf{Light regime evaluation.} Breakdown of model performance by ambient light types. Metrics reported as SIE and LFE, where lower is better.} %
    \label{tab:ambient_lighting_conditions}
    \resizebox{0.6\textwidth}{!}{
    \begin{tabular}{lcccccc}
        \toprule
        & \multicolumn{2}{c}{\textbf{Indoor}} 
        & \multicolumn{2}{c}{\textbf{Natural}} 
        & \multicolumn{2}{c}{\textbf{Mixed}} \\
        \cmidrule(lr){2-3}
        \cmidrule(lr){4-5}
        \cmidrule(lr){6-7}
        \textbf{Model} 
        & \textbf{SIE} $\downarrow$ & \textbf{LFE} $\downarrow$
        & \textbf{SIE} $\downarrow$ & \textbf{LFE} $\downarrow$
        & \textbf{SIE} $\downarrow$ & \textbf{LFE} $\downarrow$ \\
        \midrule
        Nano Banana Pro & 1.195 & 1.719 & 1.083 & 1.693 & 1.469 & 1.831 \\
        Nano Banana 2   & 1.226 & 1.716 & 1.211 & 1.701 & 1.655 & 1.825 \\
        Flux 2 Max      & 2.295 & 1.830 & 1.936 & 1.761 & 2.432 & 1.880 \\
        Flux 2 Dev      & 2.434 & 1.877 & 2.669 & 1.857 & 2.723 & 1.993 \\
        GPT Img. 1.5    & 3.009 & 2.072 & 2.813 & 2.009 & 3.233 & 2.163 \\
        Qwen-Img.-Edit  & 1.785 & 1.833 & 1.620 & 1.877 & 1.995 & 2.010 \\
        \bottomrule
    \end{tabular}
    }
\end{table}

\begin{table*}[htbp]
    \centering
    \caption{\textbf{Ambient light complexity evaluation.} SIE and LFE on ambient-light experiments for \textit{turn-on} and \textit{turn-off} using the best 80\% of images. Lower is better.}
    \label{tab:ambient_light_ablation_on_off_80}
    \resizebox{\textwidth}{!}{
    \begin{tabular}{ll cccccccccccc}
        \toprule
        & & \multicolumn{2}{c}{\textbf{Spherical}} & \multicolumn{2}{c}{\textbf{Area}} & \multicolumn{2}{c}{\textbf{Directional One}} & \multicolumn{2}{c}{\textbf{Directional Two}} & \multicolumn{2}{c}{\textbf{Complex One}} & \multicolumn{2}{c}{\textbf{Complex Two}} \\
        \cmidrule(lr){3-4} \cmidrule(lr){5-6} \cmidrule(lr){7-8} \cmidrule(lr){9-10} \cmidrule(lr){11-12} \cmidrule(lr){13-14}
        \textbf{Task} & \textbf{Method} & \textbf{SIE} $\downarrow$ & \textbf{LFE} $\downarrow$ & \textbf{SIE} $\downarrow$ & \textbf{LFE} $\downarrow$ & \textbf{SIE} $\downarrow$ & \textbf{LFE} $\downarrow$ & \textbf{SIE} $\downarrow$ & \textbf{LFE} $\downarrow$ & \textbf{SIE} $\downarrow$ & \textbf{LFE} $\downarrow$ & \textbf{SIE} $\downarrow$ & \textbf{LFE} $\downarrow$ \\
        \midrule
        \multirow{6}{*}{\rotatebox[origin=c]{90}{Turn-On}} & Nano Banana Pro & \textbf{0.819} & \textbf{1.678} & \textbf{0.980} & \underline{1.656} & \textbf{0.940} & \underline{1.645} & \textbf{1.019} & \underline{1.706} & \textbf{0.854} & \underline{1.489} & \textbf{0.862} & \underline{1.562} \\
        & Nano Banana 2 & \underline{1.006} & \underline{1.694} & \underline{0.996} & \textbf{1.640} & \underline{1.127} & 1.659 & \underline{1.076} & \textbf{1.700} & 1.049 & 1.531 & \underline{1.049} & 1.575 \\
        & Qwen-Img.-Edit & 1.141 & 1.722 & 1.363 & 1.782 & 1.234 & \textbf{1.624} & 1.145 & 1.711 & \underline{1.045} & \textbf{1.467} & 1.101 & \textbf{1.530} \\
        & Flux 2 Max & 1.473 & 1.731 & 1.499 & 1.701 & 1.468 & 1.684 & 1.583 & 1.733 & 1.500 & 1.605 & 1.653 & 1.656 \\
        & Flux 2 Dev & 1.508 & 1.806 & 1.679 & 1.755 & 1.378 & 1.706 & 1.591 & 1.796 & 1.510 & 1.615 & 1.499 & 1.639 \\
        & GPT Img. 1.5 & 2.143 & 1.870 & 2.200 & 1.879 & 1.995 & 1.810 & 1.975 & 1.892 & 1.986 & 1.710 & 1.929 & 1.720 \\
        \midrule
        \multirow{6}{*}{\rotatebox[origin=c]{90}{Turn-Off}} & Nano Banana Pro & \textbf{0.633} & 1.727 & \textbf{0.762} & \underline{1.653} & \textbf{1.110} & 1.780 & \underline{0.942} & 1.786 & \underline{1.359} & 1.698 & \underline{1.188} & 1.747 \\
        & Nano Banana 2 & 0.837 & 1.704 & \underline{0.964} & 1.657 & 1.355 & 1.760 & 1.406 & 1.791 & 1.533 & 1.686 & 1.378 & 1.702 \\
        & Qwen-Img.-Edit & \underline{0.678} & \textbf{1.649} & 1.186 & \textbf{1.605} & \underline{1.153} & \textbf{1.726} & \textbf{0.903} & \textbf{1.719} & \textbf{0.895} & \textbf{1.597} & \textbf{0.777} & \textbf{1.625} \\
        & Flux 2 Max & 1.033 & \underline{1.672} & 1.186 & 1.706 & 1.536 & \underline{1.741} & 1.342 & \underline{1.752} & 1.407 & \underline{1.614} & 1.480 & \underline{1.639} \\
        & Flux 2 Dev & 2.164 & 1.692 & 2.329 & 1.774 & 2.528 & 1.781 & 2.270 & 1.766 & 2.342 & 1.662 & 2.177 & 1.673 \\
        & GPT Img. 1.5 & 1.448 & 1.738 & 2.080 & 1.669 & 1.808 & 1.774 & 1.569 & 1.766 & 1.444 & 1.631 & 1.479 & 1.671 \\
        \bottomrule
    \end{tabular}
    }
\end{table*}

\subsection{Outdoor night-time images}
\label{supp:night_images}

To assess the feasibility of extending our evaluation to outdoor scenes, we captured an additional 100 night-time outdoor HDR image pairs.\cref{tab:outdoor_night_benchmark} reports the results for the six models evaluated in the main benchmark.

\begin{table}[htbp]
\centering
\caption{\textbf{Outdoor night-time evaluation.} SIE and LFE for the \textit{turn-on} and \textit{turn-off} tasks on 100 outdoor HDR image pairs captured at night. Best performance is marked in \textbf{bold}, and runner-up is \underline{underlined}.}
\label{tab:outdoor_night_benchmark}
\resizebox{0.71\linewidth}{!}{
\begin{tabular}{lccccc}
\toprule
& \multicolumn{2}{c}{\textbf{Turning Light On}} & \multicolumn{2}{c}{\textbf{Turning Light Off}} & \\
\cmidrule(lr){2-3} \cmidrule(lr){4-5}
\textbf{Model}
& \textbf{SIE} $\downarrow$
& \textbf{LFE} $\downarrow$
& \textbf{SIE} $\downarrow$
& \textbf{LFE} $\downarrow$
& \textbf{Avg. Rank} $\downarrow$ \\
\midrule
Nano Banana Pro & \textbf{1.675} & \textbf{1.651} & \textbf{0.881} & \textbf{1.491} & \textbf{1.0} \\
Nano Banana 2 & \underline{1.864} & \underline{1.668} & \underline{0.970} & 1.510 & \underline{2.3} \\
Qwen-Img.-Edit & 2.307 & 1.770 & 0.998 & \underline{1.497} & 2.8 \\
Flux 2 Max & 2.763 & 1.794 & 1.323 & 1.560 & 4.0 \\
Flux 2 Dev & 3.591 & 1.860 & 2.107 & 1.593 & 5.5 \\
GPT Img. 1.5 & 4.006 & 1.938 & 1.721 & 1.584 & 5.5 \\
\bottomrule
\end{tabular}
}
\end{table}

The overall model ranking remains unchanged from the indoor benchmark (\cref{tab:main_benchmark}), including the tie between Flux 2 Dev and GPT Img.\ 1.5. Individual rankings change for LFE: Qwen-Img.-Edit overtakes Flux 2 Max on the \textit{turn-on} task and Nano Banana 2 on the \textit{turn-off} task. \textit{Turn-on} SIE is substantially worse (higher) for all models than on the indoor dataset, suggesting that this setting is more challenging. Models must generate plausible content in regions that are nearly pitch black in the input, while also reproducing unpredictable lens flare present in the real light-on images. These effects complicate the evaluation of light transport. Addressing them may require new metrics or additional model guidance to reproduce lens flare, which we leave for future work.

\subsection{User study on human-metric agreement}
\label{supp:user_study}

We conducted a small user study with 19 participants, collecting 970 random A/B comparisons of outputs from two models performing the same \textit{turn-on} or \textit{turn-off} task. We measure human-metric agreement as the proportion of human decisions favouring the output with lower SIE or LFE. For each task and metric, we sort comparisons by absolute metric difference into four equally sized quartiles, from Q1 (nearly tied scores) to Q4 (largest separation). As shown in \cref{tab:user_study_agreement}, agreement generally increases with metric separation, reaching 78.4-81.8\% for SIE and 69.7-78.4\% for LFE in Q4. These results support alignment with human judgement and suggest that small score differences are less perceptually meaningful. In a post-study questionnaire, participants reported finding the \textit{turn-off} task more difficult and primarily considered missing or misaligned shadows and highlights, as well as overly bright highlights.

\begin{table}[htbp]
\centering
\caption{\textbf{Human--metric agreement.} Percentage of human decisions favouring the output with lower SIE or LFE, grouped by absolute metric difference from Q1 (smallest) to Q4 (largest). Overall agreement includes all comparisons for the respective task and metric.}
\label{tab:user_study_agreement}
\small
\begin{tabular}{llccccc}
\toprule
\textbf{Task} & \textbf{Metric} & \textbf{Q1} & \textbf{Q2} & \textbf{Q3} & \textbf{Q4} & \textbf{Overall} \\
\midrule
\multirow{2}{*}{Turn-On} & SIE & 63.8\% & 63.2\% & 76.1\% & 78.4\% & 70.4\% \\
& LFE & 49.1\% & 70.1\% & 67.5\% & 78.4\% & 66.3\% \\
\midrule
\multirow{2}{*}{Turn-Off} & SIE & 48.5\% & 65.7\% & 78.8\% & 81.8\% & 68.7\% \\
& LFE & 43.4\% & 54.5\% & 65.7\% & 69.7\% & 58.3\% \\
\bottomrule
\end{tabular}
\end{table}

\subsection{Prompt ablation on light-source descriptions}
\label{supp:light_source_prompt_ablation}

During prompt design, we extensively tested descriptions of the light source and observed that specifying its emission pattern can occasionally introduce artefacts, such as glowing spheres. We therefore omitted emission-pattern information from the final prompts, allowing models freedom in generating the lighting. To quantify the effect of additional light-source information, we evaluated 100 scenes with three prompt settings: the original prompt, a prompt augmented with the spherical bulb's omnidirectional emission pattern, and a prompt augmented with the target colour temperature. \cref{tab:light_source_prompt_ablation} reports the two augmented settings, with deltas relative to the original prompt. Specifying the emission pattern slightly degrades performance for most models. Adding the target colour temperature generally produces modest metric changes, consistent with the intended invariance of our metrics to the lamp's colour and intensity. We thus conclude that broad, general prompts do not negatively impact model performance.

\begin{table}[htbp]
\centering
\caption{\textbf{Prompt ablation on light-source descriptions.} SIE and LFE on 100 scenes when adding emission-pattern or target-colour information to the original prompt. Parentheses show changes relative to the original prompt; positive deltas indicate increased error. Lower scores are better.}
\label{tab:light_source_prompt_ablation}
\resizebox{\linewidth}{!}{
\begin{tabular}{lcccc}
\toprule
& \multicolumn{2}{c}{\textbf{Turning Light On}} & \multicolumn{2}{c}{\textbf{Turning Light Off}} \\
\cmidrule(lr){2-3} \cmidrule(lr){4-5}
\textbf{Model} & \textbf{SIE} $\downarrow$ & \textbf{LFE} $\downarrow$ & \textbf{SIE} $\downarrow$ & \textbf{LFE} $\downarrow$ \\
\midrule
\multicolumn{5}{l}{\textbf{Adding the spherical bulb's emission pattern}} \\
\midrule
Nano Banana Pro & 0.874 ($+0.036$) & 1.664 ($+0.019$) & 0.763 ($+0.026$) & 1.665 ($+0.003$) \\
Nano Banana 2 & 0.935 ($+0.052$) & 1.651 ($+0.008$) & 0.922 ($+0.063$) & 1.665 ($\pm0.000$) \\
Qwen-Img.-Edit & 1.261 ($-0.008$) & 1.770 ($+0.016$) & 0.941 ($-0.031$) & 1.670 ($\pm0.000$) \\
Flux 2 Max & 1.575 ($+0.022$) & 1.709 ($-0.008$) & 1.283 ($+0.014$) & 1.708 ($+0.004$) \\
Flux 2 Dev & 1.989 ($+0.025$) & 1.794 ($-0.001$) & 1.618 ($+0.051$) & 1.834 ($+0.013$) \\
GPT Img. 1.5 & 2.346 ($+0.017$) & 1.926 ($-0.003$) & 1.918 ($+0.026$) & 1.763 ($+0.025$) \\
\midrule
\multicolumn{5}{l}{\textbf{Adding the target colour}} \\
\midrule
Nano Banana Pro & 0.820 ($-0.018$) & 1.646 ($+0.001$) & 0.772 ($+0.035$) & 1.667 ($+0.005$) \\
Nano Banana 2 & 0.894 ($+0.011$) & 1.653 ($+0.010$) & 0.976 ($+0.117$) & 1.664 ($-0.001$) \\
Qwen-Img.-Edit & 1.281 ($+0.012$) & 1.786 ($+0.032$) & 0.918 ($-0.054$) & 1.665 ($-0.005$) \\
Flux 2 Max & 1.292 ($-0.026$) & 1.673 ($-0.044$) & 1.324 ($+0.055$) & 1.702 ($-0.002$) \\
Flux 2 Dev & 1.640 ($-0.090$) & 1.761 ($-0.034$) & 2.344 ($+0.024$) & 1.824 ($+0.003$) \\
GPT Img. 1.5 & 2.119 ($-0.021$) & 1.977 ($+0.048$) & 1.922 ($+0.030$) & 1.756 ($+0.018$) \\
\bottomrule
\end{tabular}
}
\end{table}

\subsection{Simultaneous and sequential editing of two lamps}
\label{supp:two_lamp_editing}

We captured 50 HDR image triplets, each showing three lighting states: both lamps switched off, only the lamp on a black stand switched on, and both lamps switched on. We retained the spherical light bulbs used in the main benchmark, but used a black and a grey stand to distinguish the lamps in the prompts. Models were asked to switch both lamps on or off either simultaneously in a single generation or sequentially across two generations. For sequential turn-on, the lamp on the black stand was switched on first, followed by the lamp on the grey stand.

\cref{tab:two_lamp_editing} compares the two settings. Simultaneous switching yields lower SIE for every model on both tasks. The differences in LFE are smaller and do not consistently favour simultaneous switching. The poorer SIE of sequential editing is consistent with errors accumulating across the two generations.

\begin{table}[htbp]
\centering
\caption{\textbf{Two-lamp editing.} SIE and LFE for switching two lamps on or off sequentially or simultaneously, evaluated on 50 HDR image triplets. Lower is better. Best performance within each setting is marked in \textbf{bold}, and runner-up is \underline{underlined}.}
\label{tab:two_lamp_editing}
\resizebox{0.71\linewidth}{!}{
\begin{tabular}{lccccc}
\toprule
& \multicolumn{2}{c}{\textbf{Turning Light On}} & \multicolumn{2}{c}{\textbf{Turning Light Off}} & \\
\cmidrule(lr){2-3} \cmidrule(lr){4-5}
\textbf{Model} & \textbf{SIE} $\downarrow$ & \textbf{LFE} $\downarrow$ & \textbf{SIE} $\downarrow$ & \textbf{LFE} $\downarrow$ & \textbf{Avg. Rank} $\downarrow$ \\
\midrule
\multicolumn{6}{l}{\textbf{Sequential switching}} \\
\midrule
Nano Banana Pro & \textbf{1.307} & \underline{1.790} & \textbf{1.115} & 1.739 & \textbf{1.8} \\
Nano Banana 2 & \underline{1.332} & \textbf{1.784} & \underline{1.416} & \underline{1.730} & \textbf{1.8} \\
Qwen-Img.-Edit & 1.882 & 2.066 & 1.452 & \textbf{1.713} & \underline{3.5} \\
Flux 2 Max & 1.839 & 1.834 & 1.497 & 1.765 & 3.8 \\
Flux 2 Dev & 1.927 & 1.864 & 1.910 & 1.785 & 5.0 \\
GPT Img. 1.5 & 2.657 & 1.953 & 2.140 & 1.753 & 5.3 \\
\midrule
\multicolumn{6}{l}{\textbf{Simultaneous switching}} \\
\midrule
Nano Banana Pro & \textbf{0.736} & \underline{1.686} & \textbf{0.687} & \underline{1.687} & \textbf{1.5} \\
Nano Banana 2 & \underline{0.766} & 1.696 & \underline{0.863} & \textbf{1.682} & \underline{2.0} \\
Qwen-Img.-Edit & 1.539 & 1.961 & 1.112 & 1.734 & 4.0 \\
Flux 2 Max & 1.285 & \textbf{1.651} & 1.285 & 1.748 & 3.0 \\
Flux 2 Dev & 1.488 & 1.804 & 1.368 & 1.871 & 4.8 \\
GPT Img. 1.5 & 2.561 & 1.968 & 1.880 & 1.828 & 5.8 \\
\bottomrule
\end{tabular}
}
\end{table}

\subsection{Editing coloured light sources}
\label{supp:coloured_light_sources}

We captured 50 HDR image pairs with the light switched on and off, randomly sampling the emitted colour for each pair. We used a Philips Hue 1600\,lm White \& Colour bulb and included the target light colour in each model's prompt. As shown in \cref{tab:coloured_light_sources}, the ranking is broadly consistent with the main benchmark (\cref{tab:main_benchmark}), although SIE is generally higher, particularly for the \textit{turn-on} task.

Three factors may contribute to this degradation. The Hue bulb is not perfectly spherical, which can impair performance (\cref{sec:LightSourceTypes}). Its coloured output is substantially below the advertised 1600\,lm, requiring dimmer ambient illumination. Finally, our metrics are computed per colour channel and then averaged. If the ground-truth light lacks signal in some channels, generating a different light colour (e.g., red instead of blue) can incur a large penalty.

\begin{table}[htbp]
\centering
\caption{\textbf{Coloured light-source evaluation.} SIE and LFE on 50 HDR image pairs captured with randomly sampled light colours. The target colour is specified in the prompt. Lower is better. Best performance is marked in \textbf{bold}, and runner-up is \underline{underlined}.}
\label{tab:coloured_light_sources}
\resizebox{0.71\linewidth}{!}{
\begin{tabular}{lccccc}
\toprule
& \multicolumn{2}{c}{\textbf{Turning Light On}} & \multicolumn{2}{c}{\textbf{Turning Light Off}} & \\
\cmidrule(lr){2-3} \cmidrule(lr){4-5}
\textbf{Model} & \textbf{SIE} $\downarrow$ & \textbf{LFE} $\downarrow$ & \textbf{SIE} $\downarrow$ & \textbf{LFE} $\downarrow$ & \textbf{Avg. Rank} $\downarrow$ \\
\midrule
Nano Banana Pro & \textbf{1.369} & \textbf{1.666} & \textbf{0.832} & \underline{1.650} & \textbf{1.3} \\
Nano Banana 2 & \underline{1.546} & \underline{1.692} & \underline{0.926} & 1.653 & \underline{2.3} \\
Qwen-Img.-Edit & 1.932 & 1.890 & 0.986 & \textbf{1.616} & 3.0 \\
Flux 2 Max & 2.739 & 1.808 & 1.527 & 1.666 & 4.3 \\
Flux 2 Dev & 2.071 & 1.786 & 1.817 & 1.706 & 4.3 \\
GPT Img. 1.5 & 3.135 & 1.914 & 1.946 & 1.730 & 6.0 \\
\bottomrule
\end{tabular}
}
\end{table}

\section{Details to experiments of main article}
\label{supp:details_to_experiments}
\subsection{Results for all models and percentiles}
\label{supp:All_Models_All_Percentiles}

\cref{tab:all_models_metric_percentiles} extends the results presented in the main article in \cref{tab:main_benchmark}. It shows all tested models, including Bagel 7B MoT and OmniGen2, evaluated on the top $x\%$ of images, where $x$ ranges from $100\%$ down to $50\%$. Upon visual inspection we confirmed that all six models featured in the main benchmark (\cref{tab:main_benchmark}) successfully perform the requested tasks in well over $80\%$ of the images. This was not true for Bagel 7B MoT and OmniGen2. For instance they sometimes turn on or off the wrong light probe in the scene, or make no changes to the image. Independent of this problem,  both models perform worst with respect to SIE for all percentiles, i.e from $100\%$ down to $50\%$.
Furthermore, the table demonstrates that restricting the evaluation to only the top $x\%$ of images does not significantly alter the overall ranking of the models.

\begin{table}[htbp]
\centering
\caption{\textbf{SIE and LFE on all models and percentiles.} We report SIE and LFE for the \textit{turn-on} and \textit{turn-off} task computed for the best $x\%$ of images for each model. Bagel 7B and OmniGen2 are not part of the main 3DLP benchmark (\cref{tab:main_benchmark}), as they fail to understand the task in more than 20\% of images. Best performance is marked in \textbf{bold}, while the runner-up is \underline{underlined}.}
\label{tab:all_models_metric_percentiles}
\resizebox{\linewidth}{!}{
\begin{tabular}{llcccccccccccc}
\toprule
 &  & \multicolumn{2}{c}{\textbf{Best 100\%}} & \multicolumn{2}{c}{\textbf{Best 90\%}} & \multicolumn{2}{c}{\textbf{Best 80\%}} & \multicolumn{2}{c}{\textbf{Best 70\%}} & \multicolumn{2}{c}{\textbf{Best 60\%}} & \multicolumn{2}{c}{\textbf{Best 50\%}} \\
\cmidrule(lr){3-4} \cmidrule(lr){5-6} \cmidrule(lr){7-8} \cmidrule(lr){9-10} \cmidrule(lr){11-12} \cmidrule(lr){13-14}
 \textbf{Task} & \textbf{Model} & \textbf{SIE} $\downarrow$ & \textbf{LFE} $\downarrow$ & \textbf{SIE} $\downarrow$ & \textbf{LFE} $\downarrow$ & \textbf{SIE} $\downarrow$ & \textbf{LFE} $\downarrow$ & \textbf{SIE} $\downarrow$ & \textbf{LFE} $\downarrow$ & \textbf{SIE} $\downarrow$ & \textbf{LFE} $\downarrow$ & \textbf{SIE} $\downarrow$ & \textbf{LFE} $\downarrow$ \\
\midrule
\multirow{8}{*}{\rotatebox[origin=c]{90}{Turn-On}}
& Nano Banana Pro$^\ast$ & \textbf{1.247} & \textbf{1.749} & \textbf{0.905} & \textbf{1.674} & \textbf{0.813} & \textbf{1.641} & \textbf{0.744} & \underline{1.612} & \textbf{0.687} & \underline{1.585} & \textbf{0.637} & \underline{1.559} \\
& Nano Banana 2$^\ast$ & \underline{1.377} & \underline{1.750} & \underline{0.984} & \underline{1.675} & \underline{0.882} & \underline{1.641} & \underline{0.805} & \textbf{1.611} & \underline{0.737} & \textbf{1.583} & \underline{0.680} & \textbf{1.557} \\
& Qwen-Img.-Edit$^\dagger$ & 1.790 & 1.925 & 1.361 & 1.823 & 1.239 & 1.772 & 1.147 & 1.729 & 1.068 & 1.690 & 0.998 & 1.652 \\
& Flux 2 Max$^\ast$ & 2.193 & 1.820 & 1.706 & 1.746 & 1.571 & 1.710 & 1.464 & 1.680 & 1.371 & 1.652 & 1.285 & 1.626 \\
& Flux 2 Dev$^\dagger$ & 2.640 & 1.915 & 1.959 & 1.824 & 1.747 & 1.775 & 1.591 & 1.737 & 1.460 & 1.702 & 1.343 & 1.668 \\
& GPT Img. 1.5$^\ast$ & 3.005 & 2.082 & 2.499 & 1.986 & 2.336 & 1.933 & 2.209 & 1.889 & 2.096 & 1.847 & 1.987 & 1.804 \\
& Bagel 7B MoT$^\dagger$ & 3.568 & 2.000 & 2.881 & 1.897 & 2.655 & 1.851 & 2.511 & 1.813 & 2.393 & 1.780 & 2.294 & 1.746 \\
& OmniGen2$^\dagger$ & 511.869 & 111.498 & 3.374 & 2.014 & 3.159 & 1.959 & 3.003 & 1.912 & 2.867 & 1.869 & 2.742 & 1.828 \\
\midrule
\multirow{8}{*}{\rotatebox[origin=c]{90}{Turn-Off}}
& Nano Banana Pro$^\ast$ & \textbf{1.012} & \textbf{1.727} & \textbf{0.855} & \textbf{1.676} & \textbf{0.769} & \textbf{1.651} & \textbf{0.707} & \underline{1.628} & \textbf{0.653} & \underline{1.606} & \textbf{0.604} & \underline{1.584} \\
& Nano Banana 2$^\ast$ & \underline{1.127} & \underline{1.738} & \underline{0.964} & \underline{1.679} & \underline{0.871} & \underline{1.652} & \underline{0.800} & 1.630 & \underline{0.737} & 1.607 & \underline{0.678} & 1.584 \\
& Qwen-Img.-Edit$^\dagger$ & 1.194 & 1.756 & 1.067 & 1.691 & 0.982 & 1.656 & 0.913 & \textbf{1.626} & 0.848 & \textbf{1.597} & 0.789 & \textbf{1.569} \\
& Flux 2 Max$^\ast$ & 1.509 & 1.774 & 1.388 & 1.722 & 1.303 & 1.691 & 1.228 & 1.665 & 1.157 & 1.641 & 1.086 & 1.616 \\
& Flux 2 Dev$^\dagger$ & 2.423 & 1.918 & 2.262 & 1.844 & 2.136 & 1.797 & 2.014 & 1.753 & 1.883 & 1.710 & 1.743 & 1.671 \\
& GPT Img. 1.5$^\ast$ & 2.091 & 1.826 & 1.966 & 1.766 & 1.887 & 1.734 & 1.816 & 1.705 & 1.746 & 1.678 & 1.676 & 1.651 \\
& Bagel 7B MoT$^\dagger$ & 2.599 & 1.797 & 2.420 & 1.737 & 2.322 & 1.706 & 2.245 & 1.680 & 2.178 & 1.656 & 2.116 & 1.632 \\
& OmniGen2$^\dagger$ & 3.145 & 1.882 & 2.836 & 1.809 & 2.700 & 1.771 & 2.595 & 1.739 & 2.501 & 1.709 & 2.412 & 1.680 \\
\bottomrule
\end{tabular}
}
\vspace{0.05cm}

{\footnotesize $^\ast$Commercial model. $^\dagger$Open-source model.}
\end{table}

\subsection{Metric results for annotation analysis}
\label{supp:full_metric_results_on_annotations}

\begin{figure}[htbp]
    \centering
    \includegraphics[width=\textwidth]{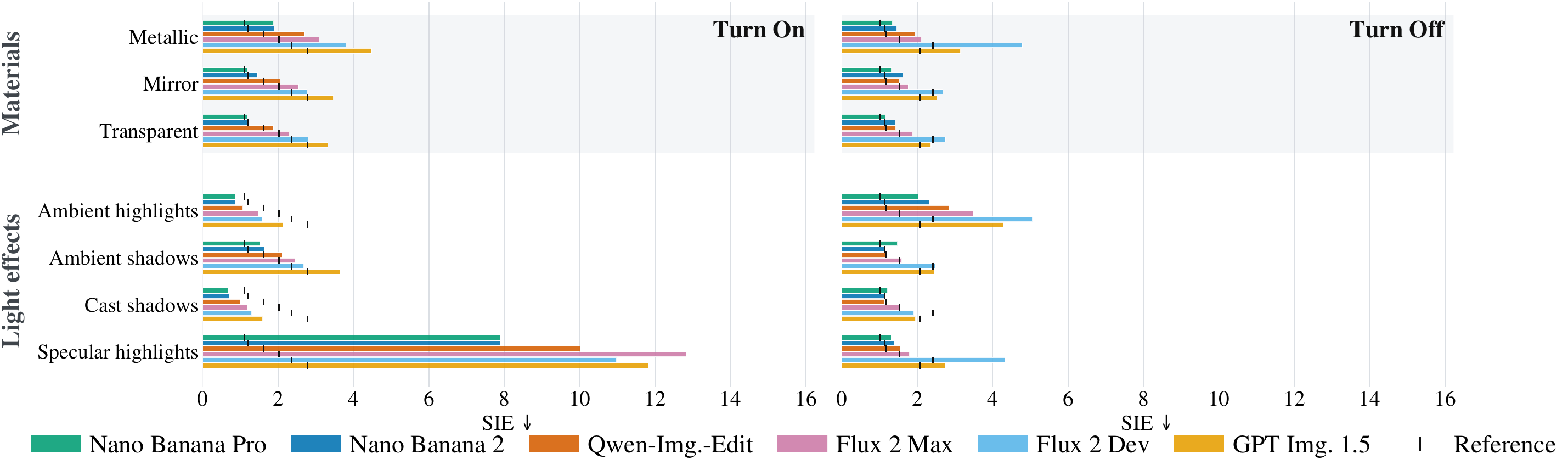}
    \caption{\textbf{SIE metric computed per annotated class.} Presented on an absolute scale (in comparison to \cref{fig:ablation_annotations}). Absolute differences between classes should not be used to rank ``hardness'', but should rather be seen as metric sensitivity and used for model performance comparisons. Black vertical bars indicate model performance on full images as a reference.}
    \label{fig:annotated_regions_all_sie}
\end{figure}

\begin{figure}[htbp]
    \centering
    \includegraphics[width=\textwidth]{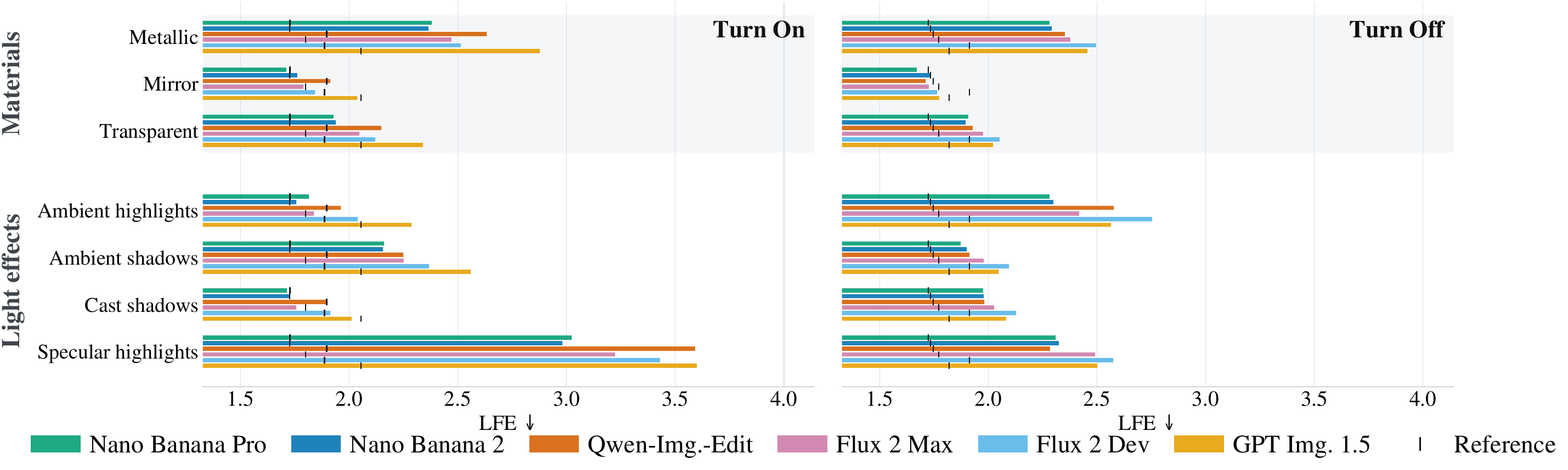}
    \caption{\textbf{LFE metric computed per annotated class.} Presented on an absolute scale (in comparison to \cref{fig:ablation_annotations}). Absolute differences between classes should not be used to rank ``hardness'', but should rather be seen as metric sensitivity and used for model performance comparisons. Black vertical bars indicate model performance on full images as a reference.}
    \label{fig:annotated_regions_all_lfe}
\end{figure}

We complement the annotation analysis experiment in \cref{subsec:Results}. Note that
\cref{fig:ablation_annotations} presented relative scores. \cref{fig:annotated_regions_all_sie} and \cref{fig:annotated_regions_all_lfe} present now the absolute scores per Material and Light Effect for SIE and LFE, respectively. 
These scores should not be interpreted as an absolute ranking of category ``hardness'', as category complexity is inherently entangled with metric sensitivity. For example, in a ``turn-on'' task, a missing highlight results in a large loss of light intensity in linear space, triggering a high error. Conversely, a missing shadow results in only a minor intensity overflow, yielding a disproportionately smaller error. Nonetheless, these annotations provide valuable insights into model comparison, benchmarking, and metric behaviour, serving as a foundation for future research on accurately annotated light and material effects.

\subsection{Lamps used for light probe ablation}
\label{supp:lamp_images}
\cref{fig:light_types} shows all six lamp types that were used in the light probe ablation experiments in \cref{sec:LightSourceTypes}. In our 3DLP dataset and benchmark we use the ``Spherical'' light, as the emission pattern is the easiest to be reproduced by generative models. See \cref{sec:LightSourceTypes} for more details on the impact of different lamp geometries on our scores.

\subsection{Intensity bands visualisations}
\label{supp:intensity_bands_vis}

\begin{figure}[htbp]
    \centering
    \includegraphics[width=\textwidth]{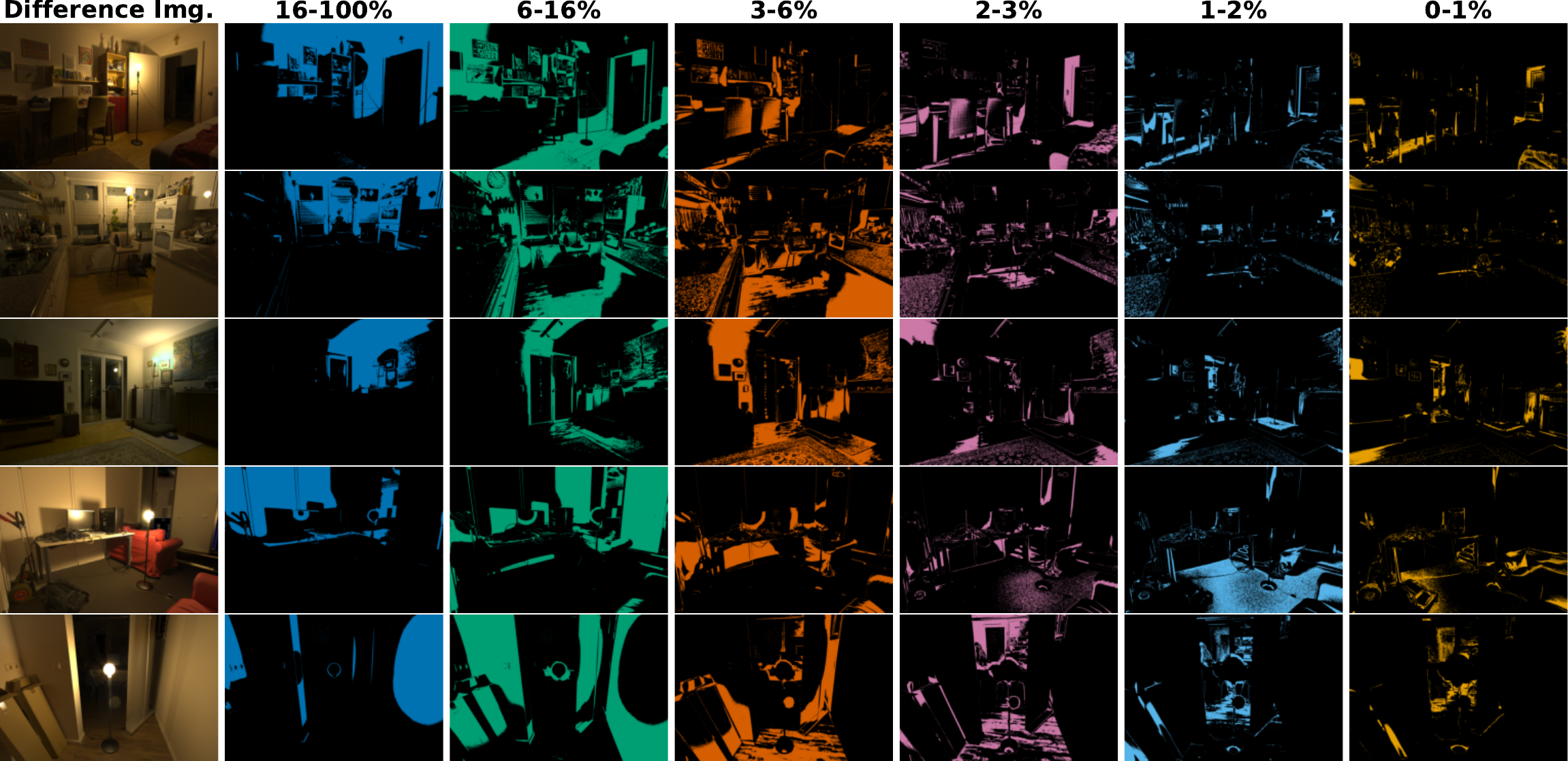}
    \caption{\textbf{Intensity band visualisations.} The difference images $I_{R}^\textrm{on}-I_R^\textrm{off}$ represent the intensity of the light probe. Each band captures a different range of intensities (colour coded for better visibility). The $16-100\%$ band includes the brightest pixels and the $0-1\%$ band the ones with lowest intensity.}
    \label{fig:light_band_visualisation}
\end{figure}

We visualise the intensity bands used in the light intensity band analysis in \cref{subsec:Results}. \cref{fig:light_band_visualisation} illustrates these bands for various images. As to be expected each band forms spatially quite compact regions. 
Generally, lower-intensity bands capture more distant regions, with respect to the light probe, though the exact distance is a function of scene geometry and surface reflectivity. 

\subsection{Additional qualitative results}

\begin{figure}[htbp]
    \centering
    \includegraphics[width=\textwidth]{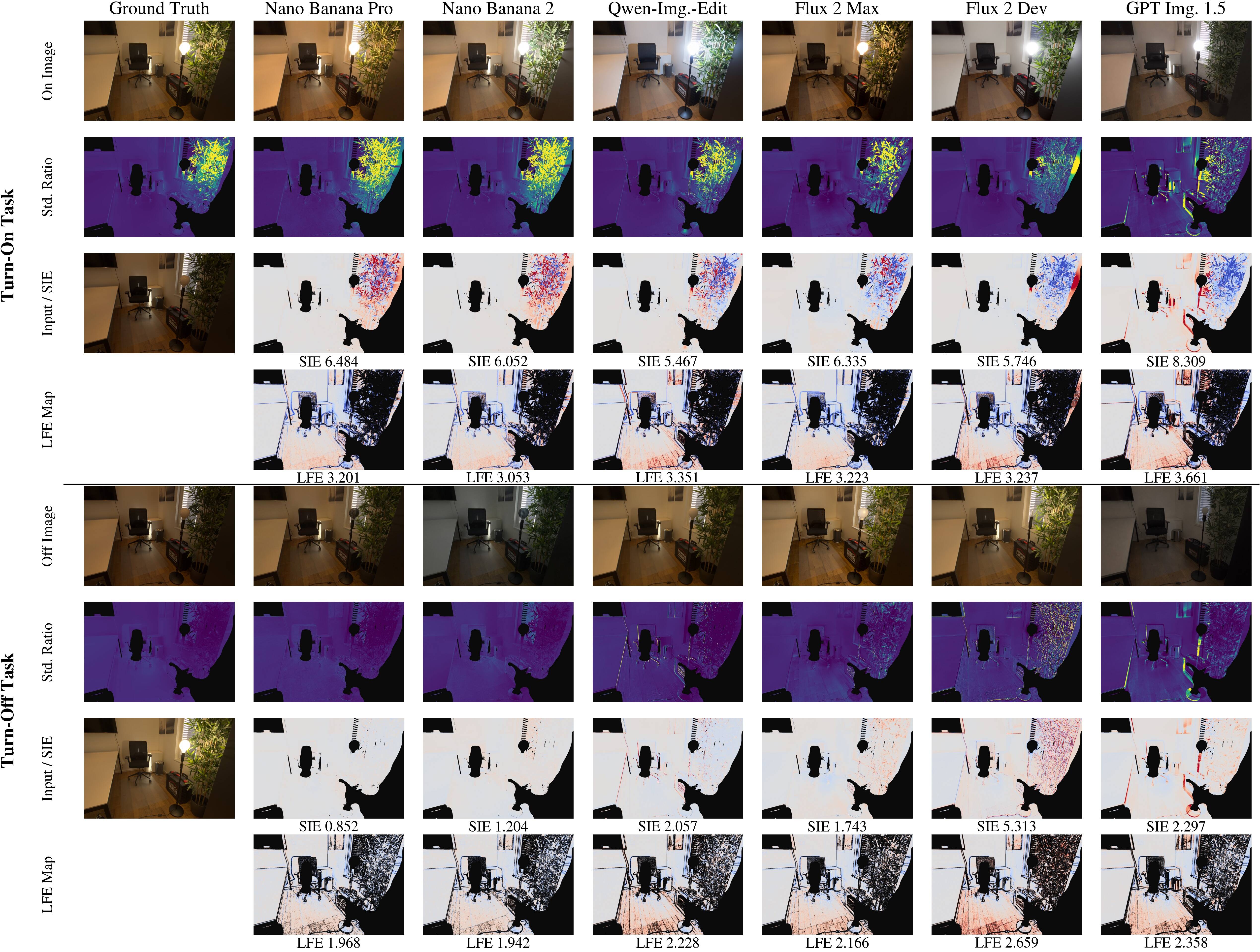}
    \caption{\textbf{Qualitative results.} This shows a challenging example with a plant next to the light bulb. Notice how GPT Image 1.5 fails to illuminate the plant at all. All other models also struggle with this complex light interaction for the turn-on task.
    The intensity ratio is defined as $E_{R}^t$ for the real image and $E_{G}^t$ for the generated images. The error map of SIE is 
    visualised as $\mathcal{S}(E^t_{R}) - \mathcal{S}(E^t_{G})$ and the LFE map as $\mathcal{S}(M_R^t)-\mathcal{S}(M_{G}^t)$. 
    }
    \label{supp:add_qualitative1}
\end{figure}

\begin{figure}[htbp]
    \centering
    \includegraphics[width=\textwidth]{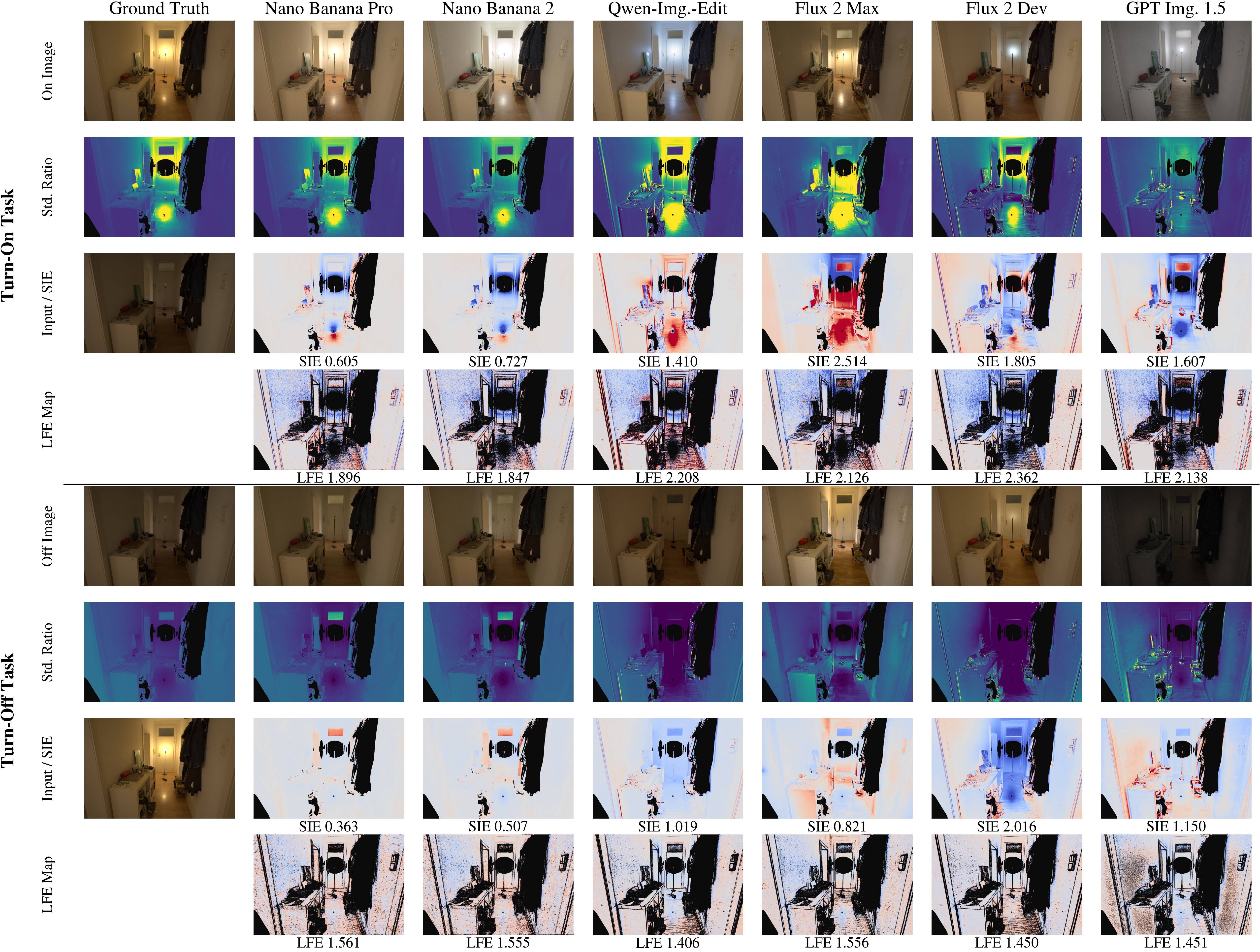}
    \caption{\textbf{Qualitative results.} Notice the highlight on the floor. While all models are able to successfully remove the highlight for the \textit{turn-off} task, all methods fail to reproduce the exact placement and shape of the highlight for the \textit{turn-on} task. The intensity ratios are defined as $E_{R}^t$ for the real image and $E_{G}^t$ for the generated images. The error map of SIE is 
    visualised as $\mathcal{S}(E^t_{R}) - \mathcal{S}(E^t_{G})$ and the LFE map as $\mathcal{S}(M_R^t)-\mathcal{S}(M_{G}^t)$. }
    \label{supp:add_qualitative2}
\end{figure}

\cref{supp:add_qualitative1,supp:add_qualitative2} show additional results for the turn-on and turn-off task. The results are shown for all models evaluated in the main paper. Additionally, we visualise the standardised intensity ratios for a better comparison to the ground truth.

\section{Dataset}
\label{supp:dataset}

\begin{figure}[htbp]
    \centering
    \includegraphics[width=\textwidth]{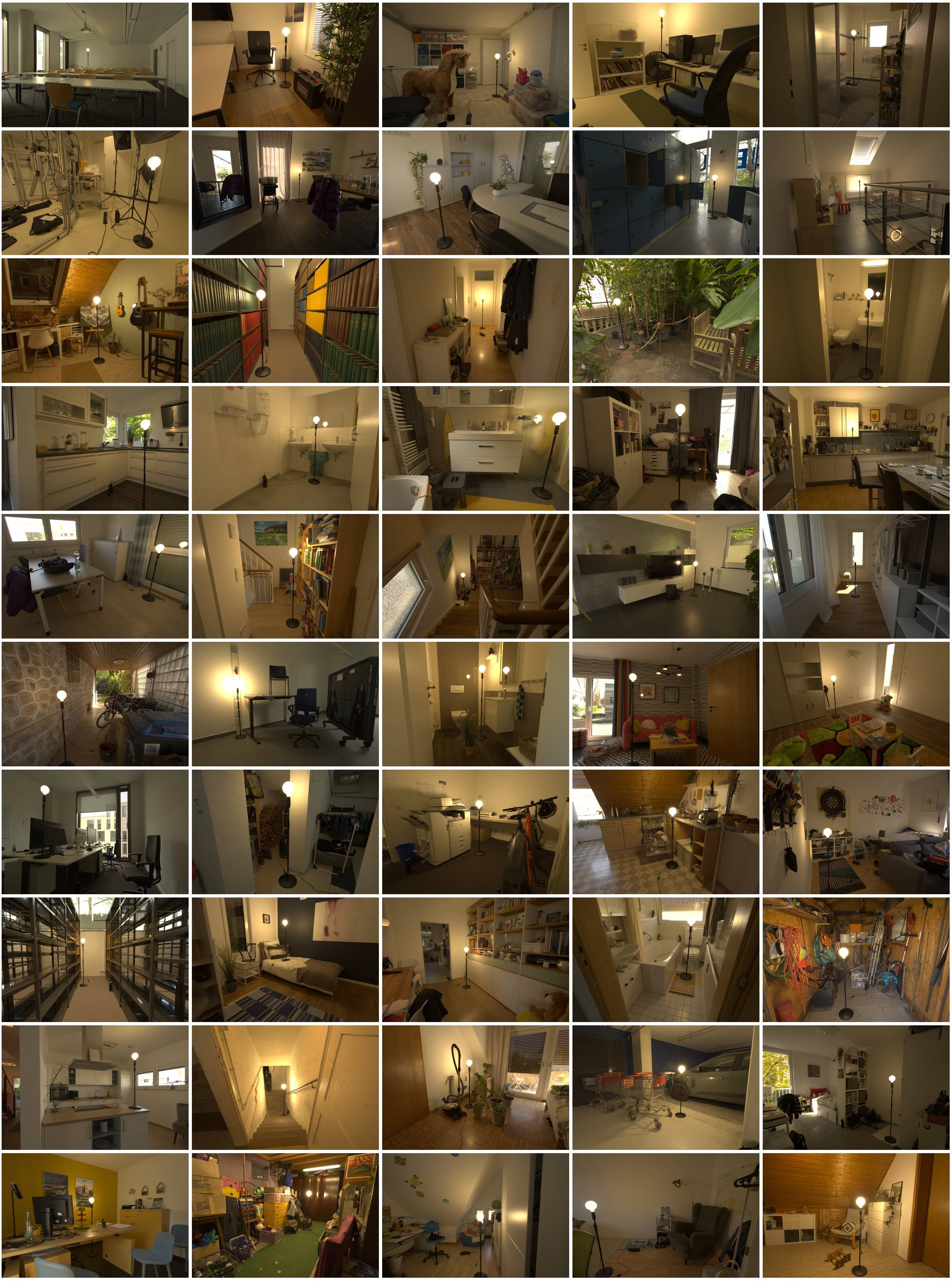}
    \caption{\textbf{3DLP Dataset diversity.} We visualise 50 randomly sampled \textit{on}-images from our dataset.} %
    \label{fig:supp_diversity_figure}
\end{figure}

We demonstrate the diversity of our dataset in \cref{fig:supp_diversity_figure} by visualising 50 randomly sampled \textit{on}-images. Note that each of these samples consists of an HDR image pair (\textit{on} and \textit{off} state), alongside the corresponding camera metadata as well as its annotation. We hope that this will encourage researchers to use our data to evaluate or improve methods for relighting or related tasks. Since the images are relightable (HDR), the camera intrinsics are known, and there are five different light positions available for each view, this dataset can also be useful for downstream tasks, such as material estimation.

\begin{figure}[htbp]
    \centering
    \includegraphics[width=\textwidth]{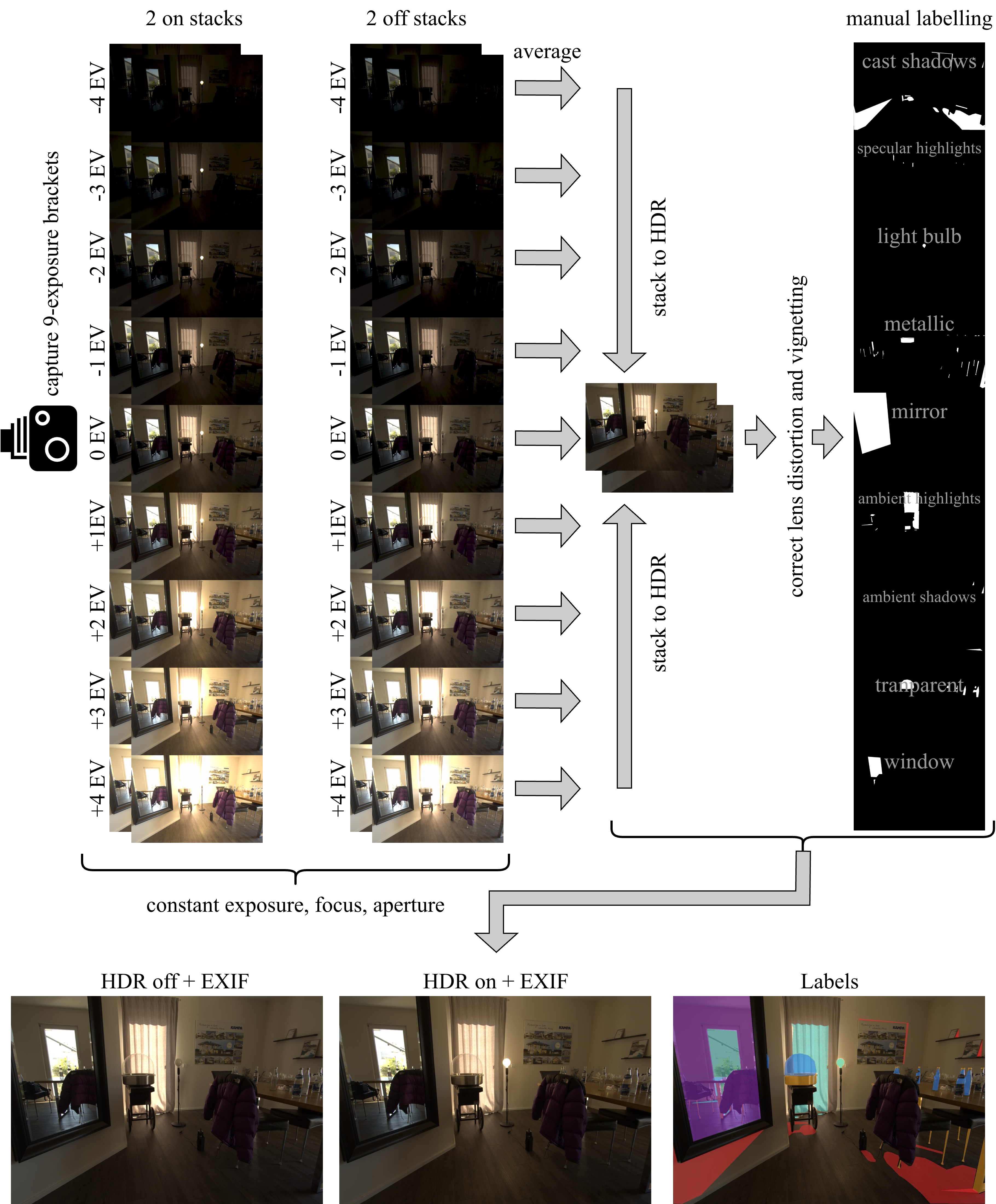}
    \caption{\textbf{Capture process.} For each pair, we capture two 9-exposure HDR brackets using a Sony $\alpha$6600 with 14-bit precision, once for the \textit{on}-image and once for the \textit{off}-image. To ensure both images have an identical ambient light noise level, we fix the aperture, focus, and exposure time for all images in a given view. The exposure is chosen to produce a well-exposed \textit{on}-image. After capturing, we average the individual RAW exposures to reduce noise and then fuse them into a single HDR \textit{on}/\textit{off} image. Then, we correct for lens distortion and vignetting using the EXIF data recorded by the camera. Finally, we annotate the images. The entire recording and processing pipeline takes up to 5 minutes per image pair (not including labelling).}
    \label{supp:data_capturing}
\end{figure}

\paragraph{Data capture and processing pipeline} To construct the 3DLP dataset, we use the following pipeline to isolate the exact light transport of our probe. The process is illustrated in \cref{supp:data_capturing}. For each viewpoint, we capture data in two states: with the spherical light probe turned \textit{on} and \textit{off}. We utilise a Sony $\alpha$6600 camera capturing 14-bit RAW images. We use a Samyang AF12mm F2.0 wide-angle lens due to the limited space in indoor spaces. The used light probe is an OSRAM LED STAR GLOBE125 150 with 2452\,lm and a colour temperature of 4,000\,K. The beam angle of this bulb is specified as ``up to $360^\circ$'' in its datasheet. In a qualitative comparison with other light bulbs we found this bulb to produce the most omnidirectional illumination.

To ensure that the ambient lighting and sensor noise profile remain identical across both states, we strictly lock the camera's aperture, focus, and base exposure time for each viewpoint. The base exposure is calibrated to produce a well-exposed image when the light probe is \textit{turned on}. For both the \textit{on} and \textit{off} state, we capture two 9-exposure bracketing sequences. 

In post-processing, the individual RAW exposures are first averaged to mitigate sensor noise and then fused into high-fidelity, high-dynamic-range (HDR) images. We subsequently apply corrections for lens distortion and vignetting utilising the camera's embedded EXIF data, ensuring geometric and photometric accuracy. The full capturing and processing can take up to five minutes per pair.

\begin{figure}[htbp]
    \centering
    \includegraphics[width=\textwidth]{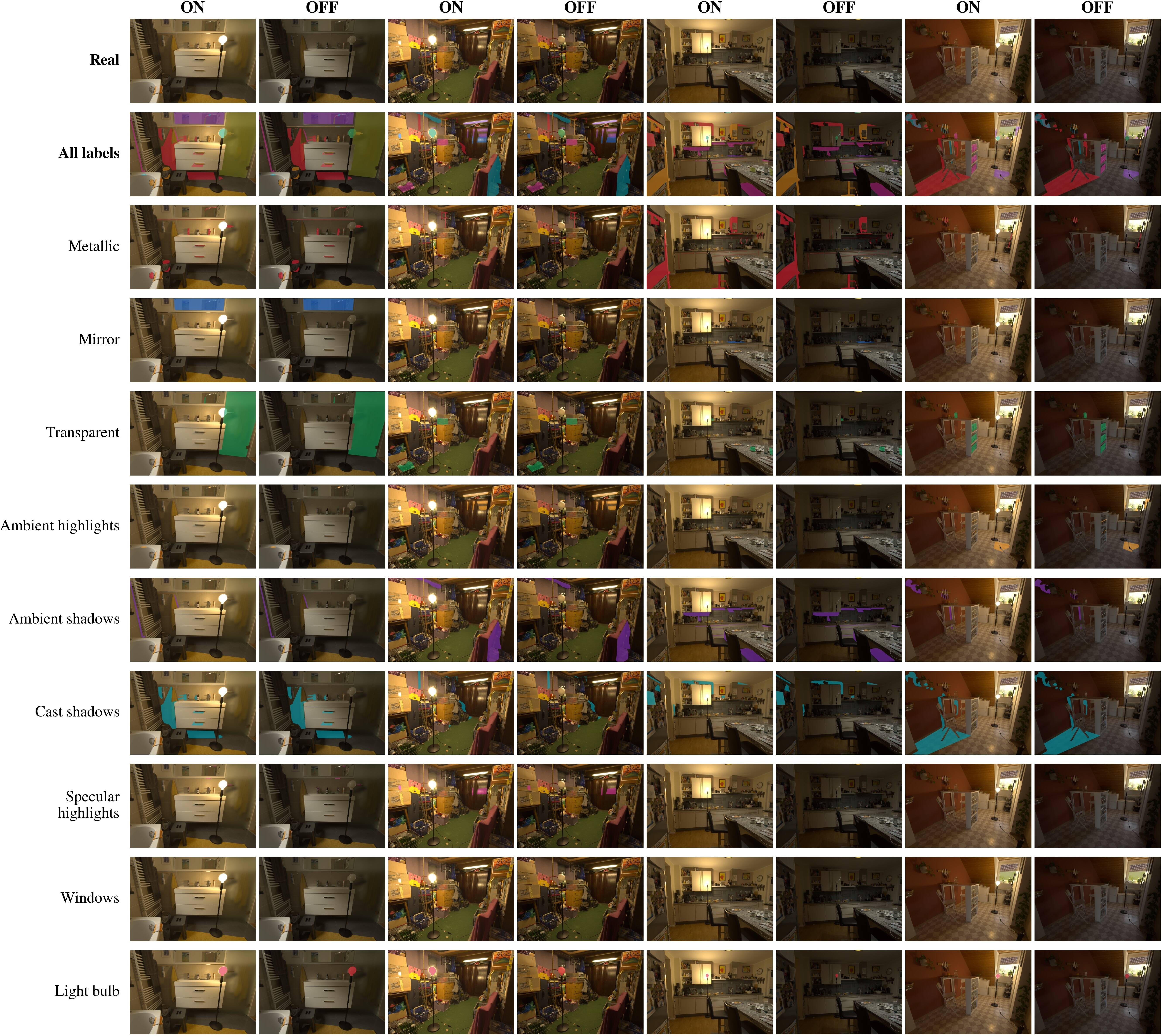}
    \caption{\textbf{Annotation examples.} Each label is visualised with a different colour. Please note that we prioritise annotation correctness over exhaustive completeness.}
    \label{supp:label_figure}
\end{figure}

\paragraph{Material and light effect annotations} We visualise our annotations in \cref{supp:label_figure}. The processed image pairs were annotated using the Roboflow platform. To ease labelling of complex light interactions, such as cast shadows and specular highlights, we computed difference images, i.e. $I_R^{\textrm{on}} - I_R^{\textrm{off}}$, to clearly isolate the probe's direct illumination.

\section{Cost and compute to run the 3DLP Benchmark}
\label{sec:Cost_Compute}
We evaluated both open-weights and closed-source image editing models to conduct our proposed 3DLP benchmark. \cref{tab:compute_costs} details the estimated compute resources and financial expenditures required for all experiments.
Note that for the main benchmark $\approx 2000$ images needed to be generated per model. For the models we used in \cref{sec:LightSourceTypes} and \cref{supp:ambient_light_complexity_evaluation} we generated an additional $\approx 1000$ images for our experiment split.
For the open-source models, we report the estimated GPU compute time utilising NVIDIA H200 and A100 GPUs. We measured a rough average generation time of approximately 60 seconds per image. For the closed-source models accessed via API, we report the estimated financial cost in USD based on per-image pricing, using Batch-API where available. 

Please note that the numbers presented in \cref{tab:compute_costs} represent a lower-bound estimate. The actual computational footprint and financial expenditure might be higher in practice. This discrepancy is due to iterative pipeline adjustments, failed generations, and the necessity to regenerate certain images throughout the experimental process.

\begin{table}[htbp]
    \centering
    \caption{\textbf{Estimated compute and cost for image generation.} Open-source GPU hours are calculated assuming 60 seconds per image. API costs are based on fixed per-image rates.}
    \label{tab:compute_costs}
    \begin{tabular}{lllrrc}
        \toprule
        \textbf{Model} & \textbf{Category} & \textbf{Hardware / Access} & \textbf{Images} & \textbf{GPU Hours} & \textbf{Cost (USD)} \\
        \midrule
        Qwen-Img.-Edit & Open-Source & NVIDIA H200 & 3,000 & 50.0 & -- \\
        Flux 2 Dev      & Open-Source & NVIDIA H200 & 3,000 & 50.0 & -- \\
        Bagel 7B MoT          & Open-Source & NVIDIA A100 & 2,000 & 33.3 & -- \\
        OmniGen2         & Open-Source & NVIDIA A100 & 2,000 & 33.3 & -- \\
        \midrule
        Nano Banana Pro & Closed-Source & API (\$0.07/img) & 3,000 & -- & \$210.00 \\
        Nano Banana 2   & Closed-Source & API (\$0.07/img) & 3,000 & -- & \$210.00 \\
        Flux 2 Max        & Closed-Source & API (\$0.10/img) & 3,000 & -- & \$300.00 \\
        GPT Img. 1.5   & Closed-Source & API (\$0.10/img) & 3,000 & -- & \$300.00 \\
        \midrule
        \textbf{Total Estimate} & & & \textbf{22,000} & \textbf{$\sim$166.6} & \textbf{$\sim$\$1,020.00} \\
        \bottomrule
    \end{tabular}
\end{table}

\section{Prompt selection}
\label{subsec:Prompt Sweep}
For each model we test different prompts to find one that the model consistently understands, i.e. in which the model turns on/off the correct lamp. Note, that we \textbf{do not} select prompts to optimise for our metrics or to improve the models light transport capabilities. Prompts were chosen by human visual inspection.

\begin{figure}[t]
    \centering
    \includegraphics[width=\textwidth]{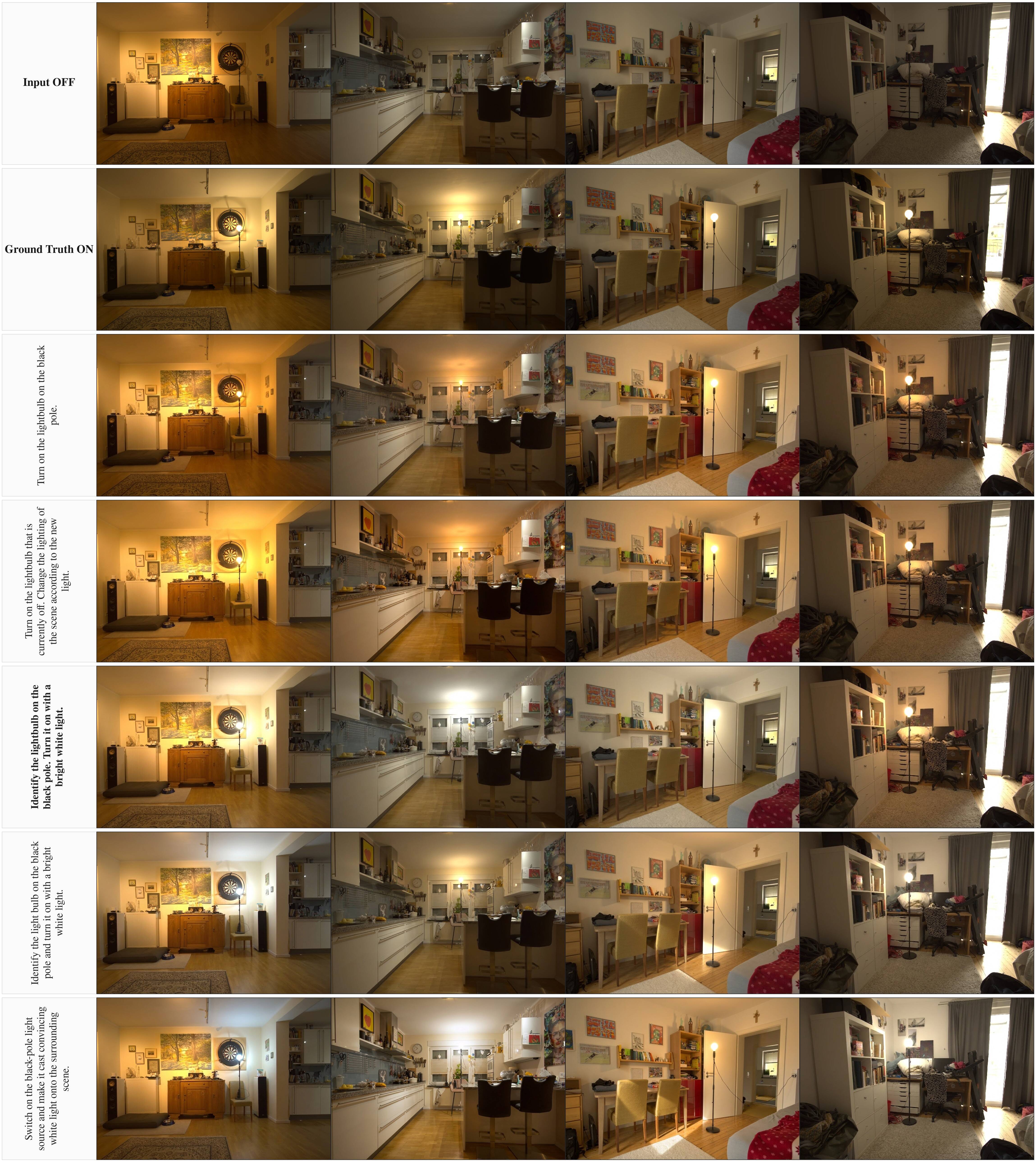}
    \caption{\textbf{Prompt selection for Nano Banana Pro:} It is evident that Nano Banana Pro reliably completes and understands the task. Across all prompts, the model successfully detects the correct lamp and turns on its light while keeping the ambient light unchanged. Based on visual inspection, we decided to use the prompt in bold.}
    \label{fig:PromptSweep_NanoBananaPro}
\end{figure}

\begin{figure}[t]
    \centering
    \includegraphics[width=\textwidth]{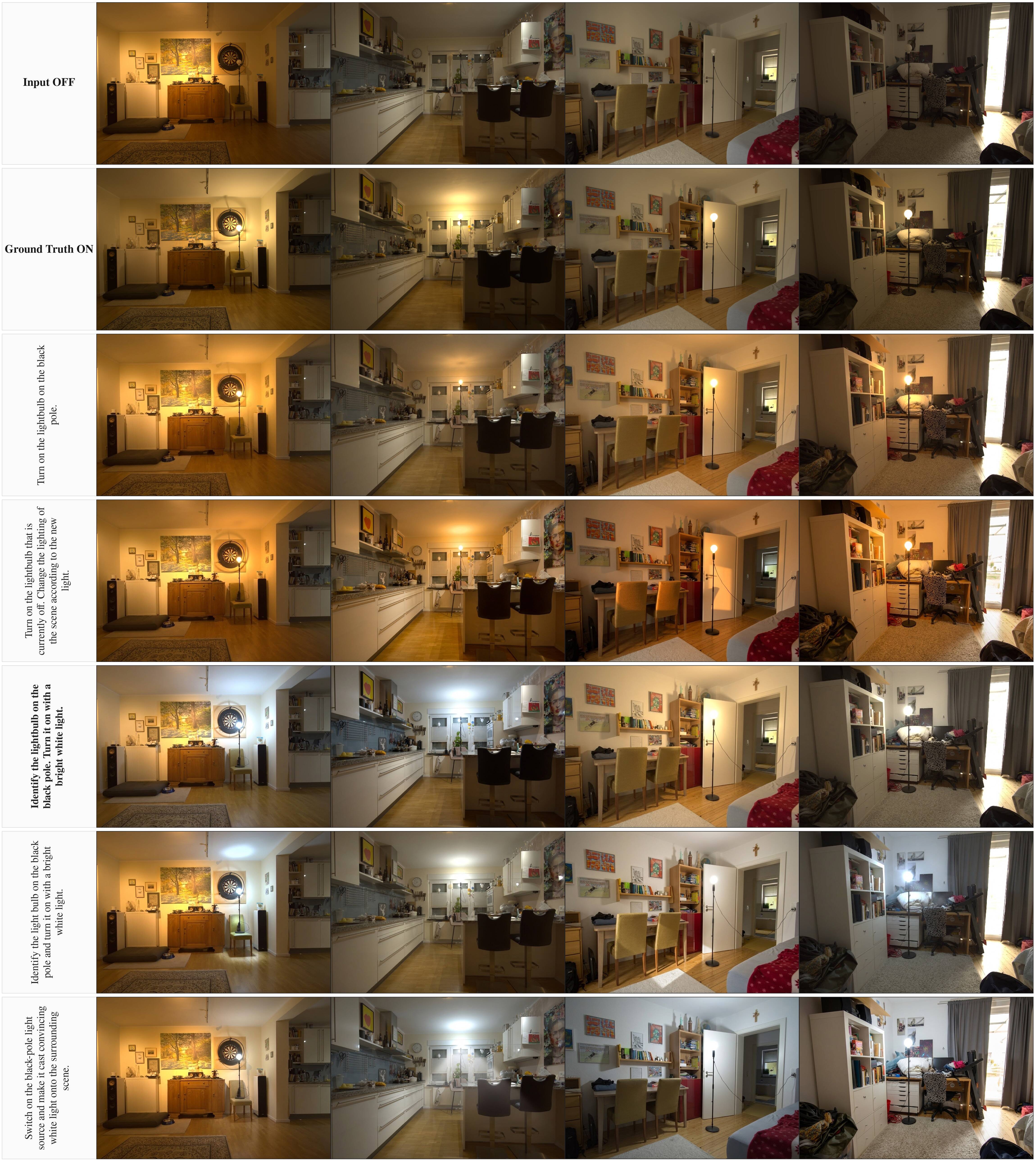}
    \caption{\textbf{Prompt selection for Nano Banana 2:} Like Nano Banana Pro, Nano Banana 2 has no problem in executing the task correctly. Similar to the Pro version, we selected the prompt based on visual inspection.}
    \label{fig:PromptSweep_NanoBananaFlash}
\end{figure}

\begin{figure}[t]
    \centering
    \includegraphics[width=\textwidth]{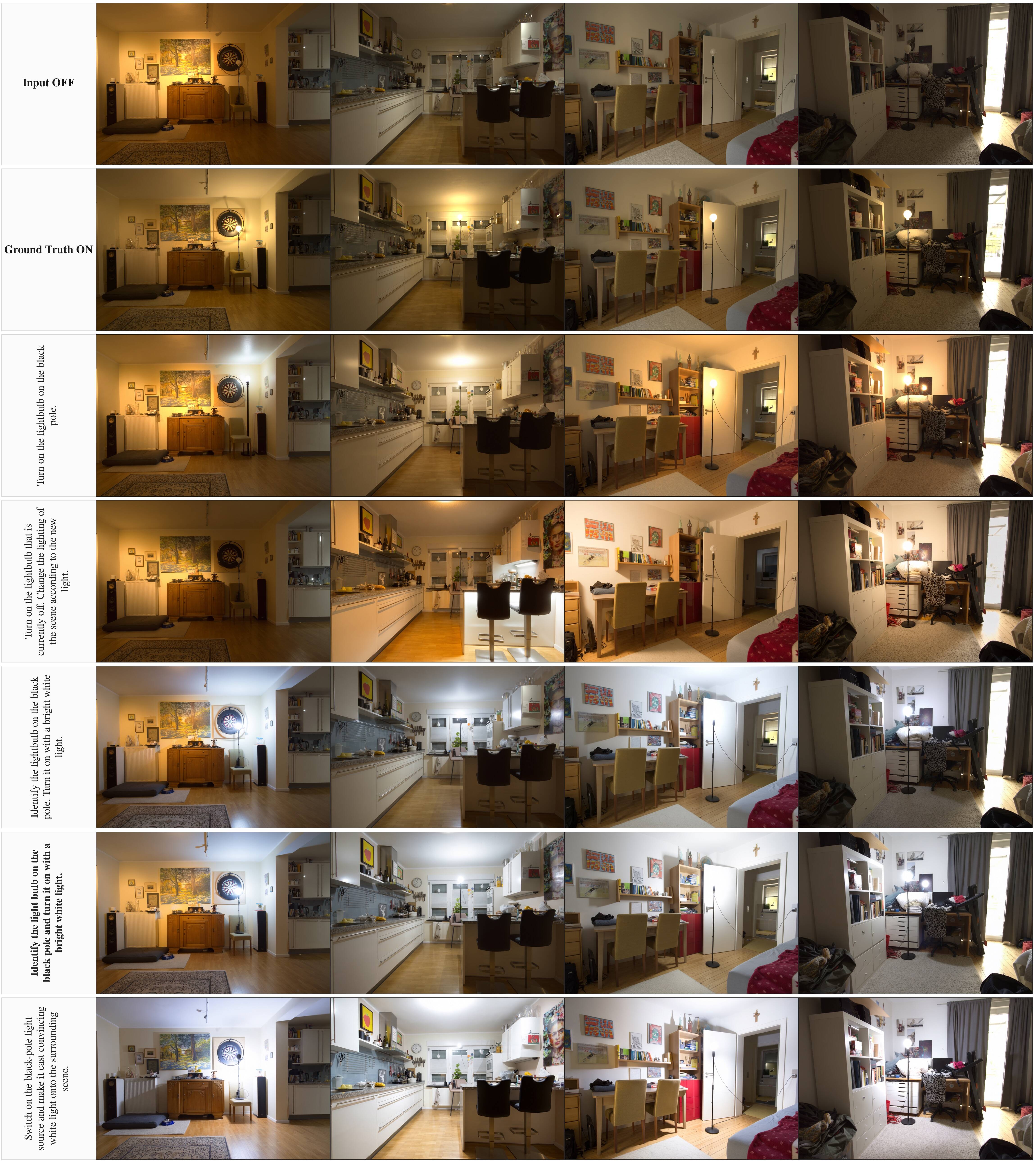}
    \caption{\textbf{Prompt selection for Qwen-Image-Edit:} As shown, Qwen-Image-Edit exhibits variations in performance depending on the prompt. It fails to understand the task for the prompt used in row 4. In three out of the four images, Qwen-Image-Edit turns on the wrong lamp or does not illuminate the scene at all. Based on visual inspection, we decided to use the prompt marked in bold.}
    \label{fig:PromptSweep_Qwen}
\end{figure}

\begin{figure}[t]
    \centering
    \includegraphics[width=\textwidth]{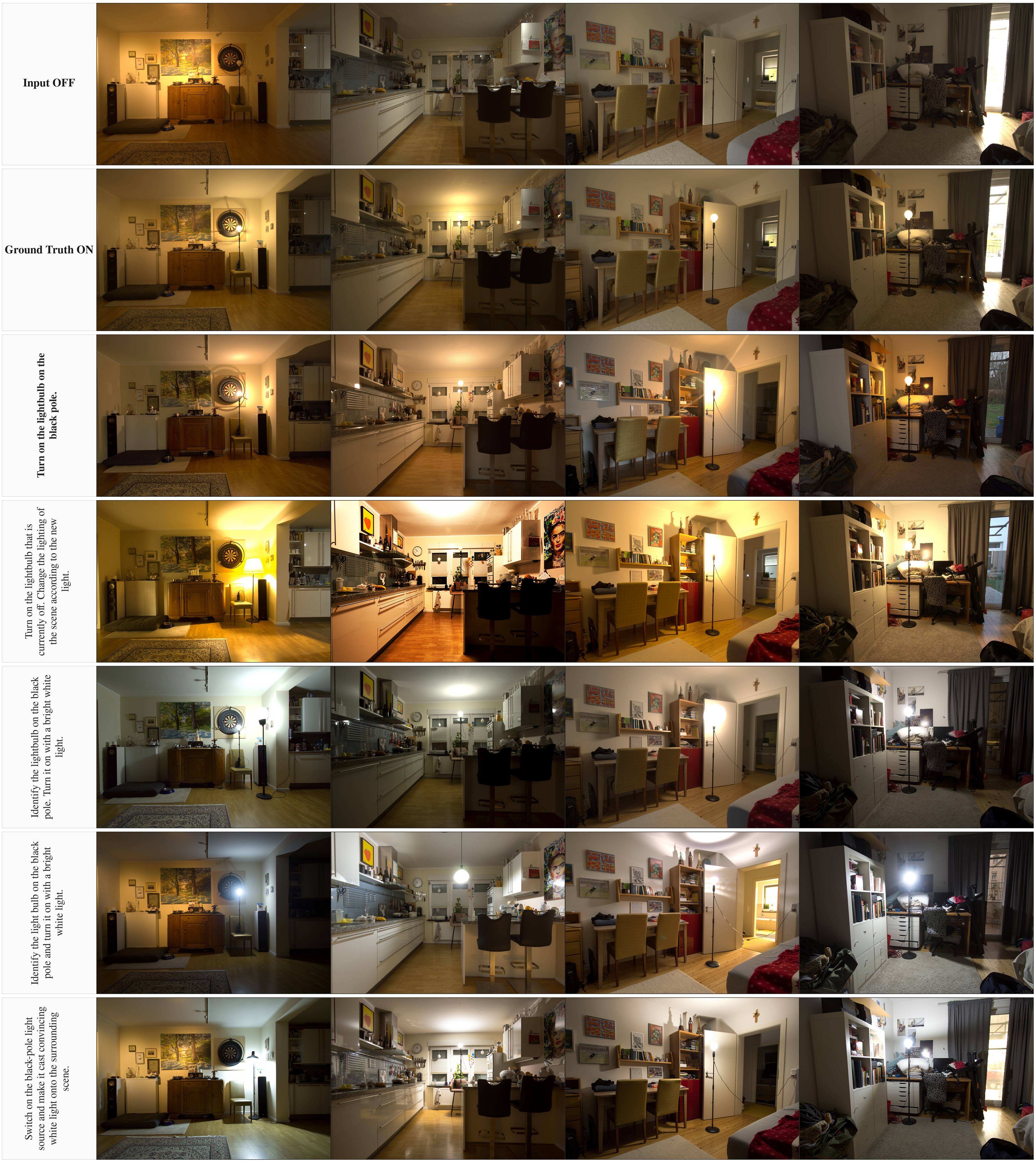}
    \caption{\textbf{Prompt selection for Flux 2 Max:} As seen in rows 5-7, Flux 2 Max tends to produce highly variable results for certain prompts. We obtained the most consistent results using a simple prompt: ``Turn on the light bulb on the black pole.''}
    \label{fig:PromptSweep_FluxMax}
\end{figure}

\begin{figure}[t]
    \centering
    \includegraphics[width=\textwidth]{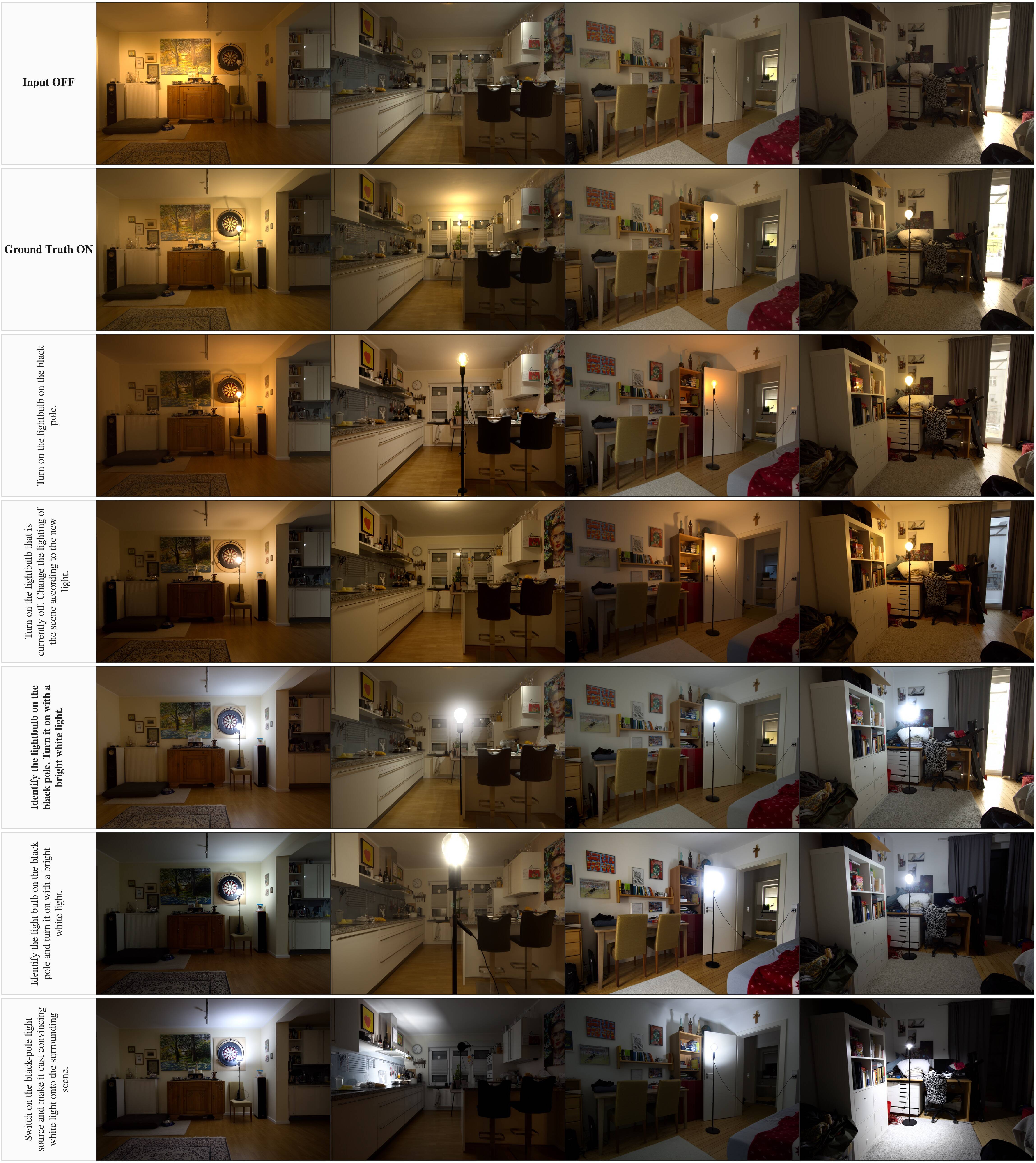}
    \caption{\textbf{Prompt selection for Flux 2 Dev:} As with Flux 2 Max, the choice of prompt affects the performance of Flux 2 Dev considerably, more than other models like Qwen-Image-Edit or Nano Banana Pro. Based on metrics and visual inspection, we chose the prompt printed in bold. Even though Flux 2 Dev misinterpreted the lamp position in one of the examples, the metrics for this prompt remained best.}
    \label{fig:PromptSweep_Flux-Dev}
\end{figure}

\begin{figure}[t]
    \centering
    \includegraphics[width=\textwidth]{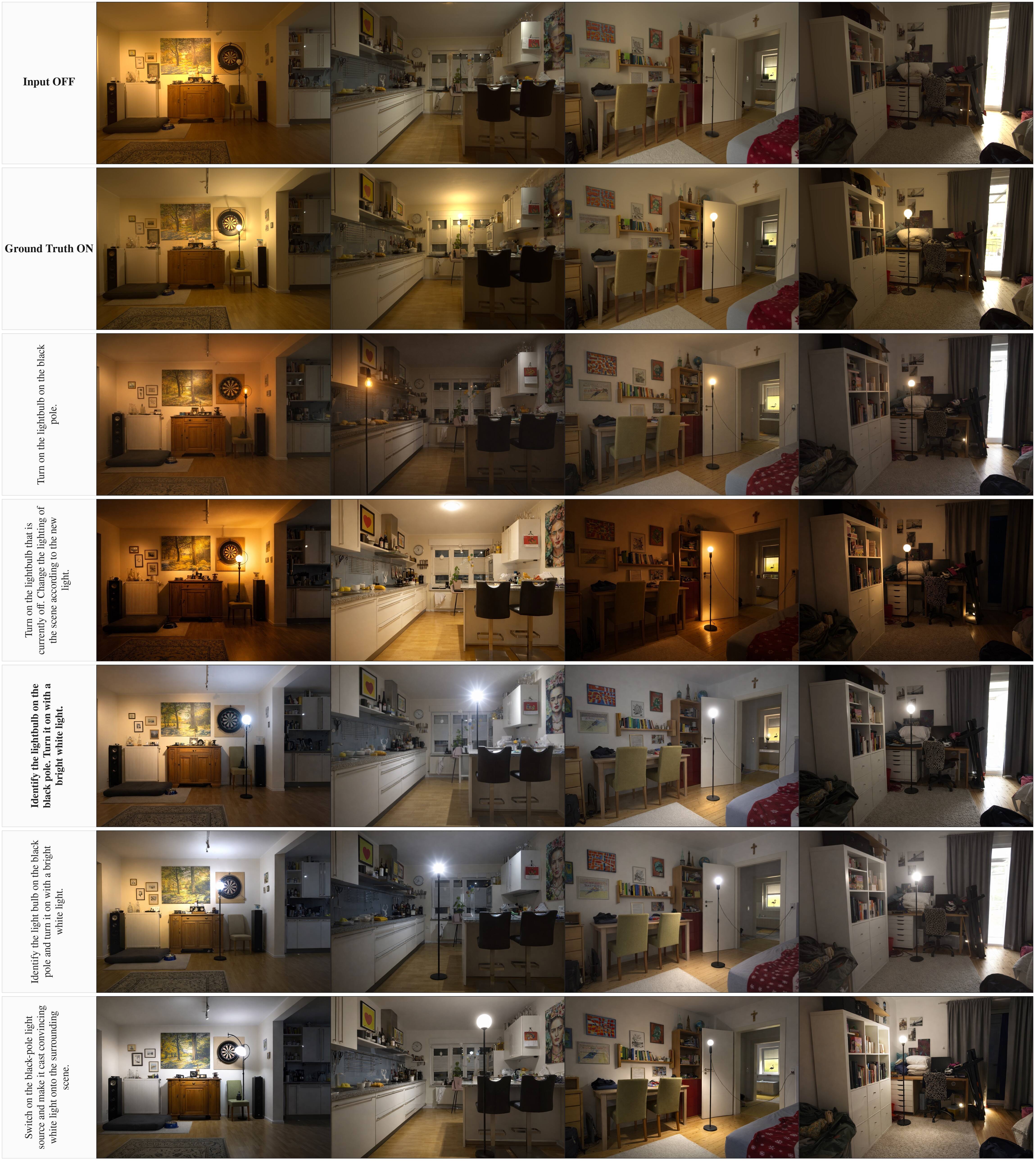}
    \caption{\textbf{Prompt selection for GPT Image 1.5:} For GPT Image 1.5, the 5th row produced the most consistent results.}
    \label{fig:PromptSweep_GPTImage}
\end{figure}

\begin{figure}[t]
    \centering
    \includegraphics[width=\textwidth]{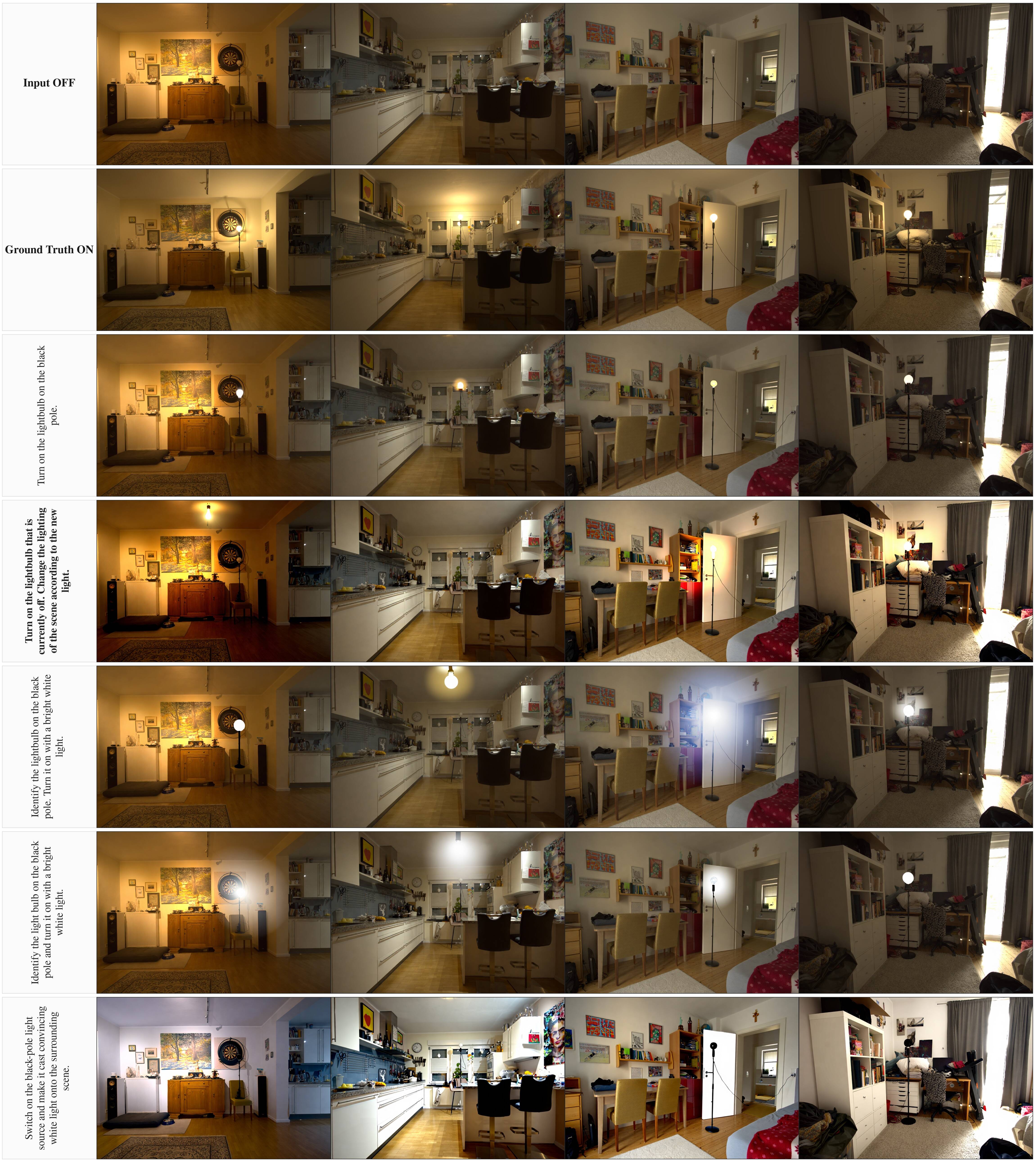}
    \caption{\textbf{Prompt selection for Bagel 7B MoT:} As seen in the images, Bagel struggles to illuminate the scene, even though it turns on the right light most of the time. Based on performance metrics we chose the prompt in row 4 (bold). Note that due to these problems the model was not selected for the main evaluation.}
    \label{fig:PromptSweep_Bagel}
\end{figure}

\begin{figure}[t]
    \centering
    \includegraphics[width=\textwidth]{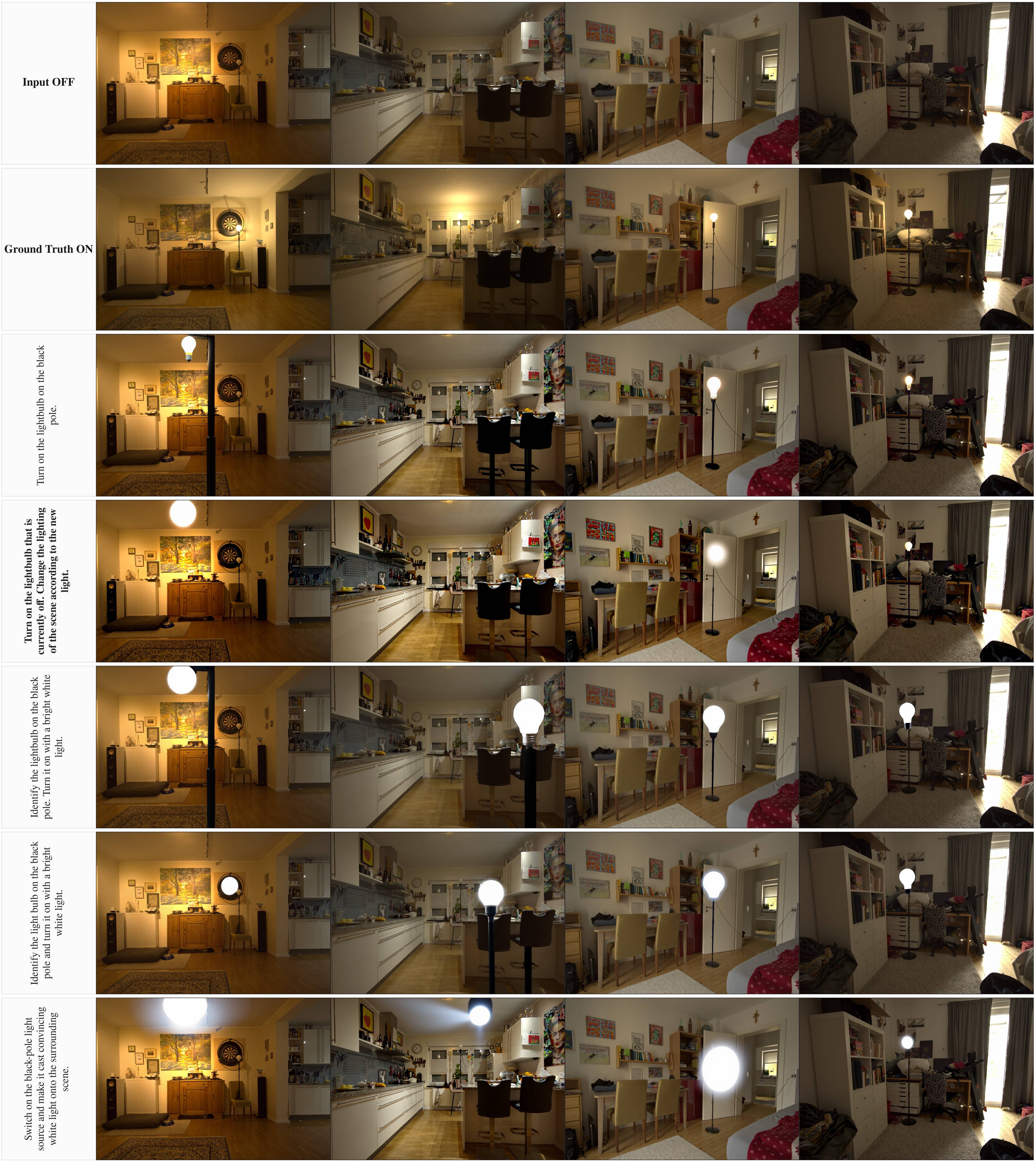}
    \caption{\textbf{Prompt selection for OmniGen2:} This model struggles the most with turning the correct light on. Oftentimes, it just applies a bright spot to the image. However, using the prompt in row 4, OmniGen2 was still able to illuminate the scene somehow. Note that due to these problems the model was not selected for the main evaluation.}
    \label{fig:PromptSweep_Omnigen}
\end{figure}

\cref{fig:PromptSweep_Qwen,fig:PromptSweep_Bagel,fig:PromptSweep_FluxMax,fig:PromptSweep_Omnigen,fig:PromptSweep_Flux-Dev,fig:PromptSweep_GPTImage,fig:PromptSweep_NanoBananaPro,fig:PromptSweep_NanoBananaFlash} show a selection of prompts that we tested for each model. The prompt that was finally used for each respective model is printed bold. In total, we tested all models with 14 different prompts. For visualisation purposes, we only show a selection of them.

\clearpage

\end{document}